\documentclass[10pt,journal,compsoc]{IEEEtran}
\makeatletter

\newcommand{\Rmnum}[1]{\expandafter\@slowromancap\romannumeral #1@}
\makeatother
\usepackage[colorlinks,urlcolor=blue,linkcolor=blue,citecolor=blue]{hyperref}
\usepackage{diagbox}
\usepackage{amsmath}
\usepackage{enumerate}
\usepackage{graphicx}
\usepackage{subfigure}
\usepackage{booktabs}
\usepackage{threeparttable}
\usepackage{bbding}
\usepackage{array}
\usepackage{amsfonts,amssymb}
\usepackage{float}
\usepackage{multirow}
\usepackage{color}
\usepackage{bm}
\usepackage{cite}
\usepackage[colorlinks,linkcolor=blue]{hyperref}
\newcounter{RNum}
\renewcommand{\theRNum}{\arabic{RNum}}
\newcommand{\Remark}{\noindent\textbf{Remark}~\refstepcounter{RNum}\textbf{\theRNum}: }
\ifCLASSINFOpdf
\else
\fi
\begin{document}
\title{Correlation Filters for Unmanned Aerial Vehicle-Based Aerial Tracking: \\ A Review and Experimental Evaluation}

\author{Changhong~Fu*, Bowen Li, Fangqiang Ding, Fuling Lin, and Geng Lu
% <-this % stops a space
\IEEEcompsocitemizethanks{
\IEEEcompsocthanksitem *Corresponding author
\protect
\IEEEcompsocthanksitem Changhong Fu, Bowen Li, Fangqiang Ding, and Fuling Lin are with the School of Mechanical Engineering, Tongji University, 201804 Shanghai, China.
\protect\\
Email: changhongfu@tongji.edu.cn.
\IEEEcompsocthanksitem Geng Lu is with Department of Automation, Tsinghua University, 10084 Beijing, China.
\protect\\
Email: lug@tsinghua.edu.cn.
}% <-this % stops an unwanted space
}


\markboth{}%
{Shell \MakeLowercase{\textit{et al.}}: Bare Demo of IEEEtran.cls for Computer Society Journals}



\IEEEtitleabstractindextext{%
\begin{abstract}
Aerial tracking, which has exhibited its omnipresent dedication and splendid performance, is one of the most active applications in the remote sensing field. Especially, unmanned aerial vehicle (UAV)-based remote sensing system, equipped with a visual tracking approach, has been widely used in aviation, navigation, agriculture, transportation, and public security, \emph{etc}. As is mentioned above, the UAV-based aerial tracking platform has been gradually developed from research to practical application stage, reaching one of the main aerial remote sensing technologies in the future. However, due to the real-world onerous situations, \textit{e.g.}, harsh external challenges, the vibration of the UAV’s mechanical structure (especially under strong wind conditions), the maneuvering flight in complex environment, and the limited computation resources onboard, accuracy, robustness, and high efficiency are all crucial for the onboard tracking methods. Recently, the discriminative correlation filter (DCF)-based trackers have stood out for their high computational efficiency and appealing robustness on a single CPU, and have flourished in the UAV visual tracking community. In this work, the basic framework of the DCF-based trackers is firstly generalized, based on which, 23 state-of-the-art DCF-based trackers are orderly summarized according to their innovations for solving various issues. Besides, exhaustive and quantitative experiments have been extended on various prevailing UAV tracking benchmarks, \emph{i.e.}, UAV123, UAV123@10fps, UAV20L, UAVDT, DTB70, and VisDrone2019-SOT, which contain 371,903 frames in total. The experiments show the performance, verify the feasibility, and demonstrate the current challenges of DCF-based trackers onboard UAV tracking. Besides, this work also implements the brilliant DCF-based trackers on a typical CPU-based onboard PC to achieve real flight UAV tracking tests to further validate their real-time capabilities and robustness under challenging scenes. A concise summary of future research trends in the area of DCF-based methods for UAV tracking is further provided. Finally, comprehensive conclusions on the directions for future research are presented.
\end{abstract}

}

\maketitle

\IEEEdisplaynontitleabstractindextext

% \fi
\IEEEpeerreviewmaketitle

\IEEEraisesectionheading{\section{Introduction}\label{sec:introduction}}

\IEEEPARstart{A}{erial} visual object tracking is currently attracting a cornucopia of attention and developing rapidly in the field of remote sensing \cite{cao2016RS, Bi2019IEEEAccess,Uzkent2019TGRS,Uzkent2016STAEORS,Fu2020TGRS}. Especially for the widely-used unmanned aerial vehicle (UAV) platforms \cite{Yi2017ICUS,Gu2019ITNEC,Shi2018ICMTMA}, which possess small size, flexible motion, and high safety, when equipped with visual tracking techniques, has flourished in extensive applications, \emph{e.g.}, wildlife rescue \cite{olivares2015Sensors}, target following \cite{Vanegas2017AC,cheng2017IROS,Yao2017ICRA}, vehicle tracking \cite{cao2016RS}, disaster response \cite{yuan2017JIRS,yuan2015ICUAS}, cinematography \cite{Bonatti2019IROS}, infrastructure inspection \cite{Martins2020IJRA,Hamelin2019ICUAS}, \textit{etc.} Specifically, the mobile UAV usually needs to continuously locate (and follow) one certain object, where real-time as well as robust single object tracking algorithm is essential. Nevertheless, under complex scenes onboard UAVs, achieving robust, accurate, and real-time tracking is a very challenging task. Compared with general tracking scenes, where the camera is usually static or slow-moving and fewer geometric and photometric variations exist, UAV tracking is confronted with more onerous challenges as follows:
\begin{itemize}
	\item Inadequate sampling resolution: The large visual scope of UAV results in more background information, ending up in the reduction of the resolution of the object, and hence weak model representations. Weak model representation can make the tracker easy to lose the object because of its poor discriminative ability.  
	
	\item Fast motion issue: UAV has a large degree of freedom and a high degree of mobility, thus making the fast motion of both the UAV and the tracking object, which bring even greater challenges to the tracking task. Besides, in the process of flight, UAV usually inevitably encounters mechanical vibration, especially under the influence of strong wind, which may even result in motion blur. Such kinds of rapid change of the object location can be more challenging for the trackers.
	
	\item Severe visual occlusion: As a common phenomenon in UAV tracking, partial or even full occlusion may cause severe degradation of the object, and then lead to tracking failure.
	
	\item Acute illumination variation: The illumination conditions for UAV can change rapidly, including from bright to dim, from indoors, canopy, and shadow regions to bright outdoors or even under sunlight. Such scenes can cause a large variety of object appearance, thus making tracking challenging. Besides, complex and harsh scenes are often encountered, such as the poor illumination conditions at night, in rainy or foggy days, and other scenes, which make it hard for the tracker to distinguish the object from the background.
	
	\item View point change: As a common scene, UAV may fly 360 degrees around the object, which makes the onboard camera capture different aspects of the 3D object, \emph{e.g.}, the back and front sides of a person, where the object appearance can undergo severe variations. Under such scenes, the trackers may lose the object without timely online learning and model update.
	
	\item Scarce computation resources: Because of the limited power supply and payload issues, most UAVs only use a single CPU as the computing platform, which greatly limits the processing speed. To meet the real-time requirements of UAV tracking, the methods need to be carefully designed to realize robust tracking without casting aside high efficiency. Besides, the algorithms need to be light-weighted enough to leave more power supplies for energy-consuming applications like maneuvering flight of UAV in complex environment.
\end{itemize}

\begin{figure*}[!t]
	\centering
	\includegraphics[width=2.04\columnwidth]{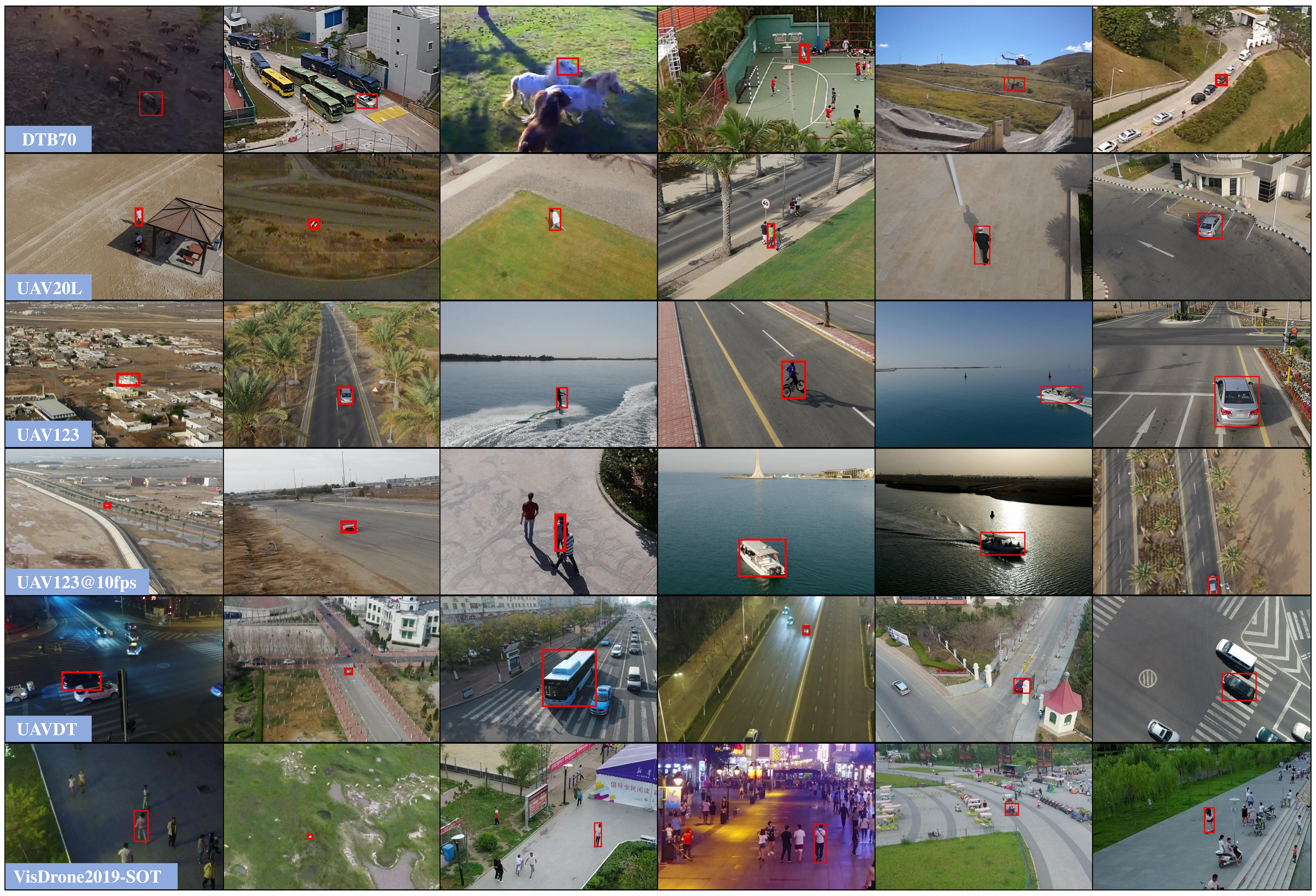}
	\vspace{-0.5cm}
	\caption{Representative scenes in 6 well-known UAV tracking benchmarks. The benchmarks from top to bottom are DTB70~\cite{li2017AAAI}, UAV20L~\cite{mueller2016ECCV}, UAV123~\cite{mueller2016ECCV}, UAV123@10fps~\cite{mueller2016ECCV}, UAVDT~\cite{du2018ECCV}, and Visdrone2019-SOT~\cite{du2019ICCVW}. Here, target ground-truths are marked out by \textbf{\textcolor{red}{red}} rectangles. Common challenges in UAV tracking are shown, \emph{e.g.}, low resolution, fast motion, similar objects, viewpoint change, illumination variation, and occlusion in UAV123@10fps (from left to right). For better display effect, please refer to the electronic version of this paper.}
	\vspace{-0.5cm}
	\label{fig:1}
\end{figure*}

Figure~\ref{fig:1} shows representative and challenging UAV-based aerial tracking scenes. Due to the challenges mentioned above, the research and development of a fast and robust visual tracking approach are extremely critical and valuable for the considerable prospect of UAV-based remote sensing applications. 

Among various tracking methods, discriminative correlation filter (DCF)-based trackers usually possess both high speed and accuracy. The most important characteristic and an amazing highlight of the DCF-based methods are that they transform the calculation of cyclic correlation or convolution in the spatial domain to the element-wise multiplication in the frequency domain through discrete Fourier transform. Such a strategy greatly enhances the operation speed of DCF-based trackers, most of which reaches over 30 frames per second (FPS) on a single CPU platform, thus meeting the real-time requirements on UAV. Due to the excellent and extraordinary performance of DCF-based trackers on a single CPU platform, the past years have witnessed the rapid expansion and the favourable results of DCF-based trackers onboard UAV tracking \cite{Fu2020TGRS,Li2020CVPR,Lin2020ICRA,Huang2019ICCV,Fu2018ROBIO,Li2020IROS,Ding2020IROS}.

Even though there have been some researches mentioning or summarizing DCF-based object tracking \cite{wu2015TPAMI,Liu2020CIS}, they paid little attention to UAV-based aerial tracking scenes (which is more complex, challenging, and resource-limited). Besides, until now, most studies reviewing UAV in the field of remote sensing have either been general reviews concerning the applications of the UAV platform \cite{nex2014AG,adao2017RS}, or detailed strategies used to control UAV \cite{Chand2017ICEECCOT,Hu2017ICUS}, which all do not focus on robust, accurate, and real-time tracking methods under complex scenes onboard UAV. 

To the best of our knowledge, there exist very few reviews about UAV real-time tracking in recent years, let alone focusing on the performance of DCF-based trackers in UAV scenarios. In other words, a systematic and comprehensive review concerning the DCF-based tracking algorithms applied to the UAV platform has not yet been conducted. DCF-based trackers have been utilized widely in the UAV tracking community, and the quantity of the related publications is currently increasing remarkably, so it appears that a systematic summary and analysis is necessary to get a comprehensive and objective understanding of the superiority and practicability of DCF for UAV tracking. Therefore, this work is conducted to offer an overall review of the DCF-based tracking algorithms in the past decade, introducing their special contributions. This work also performed extensive experiments and analyzed the advanced representative DCF-based trackers on various authoritative UAV benchmarks \cite{mueller2016ECCV,du2018ECCV,li2017AAAI,du2019ICCVW} to demonstrate their reliability and superiority in aerial tracking. Extensive experiment results confirm the excellence of the DCF-based algorithms in UAV tracking and even more exciting prospects can be expected for DCF in UAV tracking. Based on the results, we also give key bottlenecks of the DCF-based methods. Besides, the brilliant DCF-based trackers \cite{Huang2019ICCV,Li2020CVPR} are also implemented on a typical CPU-based onboard PC, \textit{i.e.}, Intel NUC8i7HVK, to achieve the UAV tracking tests. Potential directions are also indicated in this work, guiding further researches into DCF for UAV tracking.

The main contributions of this work are four-fold, which are formally summarized as follows:
\begin{itemize}
	\item \emph{Comprehensive review}. This work offers a general framework of DCF-based trackers and summarizes most state-of-the-art DCF-based trackers according to their innovations and contributions.
	\item \emph{Code library}\footnote{The code library and experimental evaluation results are located at \url{https://github.com/vision4robotics/DCFTracking4UAV}.}. This work integrates most of the publicly available DCF-based trackers in one code library. Besides, our experimental results are also organized for convenient reference.
	\item \emph{Experiment evaluation}$^1$. This work extends exhaustive experiments of the DCF-based trackers on 6 authoritative UAV benchmarks, \emph{i.e.}, UAV123, UAV123@10fps, UAV20L, UAVDT, DTB70, and Visdrone2019-SOT, to demonstrate their performance under complex scenes and their superiority against other types of trackers.
	\item \emph{Onboard test}. This work further implements the DCF-based trackers on a typical CPU-based onboard PC to achieve real flight UAV tracking tests, where their real-time capabilities and robustness under hash scenes are validated.
\end{itemize}

The remainder of this article is organized as follows: Section~\ref{Section 2} generally reviews state-of-the-art methods in the visual tracking community, presents the basic framework of DCF-based trackers, and illustrates what makes DCF-based methods suitable for UAV. Section~\ref{Section 3} presents the gradual evolution and gratifying innovations of the distinguished DCF-based tracking methods. Besides, this section integrates the trackers based on their contribution in a clear logic. Section~\ref{Section 4} provides abundant experimental results and analyses based on the trackers' performances on each UAV benchmark to demonstrate their practicability, superiority, and also tracking results in UAV special tracking challenges. Based the evaluation results, failure cases that are currently not well addressed are concluded. Moreover, in Section~\ref{Section 5}, the performances on the onboard PC in real flight tracking tests are exhibited, where the DCF-based methods are proved to be suitable and capable for UAV tracking. Further, Section~\ref{Section 6} offers potential future work in the field of DCF for aerial tracking. Finally, Section~\ref{Section 7} gives a conclusion about this work.
\\
\\
\\

\section{Related Works}\label{Section 2}
\subsection{Object tracking methods}
According to different representation schemes, object tracking methods can be generally divided into two types: generative methods \cite{Yang1995TSP,Mei2011TPAMI,Wang2013ICCV} and discriminative methods \cite{Avidan2004TPAMI,Held2016ECCV,Henriques2015TPAMI}. 

\subsubsection{Generative method}
The main idea of the generative method is to learn a feature template from the target area in the first frame and find the most similar and matching area to the template appearance in the search region of the subsequent frames as the tracking result. As is known, most of the early object tracking methods belong to the generative method. For instance, B. D. Lucas and
T. Kanade \emph{et al.} \cite{lucas1981IJCAI} proposed holistic templates which are based on raw intensity values. In order to cope with appearance changes, subspace-based tracking methods emerged \cite{degroat1992TSP,Yang1995TSP,badeau2005TSP}. As a well-known branch in the generative method, numerous tracking approaches based on sparse representation have also grabbed researchers' eyes \cite{lan2015TIP,Mei2011TPAMI,bai2012PR,Lu2012CVPR}. Nevertheless, the weaknesses of generative methods are obvious. The first is that a large number of training samples require abundant computation resources, which makes it difficult to meet real-time requirements onboard UAV. Then, traditional generative approaches have neglected the background information, which may help ensure more robust tracking. Thirdly, the generative methods usually suppose the appearance of the object won't undergo large changes in a period, while appearance variation frequently happens in UAV tracking scenes.
Recently, the discriminative method has become the mainstream of the visual tracking community.

\subsubsection{Discriminative method}
Different from the generative method, the core idea of the discriminative method (which is also called the tracking-by-detection method) is to train a classifier that can distinguish the tracking object from the background.

Belonging to discriminative method, support vector machine (SVM) based methods are the early ones that flourished in visual tracking community \cite{Avidan2004TPAMI,Ning2016CVPR,Hare2016TPAMI,Bai2012CVPR}. More concretely, S. Avidan \emph{et al.} \cite{Avidan2004TPAMI} constructed pyramids from the support vectors and used a coarse-to-fine approach in the classification stage in consideration of large motion issue. Y. Bai \emph{et al.} \cite{Bai2012CVPR} presented online Laplacian ranking support vector tracker to robustly locate the object. J. Ning \emph{et al.} \cite{Ning2016CVPR} proposed a simple while effective dual linear structured support vector machine (DLSSVM) to boost their tracking efficiency. STRUCK \cite{Hare2016TPAMI}, proposed by S. Hare \emph{et al.}, achieved great tracking result due to its kernelised structured output SVM, which is utilized to provide adaptive tracking. Though the performances of the trackers based on SVM are promising, large-scale training samples will consume much machine memory and computing time, greatly limiting their real-time capabilities onboard UAV.

\begin{figure*}[!t]
	\includegraphics[width=2.05\columnwidth]{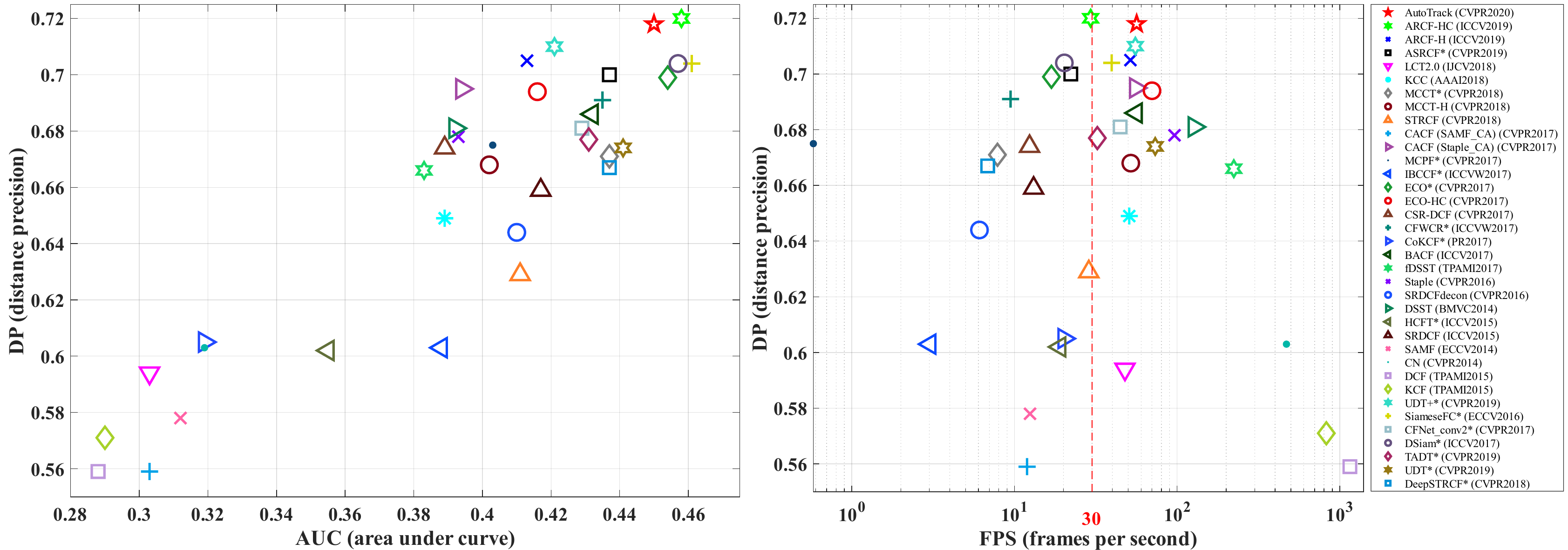}
	\vspace{-0.5cm}
	\caption{Comparison of the performance of DCF-based and deep trackers under UAVDT, the UAV tracking benchmark \cite{du2018ECCV}. The tracker with * in the legend is the result of running on GPU, which utilizes GPU to accelerate the convolution and pooling calculations. When the tracking speed reaches the red dotted line (30FPS) on a single CPU, it meets the requirements of UAV real-time tracking. AUC is related to success rate and DP depends on precision, whose exhaustive explanation is carried out in Section \ref{Section 4}. For better display effect, please refer to the electronic version of this paper.}
	\vspace{-0.5cm}
	\label{fig:Star}
\end{figure*}

Another competitive branch in the discriminative method is based on multiple instance learning (MIL) \cite{Babenko2009CVPR,babenko2011TPAMI,zhang2013PR,FU2019IS}. In \cite{Babenko2009CVPR}, B. Babenko \emph{et al.} showed that compared with traditional supervised learning methods, MIL can enable a more robust tracker with even fewer parameter tweaks. However, a serious problem in MIL is the instability of the sample labels. In other words, if there is a slight change in the training set (which is frequent in UAV tracking situation), the output sample labels can undergo drastic changes, ending up in poor robustness.

It is worth mentioning that among the previous trackers, K. Zhang \emph{et al.} proposed CT \cite{zhang2012ECCV} which employed a simple yet efficient appearance method and used a Bayes classifier in the compressed domain for tracking. The speed of CT is appreciable, nevertheless, its robustness and accuracy gradually fell behind the ranks.

With the development of the convolutional neural network (CNN) in recent years, applying a CNN for object tracking has become a research hot spot in recent years \cite{Li2018PR}, which we call deep trackers. Specifically, such methods usually train a CNN especially for object tracking, using a large number of labeled images offline.

While tracking the object, for one type of deep trackers, the object template and search region are input into the network simultaneously, and the object location and size in the search region are predicted directly end-to-end, \emph{e.g.}, \cite{Held2016ECCV,Li2018RPN,Danelljan2019atom,Bertinetto2016ECCV,Guo2017ICCV,Yan2019ICCV,Teng2020TIP}. To name a few, D. Held \emph{et al.} \cite{Held2016ECCV} used a simple feed-forward network structure which was amazingly efficient, but still hard to achieve real-time requirements on the UAV platform. Apart from the past work, L. Bertinetto \emph{et al.} novelly introduced fully-convolutional Siamese network (SiamFC) \cite{Bertinetto2016ECCV}. The network is completely trained end-to-end offline, avoiding online update of its parameter. Constructed on the general framework of SiamFC \cite{Bertinetto2016ECCV}, Q. Guo \textit{et al.} proposed dynamic Siamese network (DSiam) \cite{Guo2017ICCV}, where the fast transformation learning model can handle object appearance variation efficiently. Different from previous anchor-free fashion\cite{Bertinetto2016ECCV,Guo2017ICCV}, B. Li \textit{et al.} proposed Siamese region proposal network (SiamRPN) \cite{Li2018RPN}, where Siamese network is combined with the region proposal network (RPN) proposed in Fast R-CNN object detector \cite{girshick2015RCNN}. In this earliest anchor-based Siamese network, the traditional
multi-scale testing and online fine tuning can be abandoned, which greatly improves its speed and settles aspect ratio change issue. It's worth mentioning that M. Danelljan \textit{et al.} creatively proposed ATOM \cite{Danelljan2019atom}, where target tracking is divided into two stages: classification and estimation, different from the Siamese series methods. The first stage distinguishes the object from its background for rough localization. Aimed at fine bounding box estimation, the second stage creatively utilizes the IoU-net \cite{Jiang2018IoUnet}, which is trained offline with large-scale datasets to maximize the intersection over union (IoU) of ground-truth. Among them, some brilliant trackers are designed especially for long-term tracking. Recently, B. Yan \textit{et al.} \cite{Yan2019ICCV} proposed a novel tracker based on skimming and perusal modules. The innovative perusal module estimates target state, enabling the tracker to determine if the object disappears or not, thus deciding whether to global search or local search. In \cite{Dai2020CVPR}, K. Dai \textit{et al.} creatively off-line trained a meta-updater, which learns binary outputs to inform the tracker whether to update or not, greatly settling updating issue in long-term tracking. The other type utilizes CNN to extract deep feature of the object for model training and object detection, \emph{e.g.}, \cite{he2017CVPR,zhang2017CVPR,Li2018CVPR,Ma2015ICCV,Sun2019CVPR}.

%To name a few, D. Held \emph{et al.} \cite{Held2016ECCV} used a simple feed-forward network structure which was amazingly efficient, but still far from real-time requirements on the UAV platform. L. Bertinetto \emph{et al.} \cite{Bertinetto2016ECCV} novelly introduced fully-convolutional Siamese network (SiamFC). The network is completely trained end-to-end offline, avoiding online update of the its parameter. Q. Guo \textit{et al.} proposed dynamic Siamese network (DSiam) \cite{Guo2017ICCV}, where the fast transformation learning model can handle object appearance variation efficiently. B. Li \textit{et al.} proposed Siamese region proposal network (SiamRPN) \cite{Li2018RPN}, where Siamese network is combined with the region proposal network (RPN) proposed in Fast R-CNN object detector \cite{girshick2015RCNN}. As the classic anchor-based Siamese network, the traditional multi-scale testing and online fine tuning can be abandoned, which greatly improves its speed and settles aspect ratio change issue. M. Danelljan \textit{et al.} proposed ATOM \cite{Danelljan2019atom}, where target tracking is divided into two stages: classification and estimation. The first stage can distinguish the object from its background to realize rough localization. Aimed at fine bounding box estimation, the second stage creatively utilizes the IoU-net \cite{Jiang2018IoUnet}, which is trained offline with large-scale datasets to maximize the intersection over union (IoU) of ground-truth.

Although the brilliant deep trackers acquire promising results recently, they are generally implemented on high-performance GPU due to the high complexity of convolutional operation in the networks, which is unable to be supported onboard UAV. Besides, the offline training process requires a large number of pretreated UAV tracking images with annotations, which is hard to obtain. Moreover, the deep network is easy to lose efficacy faced with imperceptible noises \cite{Yan2020CVPR}.  Therefore, it is not an ideal approach for UAV-based aerial tracking.

Among various trackers, DCF-based trackers \cite{Henriques2015TPAMI,Zhao2019IS,Danelljan2015ICCV,Li2018CVPR,Danelljan2016ECCV,Dai2019CVPR} stand out due to their efficiency and accuracy, making them the suitable ones for UAV tracking. Compared with end-to-end methods, DCF-based methods \cite{Fu2020TGRS,Li2020CVPR,Lin2020ICRA,Huang2019ICCV,Fu2018ROBIO,Li2020IROS,Ding2020IROS} are more applicable on UAV platform tracking due to their high computational efficiency. Figure~\ref{fig:Star} showed DCF-based trackers' performances against deep trackers in terms of success rate, precision, and tracking speed with UAVDT \cite{du2018ECCV} benchmark. The next subsection presents the core idea, major steps, and the basic framework of DCF-based approaches.

\subsection{DCF-based trackers}

\subsubsection{Core idea}
The core idea of DCF-based trackers is to train a filter with the ability of classifying and scoring the search samples by minimizing the loss between the labels and the cyclic correlation between samples and filters. Classification results of the filter on the search samples can be obtained using the following formula:
\begin{equation}
\mathbf{g} = \mathbf{w} \star \mathbf{x} ~,
\end{equation}
where $\star$ symbolizes the cyclic correlation operator and $\mathbf{g}$ is supposed to be cyclic correlation between signals, \emph{e.g.}, in DCF-based tracking methods, $\mathbf{w}$ denotes the filter, $\mathbf{x}$ indicates the search samples, and $\mathbf{g}$ is the response map. To speed up the cyclic convolution calculation, cyclic correlation is computed in the Fourier domain using discrete Fourier transform (DFT), which can be expressed as: $\hat{\mathbf{g}}=\mathcal{F}(\mathbf{g})$, $\hat{\mathbf{w}}=\mathcal{F}(\mathbf{w})$, $\hat{\mathbf{x}}=\mathcal{F}(\mathbf{x})$, and the correlation calculation can be turned into:
\begin{equation}
\hat{\mathbf{g}} =\hat{\mathbf{w}}^* \odot \hat{\mathbf{x}}~,
\end{equation}
\noindent where $\odot$ denotes element-wise multiplication, and superscript $\cdot^*$ indicates the complex conjugate. As a common highlight of all DCF-based trackers, by substituting the operation with an efficient element-wise multiplication, the computational complexity can be substantially reduced as a result.

\subsubsection{Gerneral structure and major steps}
\begin{figure*}[!t]
	\includegraphics[width=2.05\columnwidth]{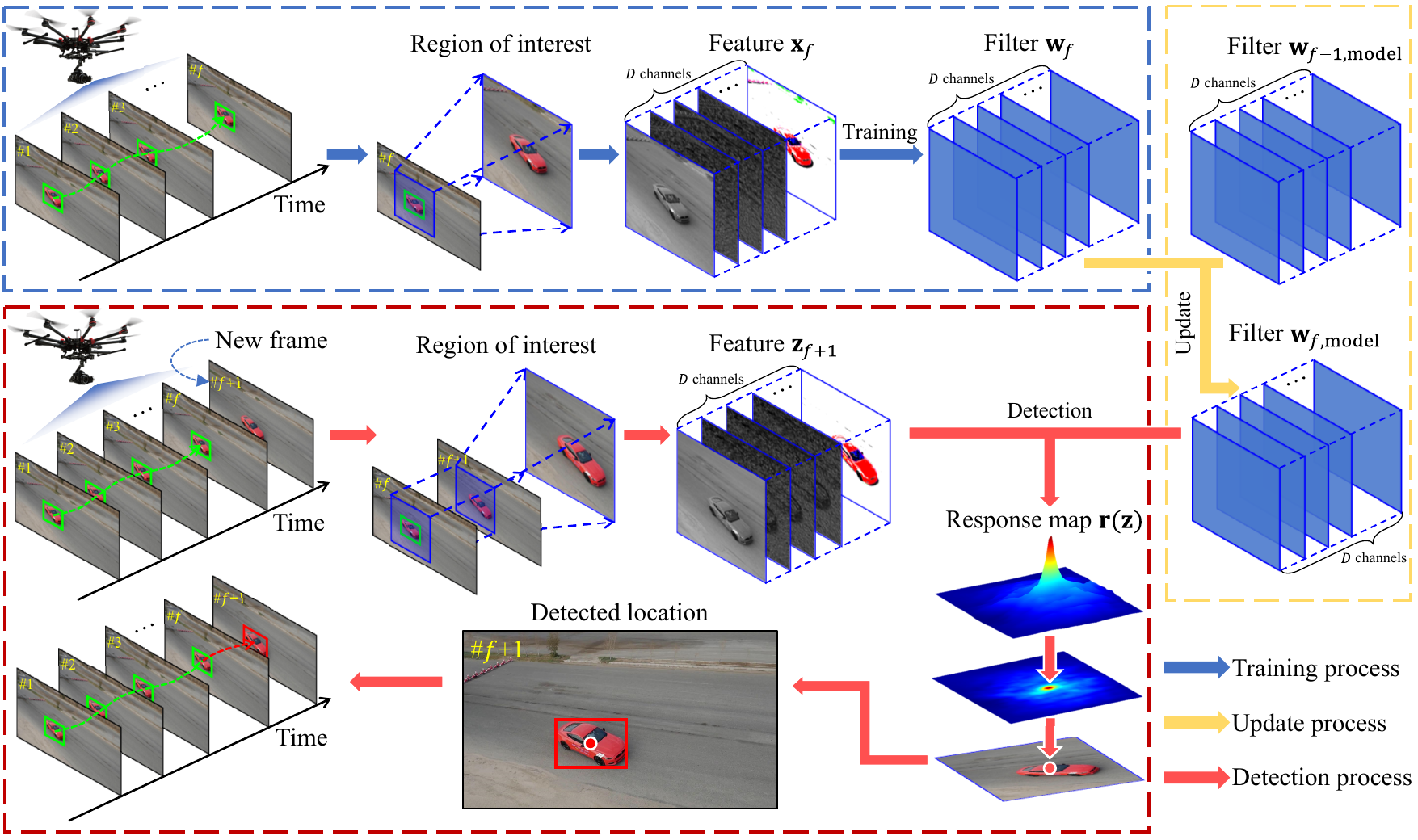}
	\vspace{-0.5cm}
	\caption{General tracking structure of DCF-based methods onboard the UAV platform, which can be divided into the training stage, model update, and detection stage. For better display effect, please refer to the electronic version of this paper.}
	\vspace{-0.5cm}
	\label{fig:main}
\end{figure*}

In general, nearly all the DCF-based trackers follow a similar structure, which mainly includes three steps: training stage, model update, and detection stage. As shown in Fig. \ref{fig:main}, in training stage, the training image patch is firstly cropped near current object center. Second, the feature of the image patch is extracted. Then, the extracted feature $\mathbf{x}_f$ is used as training samples to get the filter in the $f$-th frame $\mathbf{w}_f$ by solving the regression equation. Having gone through model update step, the filter model $\mathbf{w}_{f,\rm model}$ is obtained. During detection stage, in the $(f+1)$-th frame, the tracker firstly crops the region of interest (ROI), which is centered at current location. Then, the response map is generated by calculating the cyclic correlation between the filter model $\mathbf{w}_{f,\rm model}$ and feature of ROI $\mathbf{z}_{f+1}$ in the frequency domain, and the new location of the object in the $(f+1)$-th frame is detected according to the peak value of the response map. When the new location of the object is determined, the tracker extracts samples from the new location as new training samples. Thus in the following frames, training stage, model update, and detection stage are carried out in order.

The main differences in various filters generally lie in the three steps above, which are introduced in Section \ref{Section 3} comprehensively. In particular, the feature extraction strategy is the universal key component for all correlation filters, which can be generally divided into two camps: handcrafted feature and deep feature. The commonly used handcrafted features include grayscale, histograms of the gradient (HOG) \cite{Dalal2005CVPR}, a fast version of HOG (fHOG) \cite{Felzenszwalb2010TPAMI}, color names (CN) \cite{Weijer2006ECCV}, \emph{etc.}, which are not only easy to acquire but also robust to appearance changes. Different from handcrafted features, deep feature (used in deep trackers) is extracted from a multi-layer CNN, \emph{e.g.}, the VGGNet \cite{simonyan2014ICLR}. Deep features are usually more discriminative than those of handcrafted while bringing too much computation burden to the UAV platform as well.

\subsubsection{Basic framework}
Based on the structure above, D. S. Bolme \emph{et al.} \cite{Bolme2010CVPR} firstly proposed to train a filter $\mathbf{w}$ that minimizes the sum of squared error between preset labels and the correlation between samples and filters. Since the MOSSE tracker in \cite{Bolme2010CVPR} showed a high FPS rate in tracking performance with simple structure, plenty of DCF-based trackers with improvements based on \cite{Bolme2010CVPR} have come to the fore, whose tracking strategy can be summarized as follows.

\vspace{4mm}
(1) \textbf{Training stage}

Focusing on ridge regression, one of the goals of the DCF-based tracker is to train a filter in the $f$-th frame $\mathbf{w}_f$ that minimizes the squared error $\mathcal{E}$ of the correlation response in training samples $\mathbf{x}_f\in\mathbb{R}^{N\times D}$, which is the feature extracted from the image patch $\mathbf{O}^N\in\mathbb{R}^{N}$, compared to their regression targets $\mathbf{y}\in\mathbb{R}^{N}$, \emph{i.e.}:
\begin{equation}\label{eqn:3}
\mathcal{E}(\mathbf{w}_f)= \left\|\sum_{c=1}^{D}\mathbf{w}^c_f\star\mathbf{x}^c_f- \mathbf{y}\right\|^2_2 + \lambda \sum_{c=1}^{D}\left\|\mathbf{w}^c_f\right\|^2_2 ~,
\end{equation}
where $\mathbf{w}^c_f\in\mathbb{R}^{N}$ and $\mathbf{x}^c_f\in\mathbb{R}^{N}$ respectively denotes filter and training samples in the $c$-th feature channel, who totally contain $D$ channels. $\lambda$ is a regularization parameter that controls over-fitting.

\Remark For the convenience of derivation, this work considers the training samples and filter of a certain feature channel as one-dimensional, \emph{i.e.}, $\mathbf{x}^c, \mathbf{w}^c\in\mathbb{R}^{N}$ in most cases. In the implemented code, where the samples and filter are two-dimensional matrices with length and width, the derived results can be generalized to two-dimensional.

For convenience, the following only introduces the calculation in the $c$-th channel. The minimizer has a closed-form resolution:
\begin{equation}\label{eqn:4}
\mathbf{w}^c_f=(\mathbf{X}_f^{\rm T}\mathbf{X}_f+\lambda\mathbf{I}_N)^{-1}\mathbf{X}_f^{\rm T}\mathbf{y}~,
\end{equation}
where $\mathbf{I}_N$ is the identical matrix and $\mathbf{X}_f$ denotes the data matrix. Moreover, $\mathbf{X}_f$ is a circulant matrix, \emph{i.e.}:
\begin{equation}\label{eqn:5}
	\mathbf{X}_f=\mathcal{C}(\mathbf{x}^c)=[\mathbf{x}^c_{f,1},\mathbf{x}^c_{f,2}, \cdots, \mathbf{x}^c_{f,N}]~,
\end{equation}
where all columns in $\mathbf{X}_f$ are actually from the same original training sample $\mathbf{x}^c_f$, using circulant shift matrix $\mathbf{P}\in\mathbb{R}^{N\times N}$:
\begin{equation}
\mathbf{P}^1=\begin{bmatrix}
0  &  0  & 0 & \cdots\ &1\\
1  &  0  & 0 & \cdots\ & 0\\
0  &  1  & 0 & \cdots\ & 0\\
\vdots   & \vdots & \vdots & \ddots  & \vdots  \\
0  &  0  & \cdots\ & 1 & 0\\
\end{bmatrix}~.
\end{equation}
Thus Eq.~(\ref{eqn:5}) can also be represented as:
\begin{equation}\label{eqn:7}
\mathbf{X}_f=[\mathbf{P}^0\mathbf{x}^c_f, \mathbf{P}^1\mathbf{x}^c_f, \mathbf{P}^2\mathbf{x}^c_f, \mathbf{P}^3\mathbf{x}^c_f, \cdots, \mathbf{P}^{N-1}\mathbf{x}^c_f]~.
\end{equation}
The circulant matrix $\mathbf{X}_f$ can be diagonalized by DFT matrix $\mathbf{F}_N$, which can be expressed as:
\begin{equation}\label{eqn:8}
\mathbf{X}_f = \mathbf{F}^{\rm H}_N {\rm diag}(\hat{\mathbf{x}}^c_f) \mathbf{F}_N~,
\end{equation}
where the superscript $\cdot^{\mathrm{H}}$ is the Hermitian transpose, \emph{i.e.}, $\mathbf{F}_{N}^{\rm H}=(\mathbf{F}_{N}^*)^{\rm T}$. $\mathbf{\hat{x}}^c_f$ denotes the DFT of the generating vector, \emph{i.e.}, $\mathbf{\hat x}^c_f=\mathcal{F}(\mathbf{x}^c_f)=\sqrt{N}\mathbf{F}_N\mathbf{x}^c_f$. Therefore, Eq.~(\ref{eqn:4}) can be transformed into a simpler solution:
\begin{equation}\label{eqn:9}
\mathbf{\hat{w}}^c_f=\frac{\mathbf{\hat{x}}^{c*}_f\odot \mathbf{\hat{y}}}{\sum_{k=1}^D{\mathbf{\hat{x}}^{k*}_f\odot \mathbf{\hat{x}}^k_f}+\lambda\mathbf{I}_N}~.
\end{equation}
\Remark The filter $\mathbf{w}^c$ and training samples $\mathbf{x}^c$ here are all in the $c$-th feature channel, and the calculations between each channel do not interfere with each other.

\vspace{4mm}
(2) \textbf{Model update}

Generally, to avoid over-fitting, the filter used for detection does not directly take the calculation result of each frame but is obtained by linear interpolation. When learning rate $\eta$ is introduced, most model updating strategy for the $f$-th frame of DCF-based trackers is to update the filter model $\mathbf{w}_{f-1,\rm model}$ using linear interpolation as follows:
\begin{equation}\label{eqn:10}
\hat{\mathbf{w}}_{f,\rm model} =(1-\eta)\hat{\mathbf{w}}_{f-1,\rm model}+\eta \hat{\mathbf{w}}_f~.
\end{equation}

\vspace{4mm}
(3) \textbf{Detection stage}

Based on the concept of cyclic correlation, the detection strategy of the DCF-based method can also be expressed as follows:
\begin{equation}\label{eqn:11}
\mathbf{r}(\mathbf{z})=\mathcal{F}^{-1}\left(\sum_{c=1}^{D}\hat{\mathbf{w}}^{c*}_{f,\rm model}\odot\hat{\mathbf{z}}^c_{f+1}\right)~,
\end{equation}
where $\mathbf{z}_{f+1}$ represents the feature of ROI in the $(f+1)$-th frame, thus $\mathbf{r}(\mathbf{z})$ denotes response map.

\Remark The change in the position of the peak of the response map relative to the center indicates the displacement of the object, which can be calculated to obtain the object location in the $(f+1)$-th frame.

\begin{table}[!t]
	\centering
	\normalsize
	\fontsize{4.2}{11.7}\selectfont
	\caption{List for the frequently used symbols with their meaning and dimensions. For expressing conciseness and clarity, the different meanings of some symbols are emphasized in the corresponding subsections.}
	\begin{tabular}{ccc}
		\toprule
		\textbf{Symbol}	& \textbf{Meaning}  & \textbf{Dimension}  \\
		\midrule
		$\mathbf{w}^c$ & Correlation filter in the $c$-th channel&  $\mathbb{R}^{N}$  \\
		$\mathbf{w}$ & Correlation filter, $\mathbf{w}=[\mathbf{w}^1,\mathbf{w}^2,\cdots,\mathbf{w}^D]$&  $\mathbb{R}^{N\times D}$  \\
		$\mathcal{E}$ & Squared error &$\mathbb{R}$\\
		$\mathbf{O}^N$ & Original image patch without feature extraction & $\mathbb{R}^{N}$  \\
		$\mathbf{x}^c$& Training image samples in the $c$-th channel& $\mathbb{R}^{N}$  \\
		$\mathbf{x}$ & Training samples, $\mathbf{x}=[\mathbf{x}^1,\mathbf{x}^2,\cdots,\mathbf{x}^c,\cdots,\mathbf{x}^D]$ & $\mathbb{R}^{N\times D}$  \\
		$D$ & Total number of the feature channels & $\mathbb{R}$\\
		$x^c_i$ &Elements in training samples $\mathbf{x}^c=[x^c_1,x^c_2,\cdots,x^c_i,\cdots,x^c_N]^{\rm T}$ & $\mathbb{R}$\\
		$\mathbf{F}_N$ & DFT matrix & $\mathbb{C}^{N\times N}$  \\
		$\hat{\mathbf{x}}$&Training samples in frequency domain & $\mathbb{C}^{N\times D}$\\
		$\mathbf{y}$ & Gaussian shaped regression labels & $\mathbb{R}^{N}$\\
		$\mathcal{C}(\mathbf{x}^c)$ & Circulant data matrix for $\mathbf{x}^c$ & $\mathbb{R}^{N\times N}$\\
		$\mathbf{P}$ & Circulant shift matrix & $\mathbb{R}^{N\times N}$\\
		$\mathbf{C}$ & Cropping matrix & $\mathbb{R}^{N\times L}$\\
		$\mathbf{I}_N$ & Identity matrix & $\mathbb{R}^{N\times N}$\\
		$\mathbf{M}$ & Response map&$\mathbb{R}^N$\\
		$ \bm{\alpha}^c $& Coordinate vector in the $c$-th channel in dual space & $\mathbb{R}^N$\\
		$ \mathbf{k} $ & Kernel correlation vector & $\mathbb{R}^N $\\
		$\mathbf{z}$ & Samples the in search region (for single scale, $K$=1) & $\mathbb{R}^{N\times D\times K}$\\
		$\mathbf{z}^{s_i}$ & Samples corresponding to scale $s_i$ in search region & $\mathbb{R}^{N\times D}$\\
		$\mathbf{r}(\mathbf{z})$ & Response map generated by the search region samples& $\mathbb{C}^N $\\
		$\mathbf{x}_f$ & Training samples extracted in the $f$-th frame& $\mathbb{R}^{N}$\\
		$\bm{\alpha}_f$ & Filter parameters learned in the $f$-th frame& $\mathbb{R}^{N}$\\
		$\mathbf{x}_{f,\rm model}$ & Training template updated after the $f$-th frame& $\mathbb{R}^{N}$\\
		$\bm{\alpha}_{f,\rm model}$ & Filter parameters updated after the $f$-th frame& $\mathbb{R}^{N}$\\
		$T$ & Total number of the frames involved in regression equation & $\mathbb{R}$\\
		$\eta$ & Learning rate & $\mathbb{R}$\\
		$\mathbf{S}$ & Scaling pool containing K sizes & $\mathbb{R}^{k}$\\
		$s_i$ & Scale rate in scaling pool & $\mathbb{R}$\\
		$K$ & Total number of possible scales & $\mathbb{R}$\\
		$\mathbf{s}_{\rm T}$ & Object template size & $\mathbb{R}^{2}$\\
		$w$ & Width of the object template &$\mathbb{R}$\\
		$h$ & Height of the object template &$\mathbb{R}$\\
		$\tau$ & Threshold &$\mathbb{R}$\\
		$\bm{\omega}$ & Spatial punishment weight & $\mathbb{R}^{N}$\\
		$\theta_f$ & Weight for training samples in the $f$-th frame&$\mathbb{R}$\\
		$\gamma$ & Interpolation weight for different response map & $\mathbb{R}$\\
		$\star$   & Cyclic correlation operator & -    \\
		$\mathcal{F}$ & Discrete Fourier transform& -\\
		$\mathcal{F}^{-1}$ & Inverse discrete Fourier transform& -\\
		$\cdot^*$ & Complex conjugate & - \\
		$\left\|\cdot\right\|_2$ & $L^2$ norm known as the Euclidean norm & -\\
		$|\cdot|$ & Number of elements in a matrix & -\\
		$\otimes$&Kronecker product&-\\
		$<\cdot,\cdot>$ &Dot product & -\\
		$\mathcal{D}(\cdot)$& A diagonal matrix with elements in vector $\cdot$& -\\
		$\odot$ & Element-wise multiplication & -\\
		$\cdot^{\rm H}$ & Hermitian transpose & -\\
		$\phi(\cdot)$ & Mapping function & -\\
		$\psi(\cdot)$ & Shifting operation & -\\
		$\varphi(\cdot)$ & Feature transform function & -\\
		$R$ & Different tracking model in LCT, Staple, \emph{etc.} & - \\
		\bottomrule
	\end{tabular}%
	\label{tab:1}%
\end{table}%

Arriving here, DCF-based trackers use Eq.~(\ref{eqn:9}) for filter training, Eq.~(\ref{eqn:10}) for model update, and Eq.~(\ref{eqn:11}) for object detection. Since most of the calculation in the equations is element-wise, DCF-based trackers reduce storage and computation by several orders of magnitude compared with the traditional solution of the ridge regression problem. The DCF-based approach brings the tracking algorithms to a new level, greatly promoted the robustness and accuracy with satisfying speed, thus becoming the mainstream method in the UAV tracking field.

On the basic framework, a series of methods have been developed to deal with various challenges. For example, M. Danelljan \emph{et al.} \cite{danelljan2017TPAMI} proposed a creative one that can solve the scale change issue faster. H. K. Galoogahi \emph{et al.} \cite{Galoogahi2017ICCV} made use of the negative samples generated by the real shift to include a larger search region and real background information and applied the Alternating Direction Method of Multipliers (ADMM) to solve the filter, \emph{etc}. Specially, recent years have witnessed a number of correlation filter trackers designed for real-time UAV tracking scenes, \emph{e.g.}, \cite{Fu2020TGRS,Fu2020NCA,Li2020CVPR,Lin2020ICRA,Huang2019ICCV,Fu2018ROBIO}, which process not only excellent performance but also wide recognition. This kind of method can reduce the calculation load on UAV, thus reducing the power consumption to extend the valuable UAV's endurance time. The remaining computational resources can be put into the high-level control algorithms multi-sensor information fusion, path planning, \emph{etc.} Therefore, these advantages have made it have great advances on the platform of UAV, and finally enhances the overall performance of UAV.

\subsection{CF onboard UAV}
Generally speaking, there are three reasons why DCF-based tracker outperforms most of other tracking methods on UAV: 
\begin{itemize}
	\item \emph{Adaptability.} DCF is an online learning method. As is mentioned in the previous subsection, the tracking model is usually updated once every frame, which enables the tracker to respond to the changes of object appearance and scale in time. In UAV tracking, due to the frequent changes of view angle, height, and distance, the online updating and training ability of DCF-based tracker ensure their adaptability to object appearance changes, which has become one of its vital competitiveness onboard UAV. 
	\item \emph{Robustness.} The DCF-based method belongs to the discriminative method. It not only learns the object information but also the background information. The high discriminability of the filter enables the UAV to maintain high tracking robustness even when it encounters severe environmental changes, similar object interference, and other adverse conditions in the tracking process. 
	\item \emph{Efficiency.} Most of the operations involved in DCF are element-wise products in the frequency domain, which has an impressive running speed compared with most other tracking algorithms. The high speed of the DCF tracker not only makes the UAV realize the real-time tracking function on a single CPU but also saves the power for the UAV. Thus, the redundant computing power can be used to handle other processes, to broaden the use scenarios of the UAV.
\end{itemize}

Based on these priority mentioned above, the researches in latest years have further improved the performance of correlation filter, which raised the application of CF method in UAV to an even higher level \cite{Fu2020TGRS,Fu2020NCA,Li2020CVPR,Lin2020ICRA,Li2020ICRA,Huang2019ICCV,Fu2018ROBIO}. Specifically, Y. Li \emph{et al.} \cite{Li2020CVPR} proposed a novel approach to online automatically and adaptively adjust the spatio-temporal regularization term (AutoTrack), thus greatly reducing the workload of tuning predefined parameters. Z. Huang \emph{et al.} \cite{Huang2019ICCV} creatively trained a filter that can learn to repress the aberrances emerging in the detection stage (ARCF), resulting in its favorable robustness. C. Fu \emph{et al.} \cite{Fu2020TGRS} applied saliency detection in the filter training process and utilized dual regularization (DRCF), thus accentuating the object appearance and achieving a promising result. F. Lin \emph{et al.} \cite{Lin2020ICRA} put forward a novel bidirectional incongruity-aware correlation filter (BiCF), which is not only able to track the object forward but also able to locate the object in the previous frame, showing its advantage in both adaptability and robustness. F. Li \emph{et al.} \cite{Li2020ICRA} focused on the training samples, which proposed an original time slot-based distillation approach to optimize the training samples' quality.

As aforementioned, as an efficient and robust discriminative object tracking strategy, DCF-based methods have outstanding performance and become the mainstream method in the UAV tracking community. In the next section, the contribution of various DCF-based methods in detail are summarized.

\section{Development of DCF-based Methods}\label{Section 3}

\begin{figure*}[!t]
	\includegraphics[width=2.05\columnwidth]{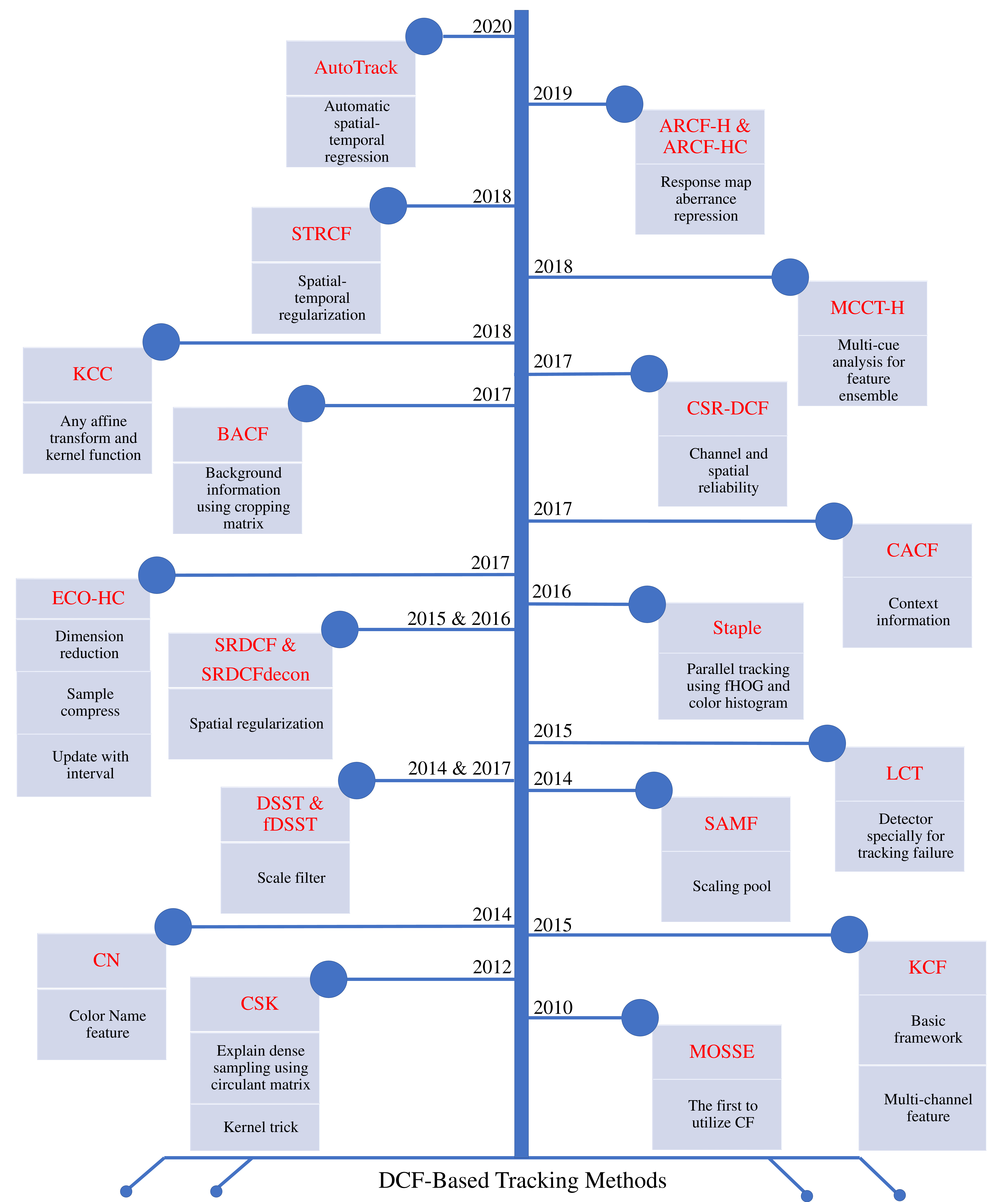}
	\vspace{-0.5cm}
	\caption{The development of DCF-based tracking methods over the years. Here focuses on their most surprising contributions, which are all introduced in detail in Section \ref{Section 3}. The words in \textcolor{red}{red} denote the trackers' names, and the black words briefly summarize their innovations. For better display effect, please refer to the electronic version of this paper.}
	\vspace{-0.5cm}
	\label{fig:tree}
\end{figure*}

Even though the general structure of most DCF-based trackers is the same, every tracker possesses its special priority and contribution. This section focuses on 23 DCF-based trackers' creative contribution in detail (some are in the same subsection, \emph{i.e.}, the fDSST tracker and DSST tracker). In general, as is shown in Fig.~\ref{fig:tree}, over the years, different trackers proposed their innovation focusing on different issues and achieved better and better results. Generally, the trackers introduced in this section are categorized into foundation, scale estimation, feature representation, boundary effect, temporal consistency, and extra types. Table~\ref{tab:1} lists most symbols used in this work for convenient reference. Table~\ref{tab:2} shows all the state-of-the-art DCF-based trackers in their categories, including their venue, features used, and other characteristics, in corresponding categories.

\Remark Considered their implement ability onboard UAVs, only trackers using handcrafted features are considered practical, which ensure promising speed on a single CPU and save the power supply onboard UAVs. Some trackers adopting CF structure but CNN features, \emph{e.g.}, the HCFT tracker \cite{Ma2015ICCV}, are classified as deep trackers, which are not the mainstream in this work. 

\begin{table*}[htbp]
	\centering
	\setlength{\tabcolsep}{2mm}
	\fontsize{6.5}{12}\selectfont
	\caption{DCF-based trackers that are suitable for UAV tracking and their characteristics. Note that scale estimation methods are introduced in detail in Section \ref{Section 3}, \emph{e.g.}, scaling pool is proposed by the SAMF tracker \cite{li2014ECCV}, scale filter is put forward by the DSST tracker \cite{danelljan2014BMVC}, and kernel scale correlator is introduced in the KCC tracker \cite{wang2018AAAI}. The features and the optimization schemes refer to the corresponding paper and code library. The typical state-of-the-art trackers listed in this table have been evaluated in the experimental section.}
	\vspace{0.08cm}
	\begin{tabular}{cccccc}
		\toprule[1.5pt]
		Category & Name  &  Venue & Features & Scale estimation & Solution \\
		\midrule
		\multirow{3}[2]{*}{Foundation} & MOSSE \cite{Bolme2010CVPR} & CVPR2010 &  Grayscale & Single scale & Analytical solution \\
		& CSK \cite{Henriques2012ECCV} & ECCV2012 & Grayscale & Single scale & Analytical solution \\
		& KCF \cite{Henriques2015TPAMI} & TPAMI2015 & fHOG  & Single scale & Analytical solution \\
		\midrule
		\multirow{3}[2]{*}{Scale estimation} & SAMF \cite{li2014ECCV} & ECCV2014 & fHOG+CN & Scaling pool & Analytical solution \\
		& DSST \cite{danelljan2014BMVC} & BMVC2014 & fHOG  & Scale filter & Analytical solution \\
		& fDSST \cite{danelljan2017TPAMI} & TPAMI2017 & fHOG+Grayscale & Scale filter & Analytical solution \\
		\midrule
		\multirow{3}[2]{*}{Feature representation} & CN \cite{Danelljan2014CVPR} & CVPR2014 & Grayscale + CN & Single scale & Analytical solution \\
		& Staple \cite{bertinetto2016CVPR} & CVPR2016 & fHOG+Color historgram & Scale filter & Analytical solution \\
		& MCCT-H \cite{wang2018CVPR} & CVPR2018 & fHOG+CN+Color historgram & Scale filter & Analytical solution \\
		\midrule
		\multirow{4}[2]{*}{Boundary effect} & SRDCF \cite{Danelljan2015ICCV} & ICCV2015 & fHOG  & Scaling pool & Numerical solution (Gauss-Seidel iteration) \\
		& SRDCFdecon \cite{danelljan2016CVPR} & CVPR2016 & fHOG  & Scaling pool & Numerical solution (Gauss-Seidel iteration) \\
		& CSR-DCF \cite{lukezic2017CVPR} & CVPR2017 & fHOG+CN+Grayscale & Scale filter & Numerical solution (ADMM iteration) \\
		& BACF \cite{kiani2017ICCV} & ICCV2017 & fHOG  & Scaling pool & Numerical solution (ADMM iteration) \\
		\midrule
		\multirow{4}[2]{*}{Temporal consistency} & STRCF \cite{Li2018CVPR} & CVPR2018 & fHOG+CN+Grayscale & Scaling pool & Numerical solution (ADMM iteration) \\
		& ARCF-H \cite{Huang2019ICCV} & ICCV2019 & fHOG  & Scale filter & Numerical solution (ADMM iteration) \\
		& ARCF-HC \cite{Huang2019ICCV} & ICCV2019 & fHOG+CN+Grayscale & Scale filter & Numerical solution (ADMM iteration) \\
		& AutoTrack \cite{Li2020CVPR} & CVPR2020 & fHOG+CN+Grayscale & Scale filter & Numerical solution (ADMM iteration) \\
		\midrule
		\multirow{4}[2]{*}{Extra types} & LCT \cite{ma2015CVPR} & CVPR2015 & fHOG  & Scale filter & Analytical solution \\
		& LCT2.0 \cite{ma2018IJCV} & IJCV2018 & fHOG+Intensity histogram  & Scale filter & Analytical solution \\
		& ECO-HC \cite{danelljan2017CVPR} & CVPR2017 & fHOG+CN & Scale filter & Numerical solution (Conjugate gradient) \\
		& SAMF\_CA \cite{mueller2017CVPR} & CVPR2017 & fHOG+CN & Scaling pool & Analytical solution \\
		& Staple\_CA \cite{mueller2017CVPR} & CVPR2017 & fHOG+CN & Scaling pool & Analytical solution \\
		& KCC \cite{wang2018AAAI} & AAAI2018 & fHOG+Color historgram & Kernel scale correlator & Analytical solution \\
		\bottomrule[1.5pt]
		\vspace{-0.3cm}
	\end{tabular}%
	\label{tab:2}%
\end{table*}%

\subsection{Foundation}

The MOSSE tracker \cite{Bolme2010CVPR}, the CSK \cite{Henriques2012ECCV}, and the KCF tracker \cite{Henriques2015TPAMI}, which are considered as the cornerstone of DCF-based methods, built the basic framework and proposed the key idea of DCF.
\subsubsection{MOSSE tracker}
Considered the first tracker that utilized CF, the MOSSE tracker \cite{Bolme2010CVPR}, proposed by 
D. S. Bolme \emph{et al.}, aimed to train a filter that minimizes the squared error between the output of cyclic correlation and the designed labels using the following regression equation:
\begin{equation}\label{eqn:12}
\mathcal{E}(\mathbf{w})=\left\|\mathbf{w}\star\mathbf{x}-\mathbf{y}\right\|^2_2~.
\end{equation}

Note that the filter $\mathbf{w}$ and training samples $\mathbf{x}$ are all in single feature channel here, \emph{i.e.}, $\mathbf{w}=\mathbf{w}^1,\mathbf{x}=\mathbf{x}^1$.

Using partial derivative in frequency domain, a closed-form solution for Eq.~(\ref{eqn:12}) can be obtained for the MOSSE tracker:
\begin{equation}\label{eqn:13}
\mathbf{\hat{w}}=\frac{\mathbf{\hat{x}}^*\odot \mathbf{\hat{y}}}{\mathbf{\hat{x}}^*\odot \mathbf{\hat{x}}}~.
\end{equation}

Note that during the model update stage, the MOSSE tracker uses the following updating scheme for more robust estimation:
\begin{align}
\mathbf{A}_{f, \rm model}&=\eta \mathbf{A}_f+(1-\eta)\mathbf{A}_{f-1, \rm model}~,\\
\mathbf{B}_{f, \rm model}&=\eta \mathbf{B}_f+(1-\eta)\mathbf{B}_{f-1, \rm model}~,
\end{align}
where $\mathbf{A}=\mathbf{\hat{x}}^*\odot \mathbf{\hat{y}}$ and $\mathbf{B}=\mathbf{\hat{x}}^*\odot \mathbf{\hat{x}}$ respectively indicates the numerator and the denominator in Eq.~(\ref{eqn:13}). Therefore, the filter model obtained after the $f$-th frame is $\hat{\mathbf{ w}}_{f,\rm model}=\frac{\mathbf{A}_{f,\rm model}}{\mathbf{B}_{f,\rm model}}$.

Having calculated the filter model $\mathbf{w}_{\rm model}$, the MOSSE tracker adopts Eq.~(\ref{eqn:11}) for detection in new frames. 

Due to its simplicity, MOSSE algorithm achieved compelling tracking speed of hundreds of frames per second. 

Large-scale variation remains hard for the MOSSE tracker to adapt to, and its single-channel grayscale feature are not powerful enough. Besides, MOSSE tracker does not consider the linear separability of the samples in high-dimensional spaces.

\subsubsection{CSK tracker}
Previous outstanding trackers, such as the SVM tracker \cite{Avidan2004TPAMI}, used sparse sampling to obtain training samples. To be specific, several candidate boxes with the same size as the object is randomly generated near the current object center in the search region. This method cannot extract sufficient samples on the one hand, and on the other overlap among the candidate, boxes cause inefficiency. The CSK \cite{Henriques2012ECCV} tracker explained for the first time that the cyclic correlation $\star$ used in CF is a kind of dense sampling in a mathematical sense, and introduced circulant matrix to calculate the cyclic correlation. Such a strategy firstly extracts the single channel feature of an image patch as a basic sample $\mathbf{x}=[\mathbf{x}^1]\in\mathbb{R}^N$, and cyclically shifting it using Eq.~(\ref{eqn:5}) to obtain the circulant data matrix Eq.~(\ref{eqn:7}) as training samples.

In addition to dense sampling and circulant matrix, the CSK tracker also improves the regression equation in the MOSSE tracker. The improved equation can be expressed as the single-channel form of Eq.~(\ref{eqn:3}), which is known as regularized least squares (RLS). The proposed regularization term can prevent the filter $\mathbf{w}$ from overfitting, which is proved to have gained much better results compared to the MOSSE tracker. 

Apart from that, another highlight of the CSK algorithm is its application of kernel trick, which maps the classification procedure to high dimensional feature space to acquire even better performance. Using kernel trick, $\mathbf{w}$ can be written as: $\mathbf{w}=\sum_{i}\alpha_i\phi(\mathbf{x}_i),$ where the mapping function $\phi(\cdot)$ can map the data to the high dimensional feature space, and sample $\mathbf{x}_i$ is from the original sample $\mathbf{x}$ using circulant shift matrix $\mathbf{P}$, \emph{e.g.}, $\mathbf{x}_1=\mathbf{P}^0\mathbf{x}$. The RLS with kernels (KRLS) has a simple closed-form solution:
\begin{equation}\label{eqn:18}
\bm{\alpha}=(\mathbf{K}+\lambda\mathbf{I}_N)^{-1}\mathbf{y}~,
\end{equation}
where $\mathbf{K}$ is the kernel matrix with elements $K_{ij}=k(\mathbf{x}_i,\mathbf{x}_j)$. Kernel $k$ is defined as:
\begin{equation}
k(\mathbf{x}_i,\mathbf{x}_j)=<\phi(\mathbf{x}_i),\phi(\mathbf{x}_j)>~,
\end{equation}
where $<\cdot,\cdot>$ denotes dot product.

\Remark Although J. F. Henriques \emph{et al.} achieved amazing tracking results using the CSK tracker, the algorithm still uses illumination intensity features, which is inferior robust. 

\subsubsection{KCF tracker}
Standing as the basic framework of most subsequent DCF-based trackers, the main contribution of the KCF tracker is that J. F. Henriques \emph{et al.} \cite{Henriques2015TPAMI} formulated their prior work of the CSK tracker \cite{Henriques2012ECCV}, integrating its core ideas into the CF tracking algorithm.

Besides, the KCF tracker handled multi-channel feature representation problem, \emph{e.g.}, fHOG \cite{Felzenszwalb2010TPAMI}, and greatly boosted the tracker's performance.

Firstly, KCF algorithm proposed the non-linear regression equation for filter training:
\begin{equation}\label{eqn:21}
\mathcal{E}(\mathbf{w})=\left\|\sum_{c=1}^{D}\mathbf{w}^c\star \phi(\mathbf{x}^c)-\mathbf{y}\right\|^2_2+\lambda\sum_{c=1}^{D}\left\|\mathbf{w}^c\right\|^2_2~.
\end{equation}

Assuming that the training sample is single feature channel, \emph{i.e.}, $\mathbf{x}\in\mathbb{R}^{N\times 1}$. By adopting the kernel tricks $\mathbf{w}=\sum_{i}\alpha_i\phi(\mathbf{x}_i)$, their solution to the filter parameter $\bm{\alpha}$ was first obtained the same as the CSK tracker, \emph{i.e.}, Eq.~(\ref{eqn:18}). Proving that the kernel matrix $\mathbf{K}$ is circulant, a faster version of Eq.~(\ref{eqn:18}) is proposed by:
\begin{equation}\label{eqn:22}
\hat{\bm {\alpha}}=\frac{\hat{\mathbf{y}}}{\hat{\mathbf{k}}^{\mathbf{x}\mathbf{x}}+\lambda\mathbf{I}_N}~,
\end{equation}
where $\mathbf{k}^{\mathbf{x}\mathbf{x}}$ denotes kernel correlation vector. For two arbitrary vectors, $\mathbf{x}$ and $\mathbf{x}'$, their kernel correlation is the vector $\mathbf{k}^{\mathbf{x}\mathbf{x}'}$ with elements:
\begin{equation}
k^{\mathbf{x}\mathbf{x}'}_i=\phi^{\rm T}(\mathbf{x}')\phi(\mathbf{P}^{i-1}\mathbf{x})~.
\end{equation}

The KCF tracker gave two commonly used kernel function:

(1) Gauss kernel:
\begin{equation}\label{eqn:24}
\mathbf{k}^{\mathbf{x}\mathbf{x}'}={\rm exp}\Big(-\frac{1}{\sigma^2}\big(\left\|\mathbf{x}\right\|^2_2+\left\|\mathbf{x'}\right\|^2_2-2\mathcal{F}^{-1}(\hat{\mathbf{x}}^*\odot\hat{\mathbf{x}}')\big)\Big)~,
\end{equation}

(2) polynomial kernel:
\begin{equation}\label{eqn:25}
\mathbf{k}^{\mathbf{x}\mathbf{x}'}=\Big(\mathcal{F}^{-1}(\hat{\mathbf{x}}^*\odot\hat{\mathbf{x}}')+a\Big)^b~.
\end{equation}

If the training sample $\mathbf{x}$ is multi-channel sample, $\mathbf{x}\in\mathbb{R}^{N\times D}$, \emph{e.g.}, fHOG \cite{Felzenszwalb2010TPAMI}, the multi-channel version of Eq.~(\ref{eqn:24}) and Eq.~(\ref{eqn:25}) are as follows: 

(1) Gauss kernel:
\begin{equation}
\mathbf{k}^{\mathbf{x}\mathbf{x}'}\!={\rm exp}\Big(-\frac{1}{\sigma^2}\big(\left\|\mathbf{x}\right\|^2_2+\left\|\mathbf{x'}\right\|^2_2-2\mathcal{F}^{-1}(\sum_c\hat{\mathbf{x}}^{*c}\odot\hat{\mathbf{x}}^{'c})\big)\Big)~,
\end{equation}

(2) polynomial kernel:
\begin{equation}
\mathbf{k}^{\mathbf{x}\mathbf{x}'}=\Big(\mathcal{F}^{-1}(\sum_c\hat{\mathbf{x}}^{*c}\odot\hat{\mathbf{x}}^{'c})+a\Big)^b~.
\end{equation}

For a linear kernel, its function can be expressed as:
\begin{equation}\label{eqn:28}
\mathbf{k}^{\mathbf{x}\mathbf{x}'}=\mathcal{F}^{-1}(\sum_c\hat{\mathbf{x}}^{*c}\odot\hat{\mathbf{x}}^{'c})~.
\end{equation}

In this case, it is a linear filter but trained in the dual space $\bm{\alpha}$.

For the model update, instead of updating filter $\mathbf{w}$, the KCF method updates the filter parameter and the sample:
\begin{equation}\label{eqn:30}
\begin{aligned}
\hat{\bm{\alpha}}_{f,\rm model} &=(1-\eta)\hat{\bm{\alpha}}_{f-1,\rm model}+\eta \hat{\bm{\alpha}}_{f}~.\\
\mathbf{x}_{f,\rm model}&=(1-\eta)\mathbf{x}_{f-1,\rm model}+\eta\mathbf{x}_f.
\end{aligned}
\end{equation}

Thus, the sample model $\mathbf{x}_{f,\rm model}$, instead of the original sample $\mathbf{x}_f$, is used for filter training in the $f$-th frame.

In the detection stage, the response map can be obtained using:
\begin{equation}\label{eqn:29}
\mathbf{r}(\mathbf{z})=\mathcal{F}^{-1}(\sum_{c}\hat{\mathbf{k}}^{\mathbf{x}\mathbf{z}}\odot\hat{\bm{\alpha}}_{\rm model})~,
\end{equation}
where $\mathbf{z}$ denotes samples in the search region.

The KCF tracker formulated the kernel tricks used in the CSK tracker into DCF-based tracking structure and put forward multi-channel samples' training strategy, thus becoming the basic framework for most DCF-based trackers.

\subsection{Scale estimation}
The three aforementioned fundamental trackers \cite{Bolme2010CVPR,Henriques2012ECCV,Henriques2015TPAMI} mentioned above can achieve target localization, but they adopted single scale during the tracking process, \textit{i.e.}, assuming that the scale of the object is fixed. Considering that the target scale changes in real-world tracking, how to accurately and effectively estimate the scale of the object has become an urgent problem to be settled.

\subsubsection{SAMF tracker}
The main creative contribution of the SAMF tracker is the proposal of an effective scale estimation method. 

In the SAMF algorithm, assuming that the original fixed size of the object template is $\mathbf{s}_{\rm T}=(w,h)$, where $w,h$ represents the width and height of the search region respectively. The scale pool can be defined as $\mathbf{S}=\{s_i\mathbf{s}_{\rm T}|s_i=\{s_1,s_2,...,s_k\}\}$. During detection stage, the image patches in $k$ different sizes in the scale pool are firstly cropped. Having extracted the feature of the resized image patches $\{\mathbf{z}^{s_1},\mathbf{z}^{s_2},\cdots,\mathbf{z}^{s_k}\}$, the SAMF tracker can determine the optimal scale $s_i$ by solving the following optimization problem:
\begin{equation}
\mathop{\arg\max_{s_i}}\ \mathbf{r}(\mathbf{z}^{s_i})~.
\end{equation}
Moreover, the location of the peak value can be used to estimate the new location of the object. Thus, the SAMF tracker achieves both location and scale prediction at the same time.

\Remark Taking into account that the new samples extracted in each frame have different scales, the SAMF tracker firstly converts the sample bilinear interpolation to the same size and then used the same linear interpolation template update strategy as the KCF tracker Eq (\ref{eqn:12}) and (\ref{eqn:30}). 

\subsubsection{DSST \& fDSST tracker}
The scaling pool algorithm given in the SAMF tracker \cite{li2014ECCV} can cope with scale variations, but its operation speed and robustness had room for further improvement. M. Danelljan \emph{et al.} made an innovative contribution to the scale estimation algorithm.

In \cite{danelljan2014BMVC}, the training regression equation adopted the ridge regression in the KCF tracker \cite{Henriques2015TPAMI}, \emph{i.e.}, Eq.~(\ref{eqn:21}), and applied linear kernel, \emph{i.e.}, Eq.~(\ref{eqn:28}). Their solution is:
\begin{equation}\label{eqn:19}
\mathbf{\hat{w}}^c=\frac{\mathbf{\hat{x}}^{c*}\odot \mathbf{\hat{y}}}{\sum_{k=1}^{D}\mathbf{\hat{x}}^{k*}\odot \mathbf{\hat{x}}^k+\lambda\mathbf{I}_N}~.
\end{equation}

Then, the core idea of the DSST tracker in \cite{danelljan2014BMVC} is to train two filters, namely a translation filter $\mathbf{w}_{\rm trans}$ and a scale filter $\mathbf{w}_{\rm scale}$. When a new frame arrives, the translation filter is first used to search for the new object location, and then samples of different scales $[\mathbf{z}^{s_1},\mathbf{z}^{s_2},\cdots,\mathbf{z}^{s_k}]$ are extracted near the new center. The samples in various scales are used for the scale filter to predict the proper scale. Then, the translation filter uses samples centered at the predicted location for training, while the scale filter utilizes samples in different scales that centered at the predicted scale for training. Such a scale evaluation method combines both robustness and speed.

\Remark In the implemented code, the translation filter adopts the promoted two-dimensional results of Eq.~(\ref{eqn:19}), where the training samples and filter in a certain feature channel are two-dimensional. Differently, the DSST tracker pulls the two-dimensional matrices into a one-dimensional vector to train the one-dimensional scale filter, adopting Eq.~(\ref{eqn:19}) directly. Such a strategy greatly boosts the processing speed of the DSST tracker.

Another reason why the DSST tracker achieves fast scale estimation is that the scale filter crops the patches about the size of the object, which is relatively small, making feature extraction more efficient.

Based on the DSST tracker, M. Danelljan further proposed a faster version called the fDSST tracker \cite{danelljan2017TPAMI}, which improves the tracker performance while obtaining even higher tracking speed. The strategy can be summarized into three points. First, the fDSST tracker uses sub-grid interpolation to reduce the size of training samples and search samples. Secondly, the fDSST tracker performs PCA on the features of the sample (similar to the principal component analysis of CN in the CN tracker \cite{Danelljan2014CVPR}) to achieve feature dimensionality reduction. Finally, the scale filter is compressed by reducing the total number of features to the number of features after PCA dimensionality reduction, which greatly cut down redundant information. Under the above three acceleration strategies, the fDSST tracker can expand its search region to obtain better tracking performance.

\subsection{Feature representation}
For the sake of better robustness and favorable discriminative ability of the trackers, it is of vital importance to learn sufficient and effective object features. The trackers introduced this part \cite{Danelljan2014CVPR,bertinetto2016CVPR,wang2018CVPR} put forward a variety of new methods to learn and effectively utilize the expressive features, thus promoting the DCF algorithms to a new level.

\subsubsection{CN tracker}
The CN tracker \cite{Danelljan2014CVPR}, which was proposed by M. Danelljan \emph{et al.}, put forward the innovative color feature based on the CSK tracker \cite{Henriques2012ECCV}, which is another powerful handcrafted feature. Besides, the CN tracker also settled the multi-channel training problem.

The color attribute \cite{Weijer2006ECCV}, also known as the color name (CN), directly denotes 11 color language labels defined by human beings. CN feature firstly maps the value of each channel in RGB image to the 11-dimensional color names probability, which sum up equals 1. Then, the 11-dimensional color space is mapped to the 10-dimensional orthogonal basis subspace, thus the dimension reduction and normalization are achieved at the same time (the CN tracker).

However, there is a linear relationship between the computational complexity of the CSK tracker and the number of dimensions of the feature used. To lessen the amount of calculation cost to guarantee high speed, the CN tracker further proposed low dimensional adaptive color attributes, where principal component analysis (PCA) is performed on 10-dimensional features to select the two dimensions with the most information (the analysis results of each frame are adaptive)(the CN\_2 tracker).

Expect for proposing the color feature, another improvement in the CN tracker \cite{Danelljan2014CVPR} is that it considers the information in the previous frames in their training progress, \emph{i.e.}:
\begin{equation}
\mathcal{E}(\mathbf{ w}_T)=\sum_{f=1}^T\theta_f\left(\left\|\sum_{c=1}^D\mathbf{w}^c_f\star\phi(\mathbf{x}^c_f)-\mathbf{y}\right\|^2_2+\lambda\sum_{c=1}^D\left\|\mathbf{w}_f\right\|^2_2\right),
\end{equation}
where $\cdot_f$ denotes values in the $f$-th frame, \emph{e.g.}, $\theta_f$ indicates the weight for the $f$-th frame. Assuming that the training sample is single feature channel, \emph{i.e.}, $\mathbf{x}\in\mathbb{R}^{N\times 1}$. Their solution for the filter parameter ${\bm \alpha}_T$ in the $T$-th frame is:
\begin{equation}
\hat{\bm {\alpha}}_T=\frac{\sum_{f=1}^T\theta_f\hat{\mathbf{y}}\odot\hat{\mathbf{k}}^{\mathbf{x}\mathbf{x}}_f}{\sum_{f=1}^T\theta_f\hat{\mathbf{k}}^{\mathbf{x}\mathbf{x}}_f\odot(\hat{\mathbf{k}}^{\mathbf{x}\mathbf{x}}_f+\lambda\mathbf{I}_N)}~.
\end{equation}
Such a strategy, where previous frames are taken into account, can improve the robustness of the tracker.

\Remark The CN feature not only improved the performance of the original CSK algorithm but also brought DCF-based trackers to a new level, making them more robust under challenging scenes.

\subsubsection{Staple tracker}
The samples used to train the filter can be various features, such as fHOG \cite{Felzenszwalb2010TPAMI}, CN \cite{Weijer2006ECCV}, and color histogram, \emph{etc}. In the Staple tracker \cite{bertinetto2016CVPR}, L. Bertinetto \emph{et al.} found that different features have different advantages under different tracking scenarios. For example, the fHOG feature has a good expression effect under illumination variation, while the color histogram is more powerful under deformation and rotation. Based on this discovery, how to make the two complementary feature fuse in tracking becomes the focus of the Staple algorithm.

In \cite{bertinetto2016CVPR}, L. Bertinetto \emph{et al.} proposed a complementary fusion tracking method, which simultaneously uses the main formula of the DSST tracker, \emph{i.e.}, Eq.~(\ref{eqn:21}), Eq.~(\ref{eqn:28}), to train template filters $\mathbf{w}$ base on fHOG feature, \emph{i.e.}, Eq.~(\ref{eqn:19}) and another ridge regression equation to train histogram weights $\mathbf{h}$:
\begin{equation}
\mathcal{E}_{\rm hist}(\mathbf{h})=\frac{1}{|\mathcal{O}|}\sum_{O_i\in \mathcal{O}}(\mathbf{h}^{\rm T}\psi[O_i]-1)^2+\frac{1}{|\mathcal{B}|}\sum_{O_i\in\mathcal{B}}(\mathbf{h}^{\rm T}\psi[O_i])^2~,
\end{equation}
where $\mathcal{O}$ and $\mathcal{B}$ represent the the object and background region respectively. $|\mathcal{B}|$ denotes total samples $O_i$ in background region in the image patch $\mathbf{O}$. $\psi[O_i]$ indicates an $M$-channel feature transform function, that maps image patch $\mathbf{O}$ to an $M$-dimensional color feature space. Then the elements $h^j$ in histogram weight $\mathbf{h}$ can be obtained:
\begin{equation}
h^j=\frac{\rho(\mathcal{O})^j}{\rho(\mathcal{B})^j+\rho(\mathcal{O})^j+\lambda}~,
\end{equation}
where $\rho(\mathcal{O})^j$ denotes the frequency of color bin $j$ in the object region, and $\rho(\mathcal{B})^j$ is similar. Thus, in search stage, the histogram score $\mathbf{r}_{\rm hist}$ can be calculated by:
\begin{equation}
r(\mathbf{O})_{\rm hist}=\frac{1}{|\mathcal{S}|}\mathbf{h}^{\rm T}\psi[\mathbf{O}]\odot\mathbf{y}~,
\end{equation}
where $\mathcal{S}$ is the search region, $\frac{1}{|\mathcal{S}|}\mathbf{h}^{\rm T}\psi[\mathbf{O}]$ denotes the likelihood map, $\mathbf{O}$ represents the original search region image patch, and $\mathbf{y}$ indicates the Gaussian shaped labels.

\Remark The response map obtained by two methods are linearly superimposed with the following formula for the final object detection.
\begin{equation}
\mathbf{r}(\mathbf{z})=\gamma_{\rm {hist}}\mathbf{r}(\mathbf{O})_{\rm {hist}}+\gamma_{\rm {template}}\mathbf{r}(\mathbf{z})_{\rm {template}}~.
\end{equation}

\subsubsection{MCCT-H tracker}
The features used by the DCF-based trackers that achieve real-time on UAV can be summarized as: fHOG \cite{Felzenszwalb2010TPAMI} and CN \cite{Weijer2006ECCV}, \emph{etc}. Now the main processing methods for multi-channel features are: (1) directly overlay by layer, such as the KCF \cite{Henriques2015TPAMI}, SAMF \cite{li2014ECCV}, DSST trackers \cite{danelljan2014BMVC}, \emph{etc.}; (2) use PCA to reduce dimension, like the ECO tracker \cite{danelljan2017CVPR}, the CN tracker \cite{Danelljan2014CVPR}, and the fDSST tracker \cite{danelljan2017TPAMI}. Nevertheless, these methods have obvious limitations. The stability and reliability of each feature channel under different frames are usually unequal and the number of reliable features is not necessarily constant. Thus, both directly overlaying by layer (also parallel tracking) and using PCA to reduce dimensionality may cause information redundancy, missing, and even unreliability.

Based on the above considerations, N. Wang \emph{et al.} proposed a multi-cue joint tracking scheme (MCCT) \cite{wang2018CVPR}, introducing a feature pool and an expert pool (different combinations of features in the feature pool) to select the most reliable expert as the features for tracking in different frames, which compared with the previous fusion methods have achieved better results.

In the MCCT tracker, basic training structure of the DSST tracker \cite{danelljan2014BMVC}, Eq.~(\ref{eqn:21}), Eq.~(\ref{eqn:28}), and Eq.~(\ref{eqn:19}), is adopted. Innovatively, it introduces feature pool and expert pool. The feature pool consists of 3 types of features $\{ \rm Low,Middle,High\}$ ,corresponding to $\{ \rm fHOG_1,fHOG_2,CN\}$ (MCCT-H). Optionally combining the features into $C^1_3+C^2_3+C^3_3=7$ experts $E_1, E_2, \cdots, E_7$. To rate the experts, they have Pair-Evaluation score:
\begin{equation}
R^t_{pair}(E_i)=\frac{M'^t_{E_i}}{V'^t_{E_i}+\xi}~,
\end{equation}
where $M'^t_{E_i}$ and $V'^t_{E_i}$ are both calculated based on the overlap ratio of the bounding box generated by expert $E_i$ and others. Differently, $M'^t_{E_i}$ directly denotes the consistency of expert $E_i$ and others, while $V'^t_{E_i}$ is the temporal stability of the overlap ratio. $\xi$ is used to avoid zero denominators. The bigger the score $R^t_{pair}(E_i)$ is, the more pair judging reliable expert $E_i$ is. They also introduced the Self-Evaluation score:
\begin{equation}
R^t_{self}(E_i)=\frac{1}{N}\sum_{\tau}W_\tau S^\tau_{E_i}~,
\end{equation}
where $S^\tau_{E_i}$ is calculated according to the Euclidean distance of the bounding box temporal change, and $\frac{1}{N}\sum_{\tau}W_\tau$ denotes the average in frames $W$. Self-Evaluation score $R^t_{self}(E_i)$ actually measures the smoothness degree. The final robustness score $R^t(E_i)$ is defined as:
\begin{equation}
R^t(E_i)=\mu\cdot R^t_{pair}(E_i)+(1-\mu)\cdot R^t_{self}(E_i)~.
\end{equation}

Thus, the expert with the highest robustness score is selected for the current tracking.

Another innovative highlight in the MCCT-H tracker is that their update scheme uses an adaptive learning rate. By computing the average PSR $P=(R_{\rm max}-m)/\sigma$ of different features:
\begin{equation} P^t_{mean}=\frac{1}{3}(P^t_H+P^t_M+P^t_L)~,
\end{equation}
to evaluate the tracking result, and learning rate $\eta$ is changed according to PSR, which is no longer a constant.

\subsection{Boundary effect}
The circulant samples used by the DCF-based trackers have a huge disadvantage, that is, the boundary effect. The aforementioned filters usually used a cosine window to weaken the boundary effect, which can only reduce the boundary effect to a certain extent. Such a method cannot solve the problem of inaccurate response value to the circulant samples that centered at the boundary to a large extent. Under such consideration, trackers such as the KCF tracker \cite{Henriques2015TPAMI} and the DSST tracker \cite{danelljan2014BMVC} the only sample 2-3 times the original object size. To settle this issue, brilliant trackers \cite{Danelljan2015ICCV,danelljan2016CVPR,lukezic2017CVPR,Galoogahi2017ICCV} which enabled sufficient negative samples learned while alleviating boundary effect, thus ensuring better tracking robustness.

\subsubsection{SRDCF \& SRDCFdecon tracker}
The most favourable highlight of the SRDCF tracker \cite{Danelljan2015ICCV} is the introduction of a spatial punishment weight in the filter training process. Their improved main training formula for filter in the $T$-th frame $\mathbf{w}_T$ is:
\begin{equation}\label{eqn:34}
\mathcal{E}(\mathbf{w}_T)=\sum_{f=1}^{T}\theta_f\left\|\sum_{c=1}^{D}\mathbf{w}^c_T\star\mathbf{x}^c_f-\mathbf{y}^c\right\|^2_2+\sum_{c=1}^{D}\left\|\bm{\omega}\odot\mathbf{w}^c_T\right\|^2_2~,
\end{equation}
where $\bm{\omega}$ is the given regularization weight, which has a larger value far from the center point and a smaller value near the center point. Thus, it can increase the weight at the center of the filter, making the more reliable center part of samples posses higher impact on response map and shielding the noise or unreliable negative samples far from the center. $\theta_f$ denotes weight of training samples in the $f$-th frame, $\mathbf{w}^c\in\mathbb{R}^N$ indicates the $c$-th feature channel of the filter $\mathbf{w}$.

The SRDCF tracker is also unique in solving the new regression equation. First, the Parcival formula is used, which turned Eq.~(\ref{eqn:34}) to:
\begin{equation}
\hat{\mathcal{E}}(\mathbf{w}_T)=\sum_{f=1}^{T}\theta_f\left\|\sum_{c=1}^{D}\hat{\mathbf{w}}_T^c\mathcal{D}(\hat{\mathbf{x}}^c_f)-\hat{\mathbf{y}}\right\|^2_2+\sum_{c=1}^{D}\left\|\mathcal{C}(\hat{\bm{\omega}})\hat{\mathbf{w}}_T^c\right\|^2_2~.
\end{equation}

Here, $\mathcal{D}(\hat{\mathbf{x}}^c)$ denotes a diagonal matrix with elements in $\hat{\mathbf{x}}^c$, and $\mathcal{C}(\hat{\bm{\omega}})$ is a $N\times N$ matrix. Then they use Gauss-Seidel iteration to solve the simplified calculation.

The SRDCF tracker is a milestone in the development of the correlation filter trackers. It supplements the KCF's ridge regression main formula with spatial penalty terms and used the Gauss-Seidel iteration method to solve it. Experiments have confirmed that the SRDCF algorithm enables the filter to learn more negative samples, which improves the robustness of tracking, which has become a popular baseline for subsequent CF trackers.

\Remark Besides, the improvement of the SRDCF tracker's performance is also due to the addition of all past samples, which slowed their calculation speed as well.

M. Danelljan further proposed the SRDCF decon tracker \cite{danelljan2016CVPR} on the basis of the SRDCF tracker. The SRDCF decon tracker adopted the similar main formula as SRDCF:
\begin{equation}
\begin{split}
\mathcal{E}(\mathbf{w}_T,\theta)=\sum_{f=1}^{T}\theta_f\left\|\sum_{c=1}^{D}\mathbf{w}_T^c\star\mathbf{x}^c_f-\mathbf{y}\right\|^2_2+\frac{1}{\mu}\sum_{f=1}^{T}\frac{\theta_f^2}{\rho_f}\\
+\sum_{c=1}^{D}\left\|\bm{\omega}\odot\mathbf{w}_T^c\right\|^2_2~,
\end{split}
\end{equation}
where $\mu>0$ represents the flexibility parameter, and $\rho_f>0$, satisfying $\sum_f\rho_f=1$, denotes the prior sample weights.

The difference is that the main formula of the SRDCF decon algorithm is binary regression, namely, filter $\mathbf{w}$ and weight $\theta$. Therefore, for SRDCF decon, the weight $\theta$ of the sample used for filter training in the previous $T$ frames is no longer fixed but is solved dynamically according to the regression equation. Thus, the online adjustment of the weight can continuously the influence of previous $T$ frames, which ensures that the impact of the frames with high confidence is greater and lessen the impact of contaminated samples.

\subsubsection{CSR-DCF tracker}
Note that the spatial punishment $\bm{\omega}$ in the Eq.~(\ref{eqn:34}) is fixed, while the importance of each pixel doesn't necessarily subjected to a fixed uniform declining from the middle to the surrounding. Besides, the reliability of each feature layer of the object sample is also different, in other words, the prior tracking methods learned some fake information that may disturb tracking. 

In order to ensure the filter learns more accurate object information with irregular shapes and weaken the unreliable feature channels. A. Lukezic \emph{et al.} made splendid improvements to the SRDCF tracker \cite{danelljan2016CVPR} in the CSR-DCF tracker \cite{lukezic2017CVPR} to change the fixed spatial punishment item $\mathbf{\omega}$ to a pixel-by-pixel confidence map $\mathbf{m}$, and introduced the channel confidence score to modify the generated response map.

In the CSR-DCF tracker, A. Lukezic \emph{et al.} introduced a spatial confidence map $\mathbf{m}$, where the value of each element $m_i\in\mathbf{m}$ represents the probability that each pixel is the object. The calculation method is divided into three steps: the prior layer, the Bayesian probability under the color model, and the Epanechnikov kernel:
\begin{equation}
p(m_i=1|\mathbf{x},x_i)\propto p(m_i=1)p(\mathbf{x}|m_i=1,x_i)p(x_i|m_i=1)~,
\end{equation}
where the prior $p(m_i=1)$ is defined by the ratio between the region sizes for color histogram, the Bayesian probability under the color model $p(\mathbf{x}|m_i=1,x_i)$ adopts the similar way as the Staple tracker \cite{bertinetto2016CVPR} which is actually a likelihood map, and $p(x_i|m_i=1)$ is calculated using a modified Epanechnikov kernel.

Having obtained the spatial confidence map, the author proposed their main training formula:
\begin{equation}\label{eqn:51}
\mathcal{E}(\mathbf{w})=\left\|\sum_{c=1}^{D}\mathbf{m}\odot\mathbf{w}^c\star\mathbf{x}^c-\mathbf{y}\right\|_2^2+\frac{\lambda}{2}\sum_{c=1}^D\left\|\mathbf{m}^c\odot\mathbf{w}^c\right\|_2^2~.
\end{equation}

In order to perform Augmented Lagrange iteration, Eq.~(\ref{eqn:51}) can be first transformed into:
\begin{equation}\label{eqn:52}
\begin{split}
\mathcal{E}(\mathbf{w})=\left\|\hat{\mathbf{w}}^{\rm H}_d \mathcal{D}(\mathbf{\hat{x}})-\hat{\mathbf{y}}\right\|_2^2+\frac{\lambda}{2}\left\|\mathbf{w}_m\right\|_2^2+\\
[\hat{\mathbf{l}}^H(\hat{\mathbf{w}}_d-\hat{\mathbf{w}}_m)
+\overline{\hat{\mathbf{l}}^H(\hat{\mathbf{w}}_d-\hat{\mathbf{w}}_m)}]+\mu\left\|\hat{\mathbf{w}}_d-\hat{\mathbf{w}}_m\right\|_2^2~,
\end{split}
\end{equation}
where $\mathbf{w}_m=\mathbf{m}\odot\mathbf{w}$, and $\mathbf{w}_d$ is a dual variable satisfying $\mathbf{w}_d-\mathbf{m}\odot\mathbf{w}\equiv0$. Note that $\hat{\mathbf{l}}$ is a complex Lagrange multiplier, and $\mu>0$. Using Augmented Lagrange iteration, Eq.~(\ref{eqn:52}) can be solved.

Besides, CSR-DCF algorithm also proposed per-channel detection reliability score:
\begin{equation}
\varsigma^c=1-{\rm min}\left(\mathbf{r}(\mathbf{z})_{\rm max 2}/\mathbf{r}(\mathbf{z})_{\rm max 1},\frac{1}{2}\right)~,
\end{equation}
where, $\hat{\mathbf{r}}(\mathbf{z})_{\rm max 2}$ and $\mathbf{r}(\mathbf{z})_{\rm max 1}$ respectively denotes the second and first major mode in the response map. In this way, it can surpass response map containing noise or distractor.

\subsubsection{BACF tracker}
Background-aware correlation filter (BACF) tracker \cite{kiani2017ICCV} proposed by H. Kiani Galooghi \emph{et al.} also enables the filter to learn abundant background information while reducing boundary effect.

\Remark The BACF tracker utilizes a cropping matrix to automatically crop the samples in the large ROI into multiple small samples of the same size as the object. To be more specific, these small samples are generated by circulant shift plus a cropping operator which are all sub-regions of ROI. This makes it possible to introduce more background samples without introducing too much boundary effect.

In \cite{kiani2017ICCV}, the author first introduced the cropping matrix $\mathbf{C}\in\mathbb{R}^{N\times L}$, which can select and extract pixels of fixed size $\mathbb{R}^{N}$ in the center area of the sample $\mathbf{x}^c\in\mathbb{R}^{L}$, and the circulant shift operator $\mathbf{P}^j$ the same in the KCF tracker \cite{Henriques2015TPAMI}. Thus, $\mathbf{C}\mathbf{x}^c\mathbf{P}^j$ can be regarded as the central area of the circulant sample $\mathbf{x}^j$. In different circulant samples, such center area corresponds to positive samples or negative samples, which is consistent with the size of the corresponding position value of the expected Gaussian label $\mathbf{y}(j)$. Note that since $L>>N$, the samples used in the BACF tracker is far larger than other trackers, ensuring abundant negative samples learning. Based on the idea above, the BACF tracker put forward their main training formula as:
\begin{equation}\label{eqn:48}
\mathcal{E}(\mathbf{w})=\frac{1}{2}\sum_{j=1}^{L}\left\|\sum_{c=1}^{D}\mathbf{w}^{c\rm T}\mathbf{Cx}^c\mathbf{P}^j-\mathbf{y}(j)\right\|_2^2+\frac{\lambda}{2}\sum_{c=1}^{D}\left\|\mathbf{w}^c\right\|_2^2~,
\end{equation}
where the regression label $\mathbf{y}$ is in $\mathbb{R}^{L}$ instead of $\mathbb{R}^{N}$. In this way, the search region is enlarged, the real negative samples increase and the circularity of samples is ensured at the same time. To apply the ADMM algorithm, Eq.~(\ref{eqn:48}) can be first divided into:
\begin{equation}
\begin{array}{ccc}
\mathcal{E}(\mathbf{w,g})&=&\frac{1}{2}\left\|\mathbf{\hat{X}\hat{g}-\hat{y}}\right\|^2_2+\frac{\lambda}{2}\left\|\mathbf{w}\right\|^2_2~,\\
&\rm s.t.&\mathbf{\hat{g}}=\sqrt{T}(\mathbf{F}_L\mathbf{P}^{\rm T}\otimes\mathbf{I}_D)\mathbf{w}~,
\end{array}
\end{equation}
where $\otimes$ denotes Kronecker product.

Due to its possession of both amazing tracking speed and impressive tracking performance, the BACF tracker's core architecture has become another milestone in the history of correlation filter tracking. Since then, many DCF-based trackers have chosen BACF tracker as their baseline and have made even more innovative contributions.

\subsection{Temporal consistency}

The appearance of the object during tracking sometimes undergoes a short-term large variation, such as partial occlusion, similar object interference, \emph{etc.} These tracking scenarios usually cause filter degradation, which leads to subsequent tracking failures. To ensure temporal consistency of the filter \cite{Li2018CVPR} of response map \cite{Huang2019ICCV,Li2020CVPR} are effective methods to encounter the unexpected abrupt degradation.

\subsubsection{STRCF tracker}
F. Li \emph{et al.} put forward the spatio-temporal regularized correlation filter (STRCF) \cite{Li2018CVPR}, which introduced the temporal regularization term  based on the previous main training formula of the SRDCF tracker \cite{danelljan2016CVPR}, \emph{i.e.}, Eq.~(\ref{eqn:34}) to suppress excessive short-term object appearance changes and obtain a more stable tracking effect on the basis of the SRDCF tracker:
\begin{equation}\label{eqn:56}
\begin{aligned}
\mathcal{E}(\mathbf{w}_f)=\frac{1}{2}\left\|\sum_{c=1}^{D}\mathbf{w}^c_f\star\mathbf{x}^c_f-\mathbf{y}\right\|_2^2&+\frac{1}{2}\sum_{c=1}^{D}\left\|\bm{\omega}\odot \mathbf{w}^c_f\right\|_2^2\\
&+\frac{\mu}{2}\sum_{c=1}^{D}\left\|\mathbf{w}^c_f-\mathbf{w}^c_{f-1}\right\|_2^2~,
\end{aligned}
\end{equation}
where $\left\|\mathbf{w}_f-\mathbf{w}_{f-1}\right\|_2^2$ represents temporal regularization term according to passive-aggressive algorithm. Also, the STRCF tracker uses ADMM iteration for efficient solution of Eq.~(\ref{eqn:56}). 

\Remark Due to the introduction of the temporal regular term, the main training formula of STRCF no longer needs the information of all previous $T$ frames, which greatly reduces the memory required for training and greatly improves the calculation speed compared to the SRDCF tracker.

Experiments have proved that the filter in the STRCF algorithm has strong temporal stability. It is especially robust under scenes where the object undergoes rapid appearance variation such as fast motion and illumination variation. The time regularization term under the PA algorithm is also very useful for further reference.

\subsubsection{ARCF-H \& ARCF-HC tracker}
Based on the aforementioned abundant DCF-based tracking research, the DCF-based tracker specifically for UAV tracking finally landed on the stage \cite{Huang2019ICCV}, \cite{Li2020CVPR}. 

A huge shortcoming of the existing DCF-based trackers in the UAV tracking environment is the poor anti-interference ability. First, the UAV tracking scene is more complex, which usually encounters large environmental disturbances, such as illumination variation, similar objects, \emph{etc.}, affecting the tracking results. Secondly, the previous correlation filter trackers usually expand the search region and impose a spatial penalty term to solve the boundary effect, which would inevitably cause more background information and thus are more likely to introduce environment noise. 

To solve this problem and improve the performance of the correlation filter tracker in the tracking environment with serious background interference onboard UAV, Z. Huang \emph{et al.} proposed the aberrance repressed correlation filter algorithm (ARCF) \cite{Huang2019ICCV}.

The core idea of the ARCF tracker is to use the aberrance of the successive response maps to suppress and learn the environmental noise. 

\Remark Different from the previous methods that utilize the reliability of current response map to suppress the corresponding position in the subsequent frames, \emph{e.g.}, the CSR-DCF tracker \cite{lukezic2017CVPR}, the ARCF tracker introduced the response map aberrance into the training formula for the first time, suppressing possible noise in the training stage, and have achieved quite considerable results.

In \cite{Huang2019ICCV}, the author first defined the noise evaluation standard:
\begin{equation}\label{eqn:57}
\left\|\mathbf{M_1}[\psi_{p,q}]-\mathbf{M}_2\right\|^2_2~,
\end{equation}
where $\mathbf{M}_1$ and $\mathbf{M}_2$ represent two response maps, $\psi_{p,q}$ indicates the shifting operation that makes the locations of the peaks in $\mathbf{M}_1$ and $\mathbf{M}_2$ the same. Thus, when aberrance occurs, the Euclidean distance of $\mathbf{M}_1[\psi_{p,q}]$ and $\mathbf{M}_2$ rises, ending up in the increase of Eq.~(\ref{eqn:57}).

Based on the above standard, the training equation of the ARCF algorithm can be expressed as:
\begin{equation}\label{eqn:58}
\begin{split}
\mathcal{E}(\mathbf{w}_f)=\frac{1}{2}\left\|\sum_{c=1}^{D}\mathbf{w}^c_f\star\mathbf{Cx}^c_f-\mathbf{y}\right\|^2_2+\frac{\lambda}{2}\sum_{c=1}^{D}\left\|\mathbf{w}^c_f\right\|^2_2+\\
\frac{\gamma}{2}\left\|\sum_{c=1}^{D}(\mathbf{w}^c_{f-1}\star\mathbf{Cx}^c_{f-1})[\psi_{p,q}]-\sum_{c=1}^{D}\mathbf{w}^c_f\star\mathbf{Cx}^c_f\right\|^2_2~,
\end{split}
\end{equation} 
where matrix $\mathbf{C}$ in the first item is the cropping matrix in the BACF tracker \cite{kiani2017ICCV}, the second item is a regular item the same as the KCF tracker \cite{Henriques2015TPAMI}, and the third item is meant to suppress the possible response map aberrance in the $f$-th frame relative to the $f-1$-th frame. Finally, the ARCF tracker adopts the ADMM iterative algorithm to solve Eq.~(\ref{eqn:58}).

Experiments have confirmed that the ARCF tracker's suppression effect on response map aberrance is very significant on most UAV-tracking benchmarks. For scenes with frequent noises such as fast movement and occlusion, the ARCF tracker stands out among the state-of-the-arts. Especially, under small object tracking scenes. Due to the low object resolution and insufficient information, aberrance is more likely to appear owing to background environment interference. The ARCF tracker showed outstanding performance under such scenes and has become one of the most preferred algorithms for UAV-based aerial tracking.

\subsubsection{AutoTrack tracker}
Another well-known correlation filter tracker specifically for UAV tracking uses the STRCF tracker \cite{Li2018CVPR} as its baseline.

Although the stability and robustness of the STRCF tracker are excellent, its spatio-temporal regularization term introduces too many parameters that need to be manually set, such as spatial punishment $\omega$ and temporal parameters $\mu$. On the one hand, these parameters usually cost much time to adjust to find the best one during the experiment. On the other hand, the determined parameters can't always perform well in every sequence.

Based on the above considerations, Y. Li \emph{et al.} proposed a DCF-based tracker with automatic spatial-temporal regularization (AutoTrack) \cite{Li2020CVPR}, which uses a local response map and global response map to respectively and automatically adjust spatial weights and temporal weights in dynamically, which is both highly adaptable to different sequences and efficient at the same time.

In \cite{Li2020CVPR}, the author firstly defined a response map reliability vector $\mathbf{\Pi}$, whose element $\Pi^i$ can be expressed as:
\begin{equation}
\Pi^i=\frac{\{\mathbf{M}_{f}[\psi_{p,q}]\}^i-\mathbf{M}_{f-1}^i}{\mathbf{M}_{f-1}^i}~,
\end{equation}
where $\mathbf{M}_{f-1}$ and $\mathbf{M}_f$ denotes response map in the $f-1$-th
frame and the $f$-th frame respectively. $\psi_{p,q}$ is the same shifting operator that makes two peaks in two response maps positionally coincide in the ARCF tracker \cite{Huang2019ICCV}. Then, the automatic spatial weight $\tilde{\bm{\omega}}$ can be defined as:
\begin{equation}\label{eqn:60}
\bm{\tilde{\omega}}=\mathbf{C}^{N}\delta\rm {log}(\mathbf{\Pi}+\mathbf{I})+\bm{\omega}~,
\end{equation}
where, $\mathbf{C}^{N}$ is used to crop the center region $\mathbb{R}^{N}$ which denotes the object template size, $\delta$ is a constant aimed to adjust the weight, and $\mathbf{\omega}$ is the same fixed spatial punishment term to settle boundary effect as in the SRDCF tracker. Thus, an automatic spatial regularization term according to local response map can be calculated, where the higher the value is, the less reliable the pixel is, resulting in smaller weight in filter $\mathbf{w}$.

In the meanwhile, the author defined automatic temporal parameter as:
\begin{equation}
\tilde{\mu}=\frac{\zeta}{1+\rm {log}(\nu \left\|\mathbf{\Pi}\right\|_2+1)}~, \left\|\mathbf{\Pi}\right\|_2\le \tau
\end{equation}
where $\zeta$ and $\nu$ are all hyper parameters. Only when $\left\|\mathbf{\Pi}\right\|_2$ is smaller than the threshold $\tau$ (denotes that the samples are reliable) does the CF learn. When $\left\|\mathbf{\Pi}\right\|_2\le \tau$, the higher $\left\|\mathbf{\Pi}\right\|_2$ is, the less reliable the samples are, resulting in a smaller $\tilde{\mu}$, making the filter $\mathbf{w}$ change little.

Based on the above, the AutoTrack tracker gave its overall objective as:
\begin{equation}
\begin{split}
\mathcal{E}(\mathbf{w}_f,\mu_f)=\frac{1}{2}\left\|\sum_{c=1}^{D}\mathbf{w}_f^c\star\mathbf{x}_f^c-\mathbf{y}\right\|^2_2+\frac{1}{2}\sum_{c=1}^{D}\left\|\bm{\tilde{\omega}}\odot\mathbf{w}_f^c\right\|_2^2\\
+\frac{\mu_f}{2}\sum_{c=1}^{D}\left\|\mathbf{w}_f^c-\mathbf{w}_{f-1}^c\right\|2_2+\frac{1}{2}\left\|\mu_f-\tilde{\mu}\right\|_2^2~,
\end{split}
\end{equation}
where $\bm{\tilde{\omega}}$ denotes the automatic spatial weights obtained by Eq.~(\ref{eqn:60}). $\mu_f$ and $\tilde{\mu}$ indicates the optimized temporal parameter and reference parameter respectively.

Abundant experiments have demonstrated that the AutoTarck tracker is extremely robust to tracking scenes where object undergoes appearance changes, such as illumination variation, viewpoint change, \emph{etc.}

\subsection{Extra types}
Some DCF-based trackers provide extra solutions for long-term tracking \cite{ma2015CVPR}, computation acceleration \cite{danelljan2017CVPR}, background suppression \cite{mueller2017CVPR}, and calculation tricks \cite{wang2018AAAI} make outstanding contributions to DCF based tracking communities.

\subsubsection{LCT tracker}
Compared to short-term tracking, long-term tracking is usually more challenging. Due to more extreme challenges like full-occlusion and out of view, the DCF-based trackers in the long-term tracking scenes are more likely to encounter tracking failures. To settle this problem, C. Ma \emph{et al.} gave an innovative tracking idea in the LCT tracker \cite{ma2015CVPR}. The idea in the LCT tracker is similar to a famous previous work the TLD tracker \cite{Kalal2012TPAMI}. In the TLD tracker, long-term tracking is divided into tracking learning and detection phases. The tracker learns frame by frame and determines the position of the object in each frame, while the detector learns the large appearance change of the object and searches for the object globally when the tracking fails.

The first innovation is that two training models are introduced in the LCT tracker, \emph{i.e.}, the temporal regression model $R_c$, and the object appearance regression $R_t$. Among them, $R_c$ used both object and the surrounding context to train the model, which is the same as the KCF tracker \cite{Henriques2015TPAMI}.

As a remedy, the main purpose of $R_t$ is to predict the current state of the object and make an appropriate scale estimation, which is similar to the scale filter in the DSST tracker \cite{danelljan2014BMVC}. Therefore, $R_t$ only needs to learn the information of the object, and updates only when the peak value of the corresponding response map $\mathbf{r}(\mathbf{z})$ exceeds the fixed threshold $\tau$ (indicates that the object state is reliable).

When the peak value of the response map $\rm {max}(\mathbf{r}(\mathbf{z}))$, the LCT tracker judges that tracking failure occurs. Then it turns on a random fern re-detector to search for the object, and update $R_t$ only when the peak of the corresponding response map exceeds the threshold to ensure the reliability of $R_t$. Note that other tracking-by-detection methods usually perform re-detection frame by frame, while the LCT tracker re-detects only when tracking failure occurs, which ensured both speed and robustness.

Later, C. Ma \emph{et al.} further proposed the LCT2.0 tracker \cite{ma2018IJCV}, where the random fern classifier in the LCT tracker is substituted by an SVM classifier, and applied histogram of local intensity as an additional expression.

\subsubsection{ECO-HC tracker}
Although the results of the previous work \cite{Danelljan2016ECCV} are considerable, it has two serious problems: first, the operation speed is slow due to excessive calculation. Second, too many parameter settings, \emph{e.g.}, too many feature dimensions, too many frames involved, made the tracker prone to overfitting. Therefore, on the basis of their previous work \cite{Danelljan2016ECCV}, M. Danelljan \emph{et al.} further proposed a faster and more robust ECO tracker \cite{danelljan2017CVPR}, which improved the performance of the correlation filter tracking method to a new level. Note that, here we only discuss the ECO tracker using handcrafted features, instead of CNN features, \emph{i.e.}, ECO-HC.

In \cite{danelljan2017CVPR}, M. Danelljan \emph{et al.} first analyzed the time complexity of the C-COT tracker, which can be expressed as $\mathcal{O}(N_{\rm CG}DT\bar{K})$. $N_{\rm CG}$ denotes the number of conjugate gradient iterations, $D$ indicates the number of feature channels of the training samples, $T$ the total number of frames involved in the operation, and $\bar{K}$ represents the average number of Fourier coefficient of each filter channel. Based on such analysis, three methods are proposed to reduce computational complexity respectively.

Firstly, the ECO tracker reduces the number of feature dimensions for higher speed. The original $D$ dimension is reduced to $E$ dimension samples, and a matrix $\mathbf{G}$ is introduced to represent the original samples as follows:
\begin{equation}
	\mathbf{x}=\left(\begin{array}{c}
	\mathbf{x}^1\\
	\vdots\\
	\mathbf{x}^d
	\end{array}\right)
	=\left(\begin{array}{ccc}
	G_{11}&\cdots&G_{1e}\\
	\vdots&\cdots&\vdots\\
	G_{d1}&\cdots&G_{de}
	\end{array}\right)\left(\begin{array}{c}
	\mathbf{x}^1\\\vdots\\\mathbf{x}^e
	\end{array}\right)=\mathbf{G}\mathbf{x}'~,
\end{equation}
which realized dimensionality reduction, and then updated the loss function to binary nonlinear regression as:
\begin{equation}
\begin{split}
\mathcal{E}(\mathbf{w}_T,\mathbf{G})=\sum_{f=1}^{T}\theta_f\left\|\sum_{c=1}^{E}\mathbf{w}_T^c\star J\{\mathbf{G}\mathbf{x}^c\}-\mathbf{y}\right\|_2^2+\\
\sum_{c=1}^{E}\left\|\bm{\omega}\odot\mathbf{w}_T^c\right\|_2^2+\lambda\left\|\mathbf{G}\right\|_{\rm F}^2~.
\end{split}
\end{equation}

Note that $J\{\mathbf{x}\}$ denotes the new training samples in the transformed continuous time-space domain \cite{Danelljan2016ECCV}. $\left\|\cdot\right\|^2_{\rm F}$ indicates Frobenius norm. Utilizing Conjugate Gradient method $\mathbf{w}$ and $\mathbf{G}$ can be solved.

Secondly, the ECO tracker simplifies and compresses the prior sample model using the Gaussian Mixture Model to prevent overfitting. Their main equation can be expressed as:
\begin{equation}
\begin{split}
\mathcal{E}(\mathbf{w}_T,\mathbf{G})=\sum_{f=1}^{V}\pi_f\left\|\sum_{c=1}^{E}\mathbf{w}_T^c\star J\{\mathbf{G}\bm{\mu}_f^c\}-\mathbf{y}\right\|_2^2+\\
\sum_{c=1}^{E}\left\|\bm{\omega}\odot\mathbf{w}_T^c\right\|_2^2+\lambda\left\|\mathbf{G}\right\|_{\rm F}^2~.
\end{split}
\end{equation}
where previous samples in the $f$-th frame $\mathbf{x}_f$ are replaced by Gaussian means $\bm{\mu}_f$, weights $\theta_f$ is substituted by $\pi_f$, and the number of total frames $T$ is reduced to $V$.

Lastly, in the ECO algorithm, the filter model is updated every 5 frames instead of frame by frame. Thus, the ECO tracker achieved both even more favorable performance and amazing speed.

\subsubsection{CACF tracker}
Although the spatial punishment in the SRDCF tracker \cite{danelljan2016CVPR} can repress background noises, it utilizes a fixed global punishment, that is, artificially reducing the weight of the filter periphery to suppress the background response. While in the means of improving the discriminative ability of the filter, especially learning background information can usually achieve a better effect. Aiming to learn context information, M. Mueller \emph{et al.} proposed context-aware correlation filter tracker (CACF) \cite{mueller2017CVPR}.

In the CACF tracker, during the training stage, the author changed the original ridge regression formula Eq.~(\ref{eqn:3}) to:
\begin{equation}
\begin{split}
\mathcal{E}(\mathbf{w})=\left\|\sum_{c=1}^{D}\mathbf{w}^c\star\mathbf{x}^c_0-\mathbf{y}\right\|_2^2+\lambda_1\sum_{c=1}^{D}\left\|\mathbf{w}^c\right\|_2^2\\
+\lambda_2\sum_{i=1}^{k}\left\|\sum_{c=1}^{D}\mathbf{w}^c\star\mathbf{ x}^c_i\right\|_2^2~,
\end{split}
\end{equation}
where $\mathbf{x}_0$ denotes training samples from the object region, and $\mathbf{x}_i$ represents samples from the context around the object region.

\Remark Innovatively, to obtain a structure similar to the ridge regression equation, the author further derives the deformation of the main formula as:
\begin{equation}\label{eqn:44}
	\mathcal{E}=\left\|\sum_{c=1}^{D}\mathbf{w}^c\star \mathbf{\bar{x}}^c-\mathbf{\bar{y}}\right\|_2^2+\lambda_1\sum_{c=1}^{D}\left\|\mathbf{w}^c\right\|_2^2~,
\end{equation}
where $\mathbf{\bar{x}}^c$ and $\mathbf{\bar{y}}$ can be expressed as: 
\begin{equation}
	\mathbf{\bar{x}}^c=\left[\begin{array}{c}
	\mathbf{x}^c_0 \\
	\sqrt{\lambda_2}\mathbf{x}^c_1\\
	\vdots\\
	\sqrt{\lambda_2}\mathbf{x}^c_k
	\end{array}\right]\quad
	\bar{\mathbf{y}}=\left[\begin{array}{c}
	\mathbf{y} \\
	{\bm 0}_N\\
	\vdots\\
	{\bm 0}_N
	\end{array}\right]~,
\end{equation}
where ${\bm 0}_N \in \mathbb{R}^N$ denotes zero vector. The solution to Eq.~(\ref{eqn:44}) can be obtained:
\begin{equation}
	\mathbf{\hat{w}}^c=\frac{\mathbf{\hat{x}}^{c*}_0\odot\mathbf{\hat{y}}}{\mathbf{\hat{x}}^{c*}_0\odot\mathbf{\hat{x}}^c_0+\lambda_1+\lambda_2\sum_{i=1}^{k}\mathbf{\hat{x}}^{c*}_i\odot\mathbf{\hat{x}}^c_i}~.
\end{equation}

Experiments have shown that trackers with the context-aware algorithm in the CACF tracker, \emph{i.e.}, the Staple\_CA tracker and the SAMF\_CA tracker can be more robust compared with the original tracker in most scenes.

\Remark Although the CACF tracker \cite{mueller2017CVPR} can make the filter learn the background information near the object bounding box, it has an obvious shortcoming: the background information learned by the CACF tracker is the fixed 4 background boxes in the relative position around the object, which is not enough for the tracker to tackle complex scenes.

\subsubsection{KCC tracker}
The kernel trick used in the KCF tracker \cite{Henriques2015TPAMI} can map samples $\mathbf{x}$ to high-dimension space for division, and thus has a relatively good shielding effect on noise and distractors. However, there are two obvious limitations in the KCF tracker: (1) due to the application of Eq.~(\ref{eqn:9}), the training samples must be circulant; (2) the kernel function used $\mathbf{k}^{\mathbf{xx'}}$ is also required to have the same weight for each pixel in the samples. 

In the KCC tracker \cite{wang2018AAAI}, C. Wang \emph{et al.} unearthed the huge potential of the kernel method, which not only eliminates the two limitations in the KCF tracker but also further expanded the kernel method to the calculation of changes such as scale and rotation (not just translation), proved to be more robust than scaling pool in the SAMF tracker \cite{li2014ECCV} and the scale filter in the DSST tracker \cite{danelljan2014BMVC}.

In \cite{wang2018AAAI}, the author first defined a kernel vector $\bm{\kappa}_{\mathbf{x}}(\mathbf{z})=[\kappa(\mathbf{z},\mathbf{x}_1),\cdots,\kappa(\mathbf{z},\mathbf{x}_k)] $, where $\mathbf{x}_i$ denotes any affine transform of $\mathbf{x}$, which is not limited translation. $\kappa(\cdot)$ is the kernel function the same as $\mathbf{k}$. Then, the kernel cross-correlator (response output) is defined as:
\begin{equation}
	\hat{\mathbf{r}}=\sum_{c=1}^{D}\hat{\mathbf{\bm{\kappa}}}_{\mathbf{x}^c}(\mathbf{z}^c)\odot\hat{\mathbf{w}}^{c*}~.
\end{equation}

Based on the definition above, the KCC tracker proposed their training regression equation as:
\begin{equation}\label{eqn:53}
	\mathcal{E}(\mathbf{w})=\sum_{c=1}^{D}\left\|\hat{\bm{\kappa}}_{\mathbf{x}^c}(\mathbf{x}^c)\odot\hat{\mathbf{w}}^{c*}-\hat{\mathbf{y}}\right\|_2^2+\lambda\sum_{c=1}^D\left\|\hat{\mathbf{w}}^c\right\|_2^2~.
\end{equation}

By applying partial derivative, a closed solution to Eq.~(\ref{eqn:53}) can be obtained:
\begin{equation}
\hat{\mathbf{w}}^{c*}=\frac{\hat{\mathbf{y}}\odot\hat{\bm{\kappa}}^*_{\mathbf{x}^c}(\mathbf{x}^c)}{\hat{\bm{\kappa}}_{\mathbf{x}^c}(\mathbf{x}^c)\odot\hat{\bm{\kappa}}^*_{\mathbf{x}^c}(\mathbf{x}^c)+\lambda}~.
\end{equation}

\Remark Since this result was derived without any theorems or restrictions, it is theoretically possible to predict any affine transform of $\mathbf{x}$, \emph{e.g.}, translation (the KTC tracker), scale change (the KSC tracker), and rotation (the KRC tracker) and apply any kernel function.

\section{Experimental Evaluation and Analysis }\label{Section 4}

This section exhibits the experimental results and analysis, which is divided into five subsections. Subsection~\ref{Sec:4.1} firstly introduces some implementation information, including the commonly used evaluation metrics for object tracking, the experiment platform, parameter settings, and the benchmarks used in the experiment. Secondly, a comprehensive analysis of the overall performance and the tracking results of DCF-based trackers are given in subsection~\ref{Sec:4.2}. Moreover, subsection~\ref{Sec:4.3} proposes the redefined UAV tracking attributes and analyzes the performance of DCF-based trackers by attribute. Then, in subsection~\ref{Sec:4.4}, the performance of the DCF-based tracker (only using the handcrafted feature) against the deep trackers is supplemented in the fourth subsection. Finally, on the basis of the comprehensive experiments, we conclude the typical failure cases and challenges that have not been well addressed in DCF-based methods for UAV tracking in subsection~
\ref{Sec:4.5}.

\subsection{Implementation information}\label{Sec:4.1}
Before stepping into experimental evaluation and analysis, some experiment implementation information is given firstly in this subsection, \emph{i.e.}, two metrics adopted in the experiment for trackers' evaluation, the experiment platform where all the experiments were extended, the parameter settings in the implemented code, and the UAV benchmarks utilized in the experiment.

\subsubsection{Evaluation metrics}
\begin{figure}[!b]
	\includegraphics[width=0.98\columnwidth]{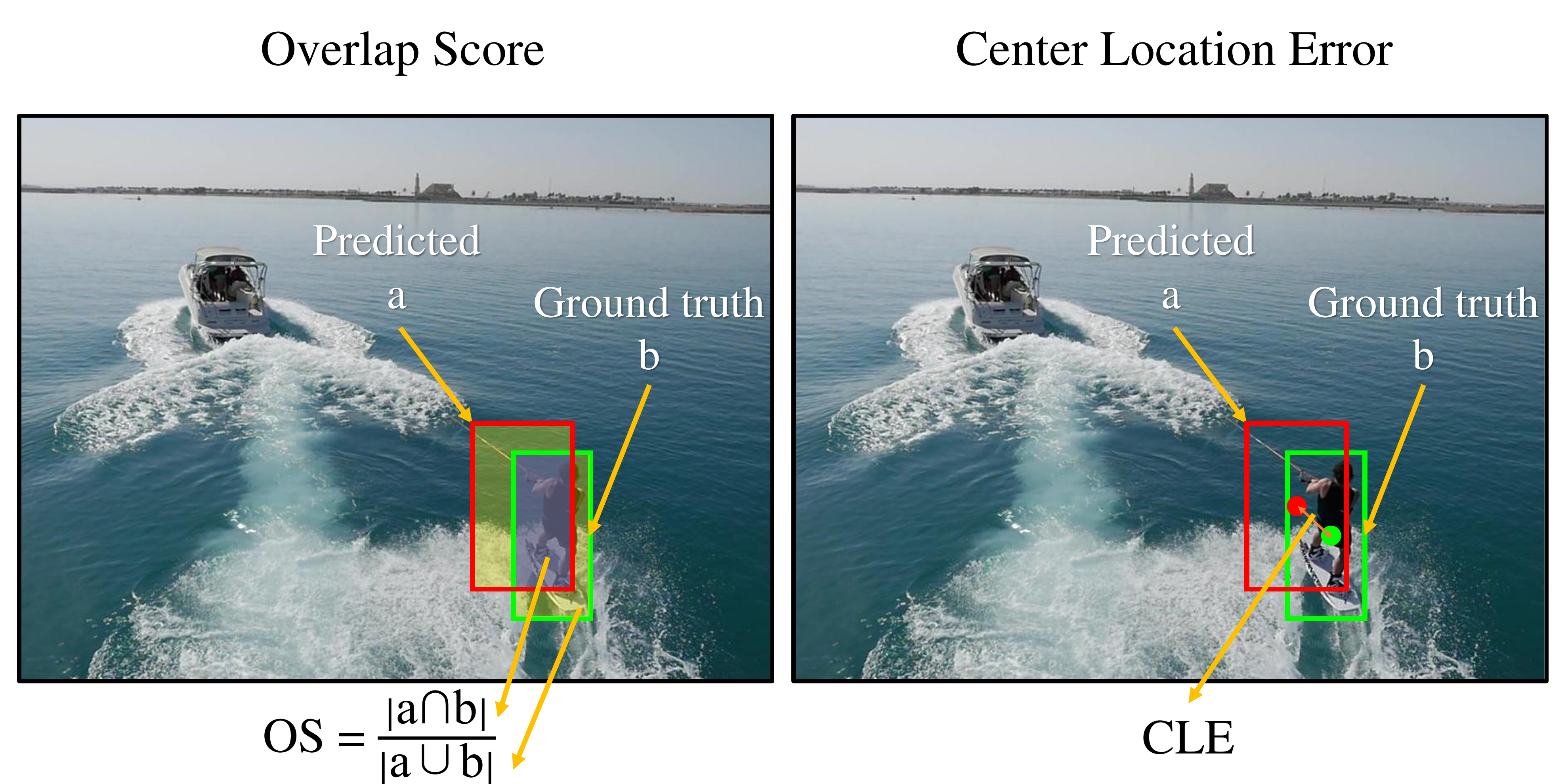}
	\caption{Two commonly used evaluation metrics in tracking: OS and CLE, and their calculation diagram. Object ground-truth and location center are marked out by \textbf{\textcolor{green}{green}} box and dot, while the predicted ones are marked out by \textbf{\textcolor{red}{red}} ones. For better display effect, please refer to the electronic version of this paper. } 
	\label{fig:3}
\end{figure}

Two authoritative and objective evaluation metrics commonly used in object tracking are introduced here, namely center location error (CLE) and overlap score (OS), based on the one-pass evaluation (OPE) \cite{wu2015TPAMI}. 

OPE refers to initializing the first frame with the location and size of the object in the ground-truth and then running the tracking algorithm to obtain the bounding boxes in subsequent frames, which can be used to draw the precision plot and success rate plot. 

To obtain the precision plot, CLE needs to be calculated in every frame which is defined by the distance between the center point of the bounding box estimated by the tracker and the center point in the ground-truth, as explained in Fig.~\ref{fig:3}. By calculating the percentage of all video frames in a sequence where the CLE is less than a given threshold, a pair of precision score and threshold is obtained. For evaluation of the whole benchmark, the final precision score can be produced by averaging scores from all sequences. Different thresholds result in different percentages, therefore a precision curve plot can be obtained, as exhibited in Fig.~\ref{fig:overall}. The threshold is set to 20 pixels during general evaluation (also in this experiment) for the final ranking of all trackers, \emph{i.e.}, distance precision (DP) at CLE = 20 pixels.

For the success rate plot, OS is firstly calculated in every frame, as shown in Fig.~\ref{fig:3}. Utilizing the bounding box obtained by the tracking algorithm (recorded as rectangle a), and the box given by ground-truth (recorded as rectangle b), OS can be obtained by:
\begin{equation}
	\text{OS}= \frac{|a \cap b|}{|a \cup b|}~,
\end{equation}

\begin{table*}[t]
	\centering
	\setlength{\tabcolsep}{2mm}
	\fontsize{6}{8}\selectfont
	\begin{threeparttable}
		\caption{Numbers of sequences, minimum, maximum, mean frames in each sequence, and total frames in 6 benchmarks, \emph{i.e.}, UAV123, UAV123@10fps, UAV20L \cite{mueller2016ECCV}, UAVDT \cite{du2018ECCV}, DTB70 \cite{li2017AAAI}, and VisDrone2019-SOT \cite{du2019ICCVW}. \textbf{\textcolor[rgb]{ 1,  0,  0}{Red}}, \textbf{\textcolor[rgb]{ 0,  1,  0}{green}}, and \textbf{\textcolor[rgb]{ 0,  0,  1}{blue}} denotes the first, second and third place respectively.}
		\vspace{0.08cm}
		\begin{tabular}{cccccc|ccc|ccc|ccc|ccc|cccc}
			%\linespread{1.5}
			\toprule[1.5pt]
			
			\multicolumn{3}{c}{Benchmark}&\multicolumn{3}{c}{UAV123}&\multicolumn{3}{c}{UAV123@10fps}&\multicolumn{3}{c}{UAV20L}&\multicolumn{3}{c}{DTB70}&\multicolumn{3}{c}{UAVDT}&\multicolumn{3}{c}{Visdrone2019-SOT}\cr
			\midrule
			
			\multicolumn{3}{c}{Sequences}&\multicolumn{3}{c}{\textbf{\textcolor{green}{123}}}&\multicolumn{3}{c}{\textbf{\textcolor{green}{123}}}&\multicolumn{3}{c}{20}&\multicolumn{3}{c}{\textbf{\textcolor{blue}{70}}}&\multicolumn{3}{c}{50}&\multicolumn{3}{c}{\textbf{\textcolor{red}{132}}}\cr
			\midrule
			
			\multicolumn{3}{c}{Each sequence}&\multirow{2}{*}{\textbf{\textcolor{green}{109}}}&\multirow{2}{*}{\textbf{\textcolor{green}{3085}}}&\multirow{2}{*}{\textbf{\textcolor{green}{915}}}&\multirow{2}{*}{37}&\multirow{2}{*}{1029}&\multirow{2}{*}{308}&\multirow{2}{*}{\textbf{\textcolor{red}{1717}}}&\multirow{2}{*}{\textbf{\textcolor{red}{5527}}}&\multirow{2}{*}{\textbf{\textcolor{red}{2934}}}&\multirow{2}{*}{68}&\multirow{2}{*}{699}&\multirow{2}{*}{225}&\multirow{2}{*}{82}&\multirow{2}{*}{2969}&\multirow{2}{*}{742}&\multirow{2}{*}{\textbf{\textcolor{blue}{90}}}&\multirow{2}{*}{\textbf{\textcolor{blue}{2970}}}&\multirow{2}{*}{\textbf{\textcolor{blue}{833}}}\cr
			\cmidrule(lr){1-3} 
			Min&Max&Mean& & & & & & & & & & & & & & & & & & \cr
			\midrule
			
			\multicolumn{3}{c}{Total frames}&\multicolumn{3}{c}{\textbf{\textcolor{red}{112578}}}&\multicolumn{3}{c}{37885}&\multicolumn{3}{c}{\textbf{\textcolor{blue}{58670}}}&\multicolumn{3}{c}{15777}&\multicolumn{3}{c}{37084}&\multicolumn{3}{c}{\textbf{\textcolor{green}{109909}}}\cr
			
			\bottomrule[1.5pt]
			\label{tab:4}
		\end{tabular}
	\end{threeparttable}
\vspace{-0.5cm}
\end{table*}
\noindent where $|\cdot|$ represents the number of pixels in the area. When the OS of a frame is greater than the set threshold, this frame is regarded as a successful one, and the percentage of the total successful frames to all frames is the success rate under one threshold. The value of OS ranges from 0 to 1, so a curve plot can be drawn, which is the success rate plot in Fig.~\ref{fig:overall}. In general evaluation (also in this experiment), the area under the curve (AUC) is calculated as the trackers' ranking standard. For the ease of understanding, Fig.~\ref{fig:3} graphically shows the two evaluation metrics.

\subsubsection{Experiment platform}
The large-scale evaluation experiments in this work were extended on MATLAB R2019a. The main hardware consists of a single Intel Core i7-8700K CPU, 32GB RAM, and an NVIDIA RTX 2080 GPU. 

\subsubsection{Parameter settings}
In order to ensure the fairness and objectivity of the experiment, all the trackers evaluated have maintained their official initial parameters. For trackers that use various features, such as the ECO tracker \cite{danelljan2017CVPR}, the ARCF tracker \cite{Huang2019ICCV}, the specific features used in the experiment are noted, \emph{e.g.}, the ARCF-H tracker utilizes fHOG feature \cite{Felzenszwalb2010TPAMI} only, and the ARCF-HC tracker uses fHOG, CN \cite{Weijer2006ECCV}, and grayscale.

\subsubsection{Benchmarks}
The experiments used a total of six well-known authoritative UAV tracking benchmarks, \emph{i.e.}, UAV123, UAV20L, UAV123@10fps~\cite{mueller2016ECCV}, UAVDT~\cite{du2018ECCV}, DTB70~\cite{li2017AAAI}, and VisDrone2019-SOT~\cite{du2019ICCVW}. Here the characteristics of each benchmark are introduced one by one.

M. Mueller \emph{et al.} \cite{mueller2016ECCV} compiled benchmark UAV123, which contains 123 fully annotated HD video sequences captured by a low-altitude aerial perspective, including totally 112,578 frames and covering a wide variety of scenes and objects. As a subset of UAV123, UAV20L is designed especially for long-term tracking, which includes 20 longest sequences. For the investigation of the impact of the camera capture speed on tracking performance, M. Mueller also temporally downsampled the UAV123 benchmark to 10 FPS where most sequences are originally provided at 30FPS, thus the benchmark UAV123@10fps were created. Note that since the frame interval becomes larger, the object's location changes between frames become bigger, increasing the difficulty of accurate tracking.

Consisting of 70 videos and totally 15,777 frames, DTB70 was built by S. Li \emph{et al.} \cite{li2017AAAI}. The highlight of DTB70 is their attention to the severe camera motion issue, which mainly focuses on tracking people and vehicles.

UAVDT \cite{du2018ECCV}, constructed by D. Du \emph{et al.}, contains 50 sequences and 37,084 frames (here refers to their single object tracking (SOT) task). UAVDT primarily focuses on cars under a variety of new challenges, \emph{e.g.}, various weather conditions, flying altitude, and camera view.

For VisDrone2019-SOT \cite{du2019ICCVW}, VisDrone2019-SOT-test-dev, VisDrone2019-SOT-val, and VisDrone2019-SOT-train are combined into totally 132 sequences and 109,909 frames. This benchmark is from the VisDrone2019 single object tracking challenge, which focused on evaluating single-object tracking algorithms on drones and was held in conjunction with the international conference on computer vision (ICCV2019). 

Table~\ref{tab:4} shows the numbers of sequences, minimum, maximum, mean frames in each sequence, and total frames in 6 benchmarks.

\subsection{Overall performance of DCF-based trackers}\label{Sec:4.2}
\begin{figure*}[!t]
	\begin{center}
		\subfigure[]{ \label{fig:UAVDT} 
			\begin{minipage}{0.315\textwidth}
				\centering
				\includegraphics[width=1\columnwidth]{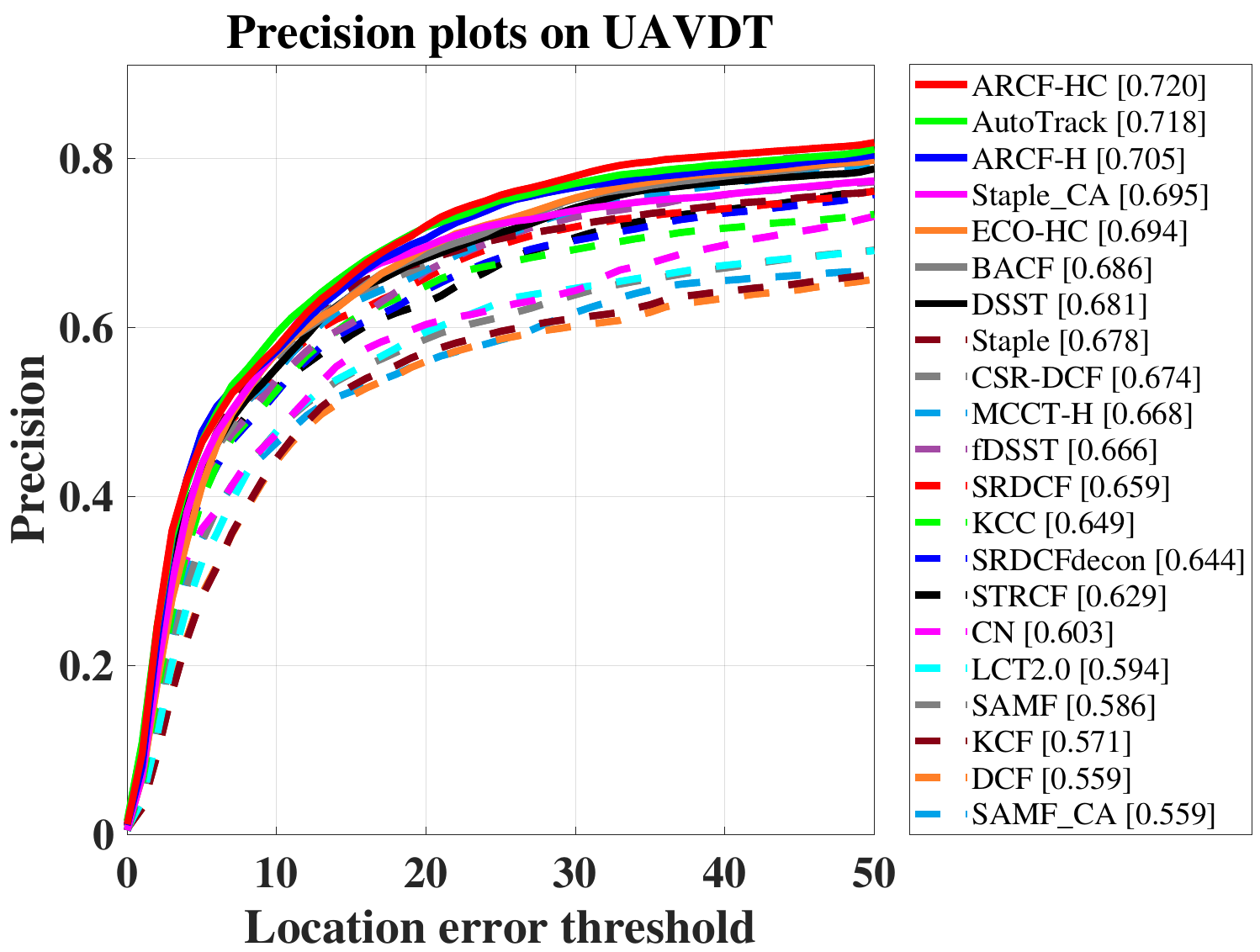}
				\\
				\includegraphics[width=1\columnwidth]{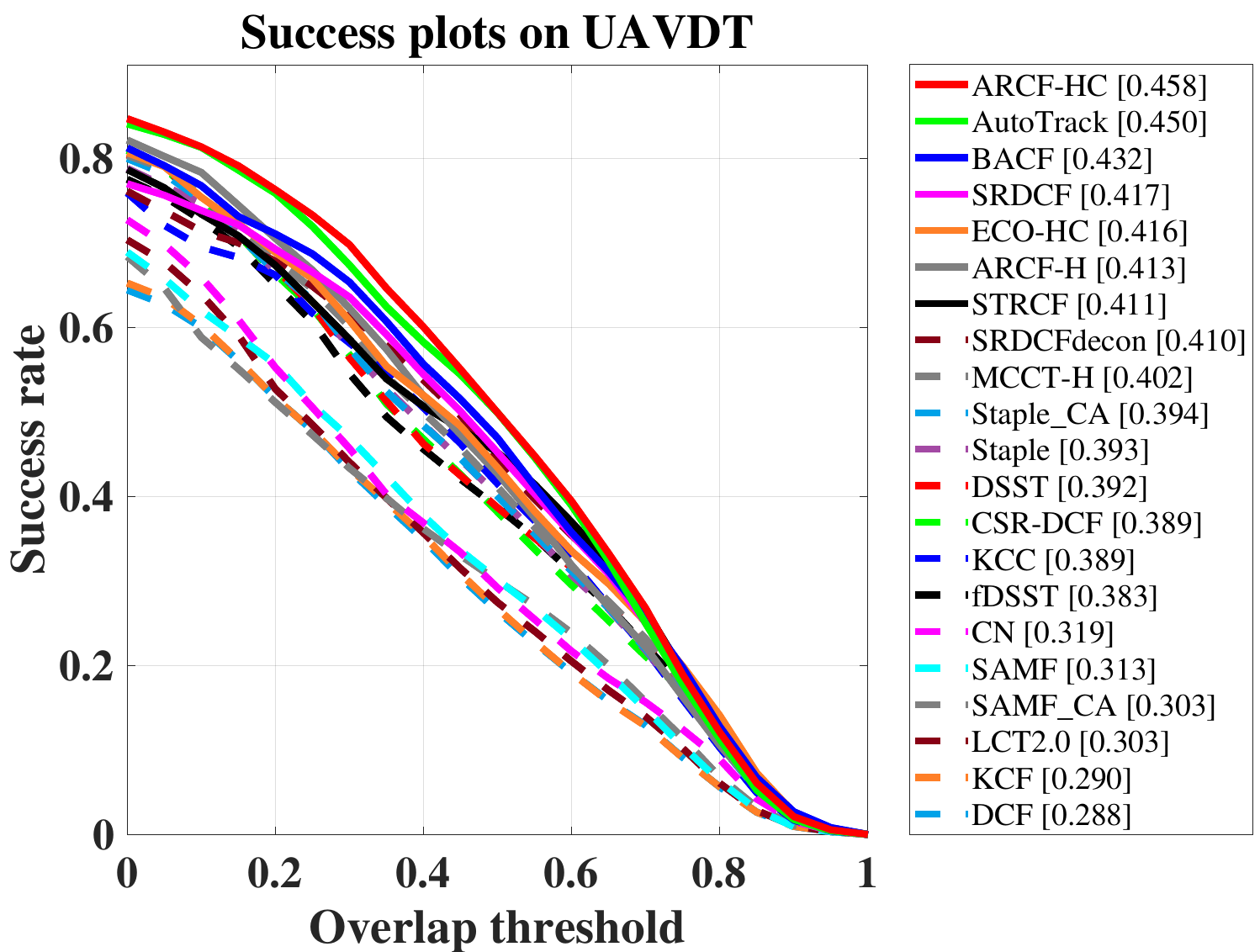}
			\end{minipage}
		}
		\subfigure[] { \label{fig:UAV123} 
			\begin{minipage}{0.315\textwidth}
				\centering
				\includegraphics[width=1\columnwidth]{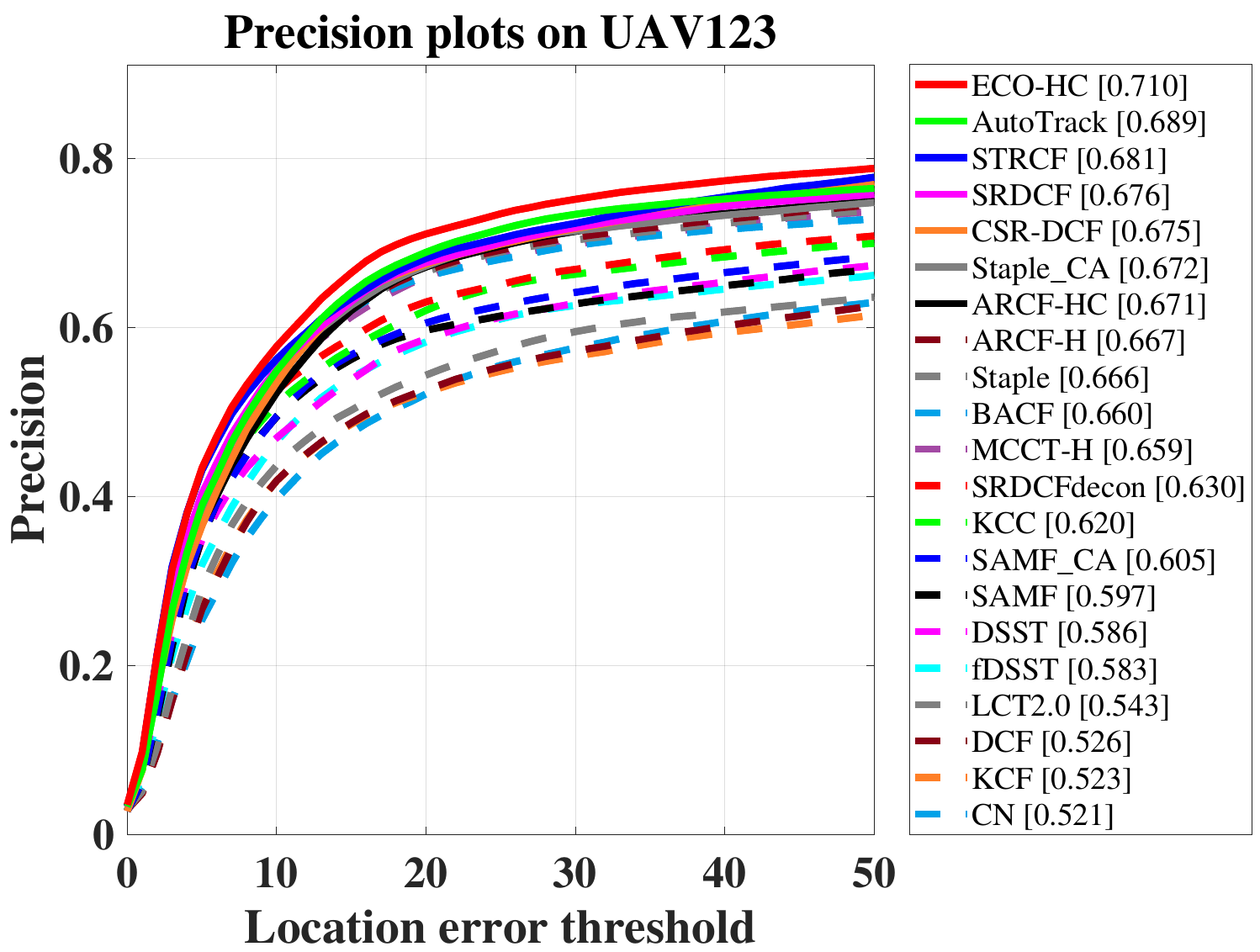}
				\\
				\includegraphics[width=1\columnwidth]{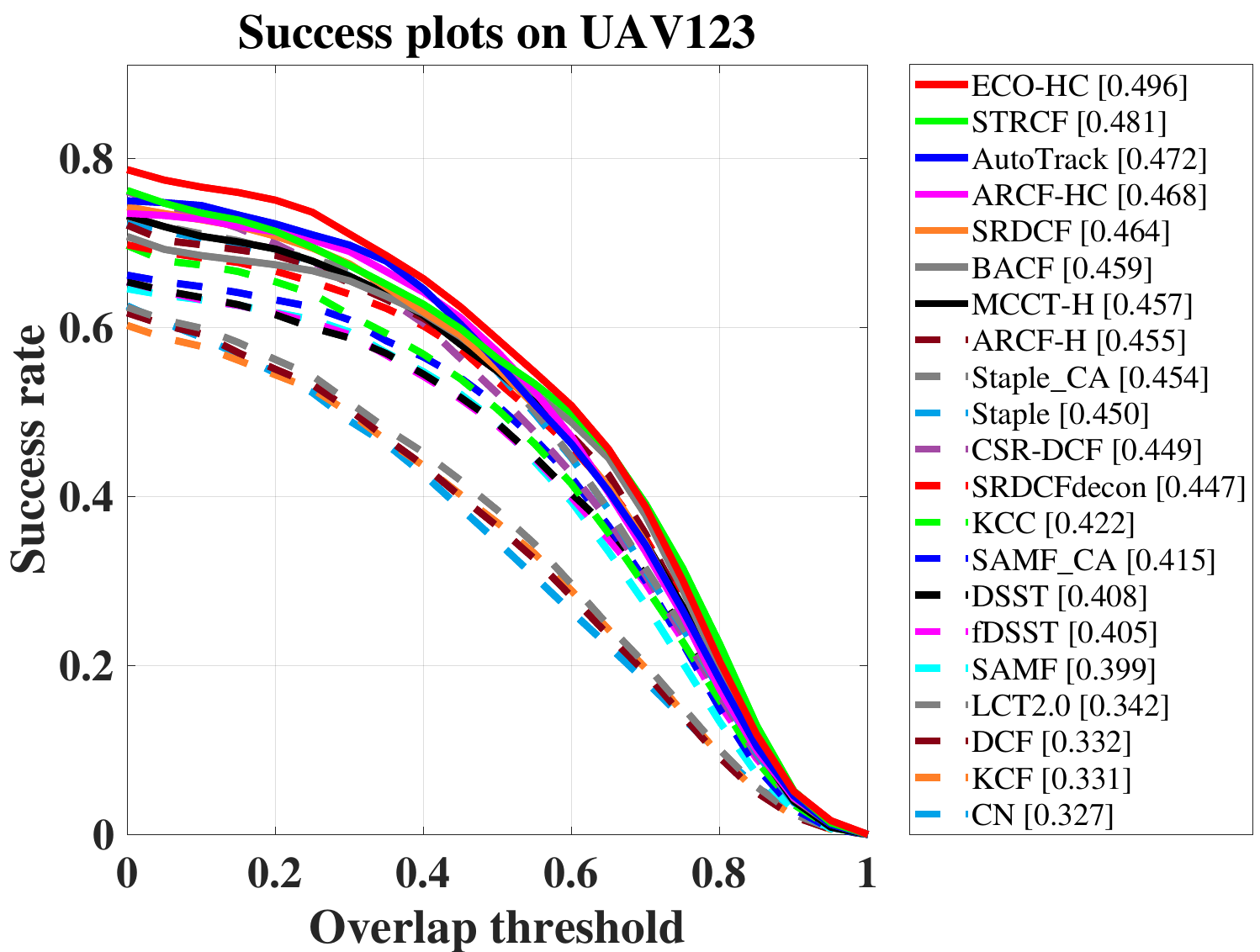}
			\end{minipage}
		}
		\subfigure[] { \label{fig:DTB70} 
			\begin{minipage}{0.315\textwidth}
				\centering
				\includegraphics[width=1\columnwidth]{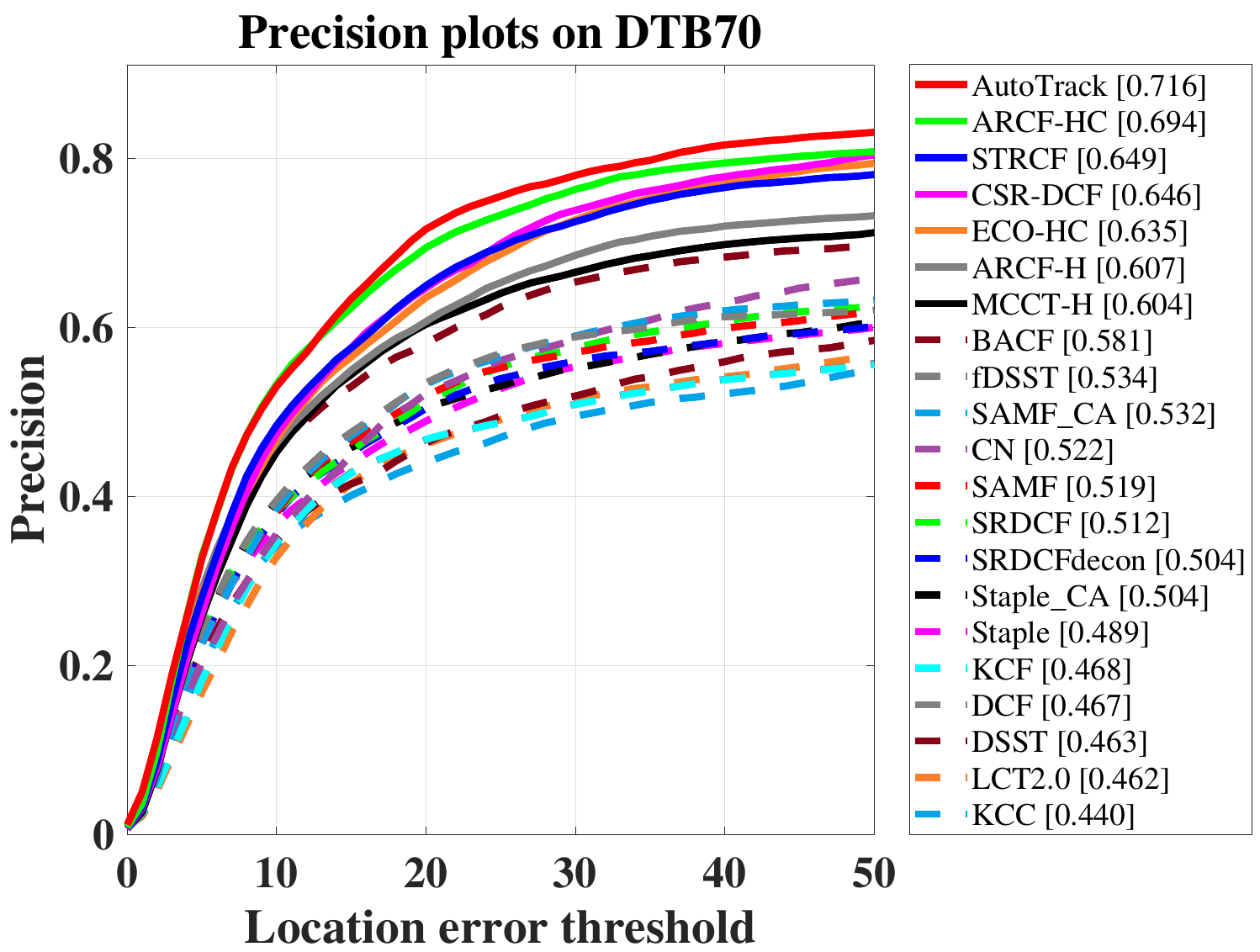}
				\\
				\includegraphics[width=1\columnwidth]{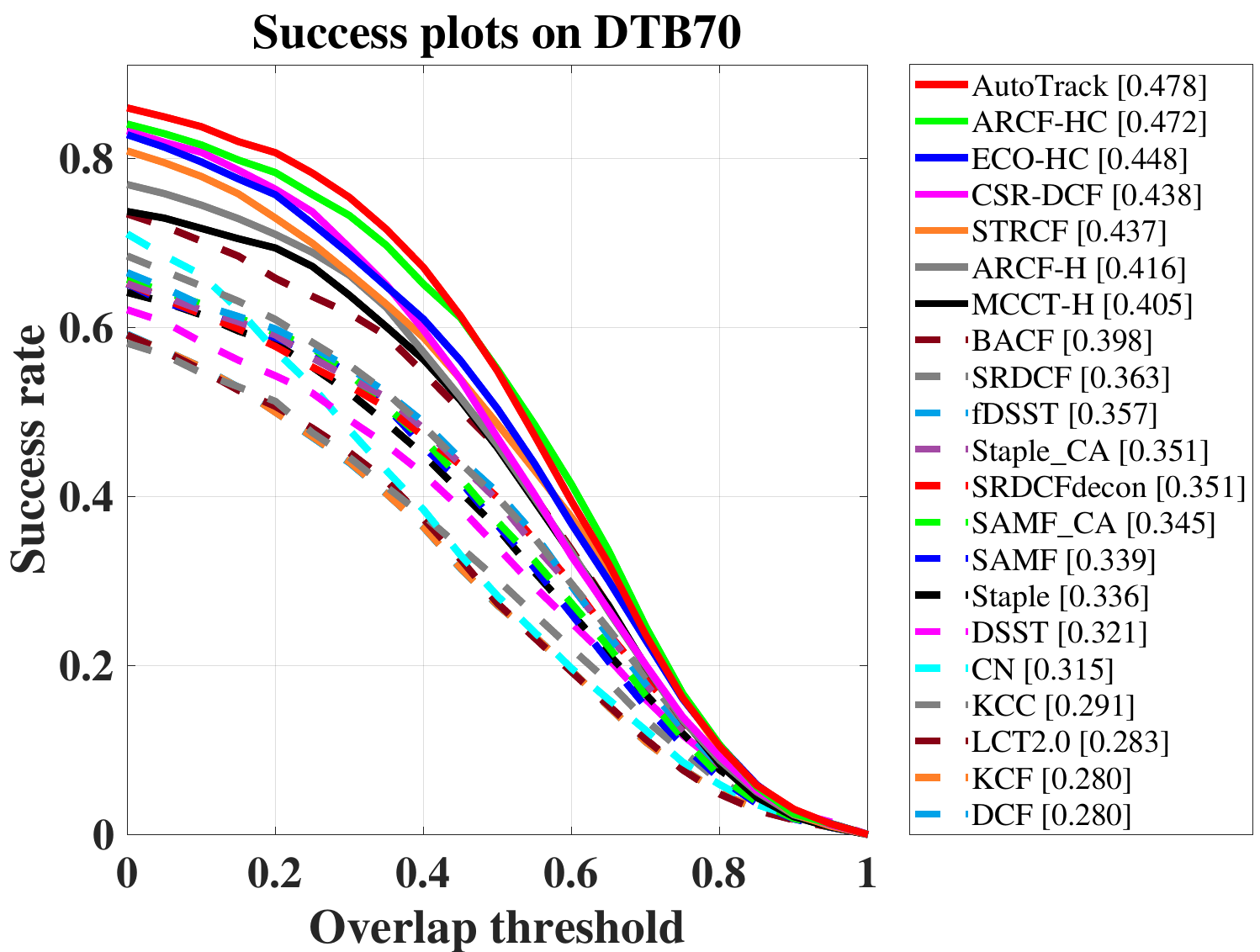}
			\end{minipage}
		}
		\\
		\subfigure[] { \label{fig:UAV123@10fps} 
			\begin{minipage}{0.315\textwidth}
				\centering
				\includegraphics[width=1\columnwidth]{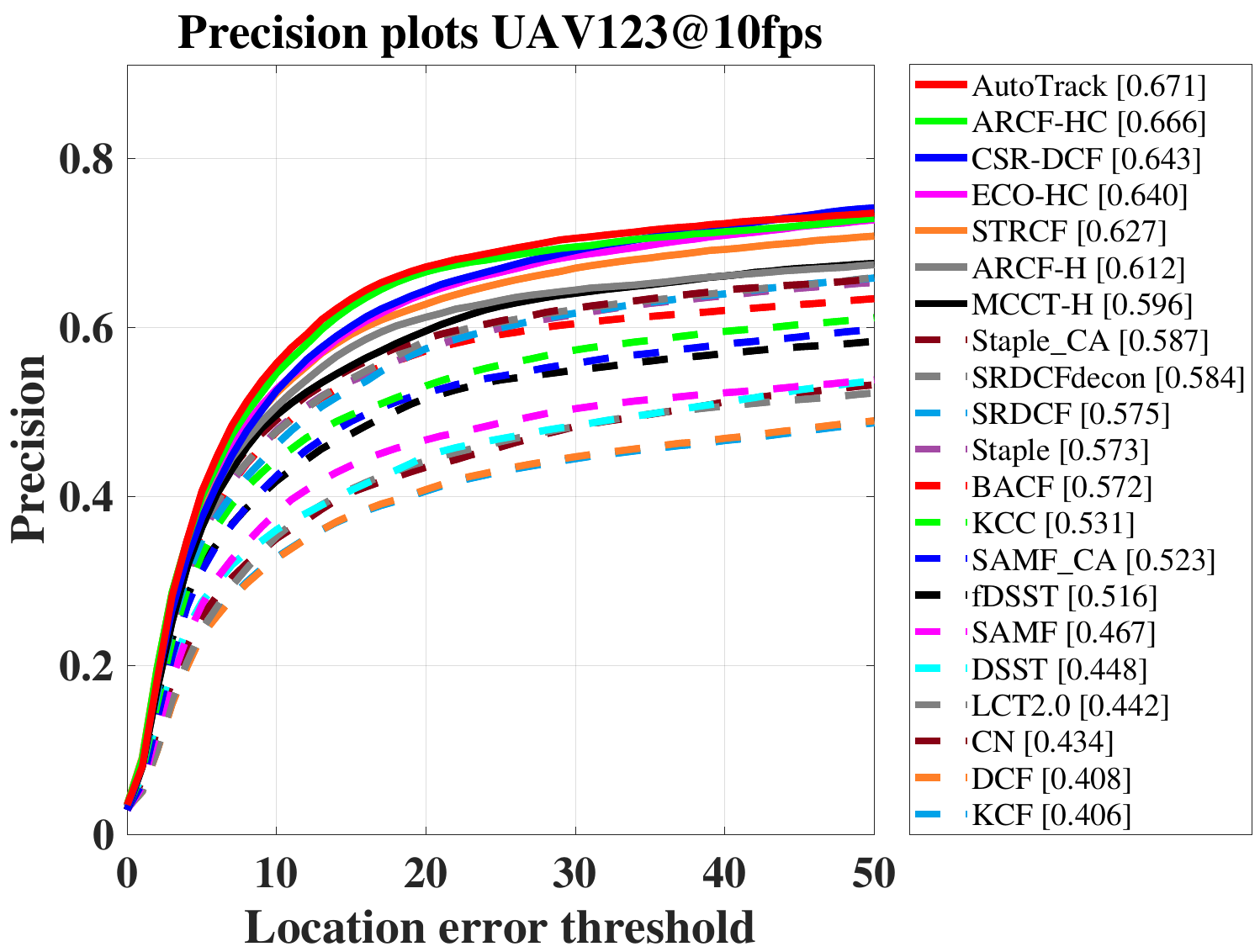}
				\\
				\includegraphics[width=1\columnwidth]{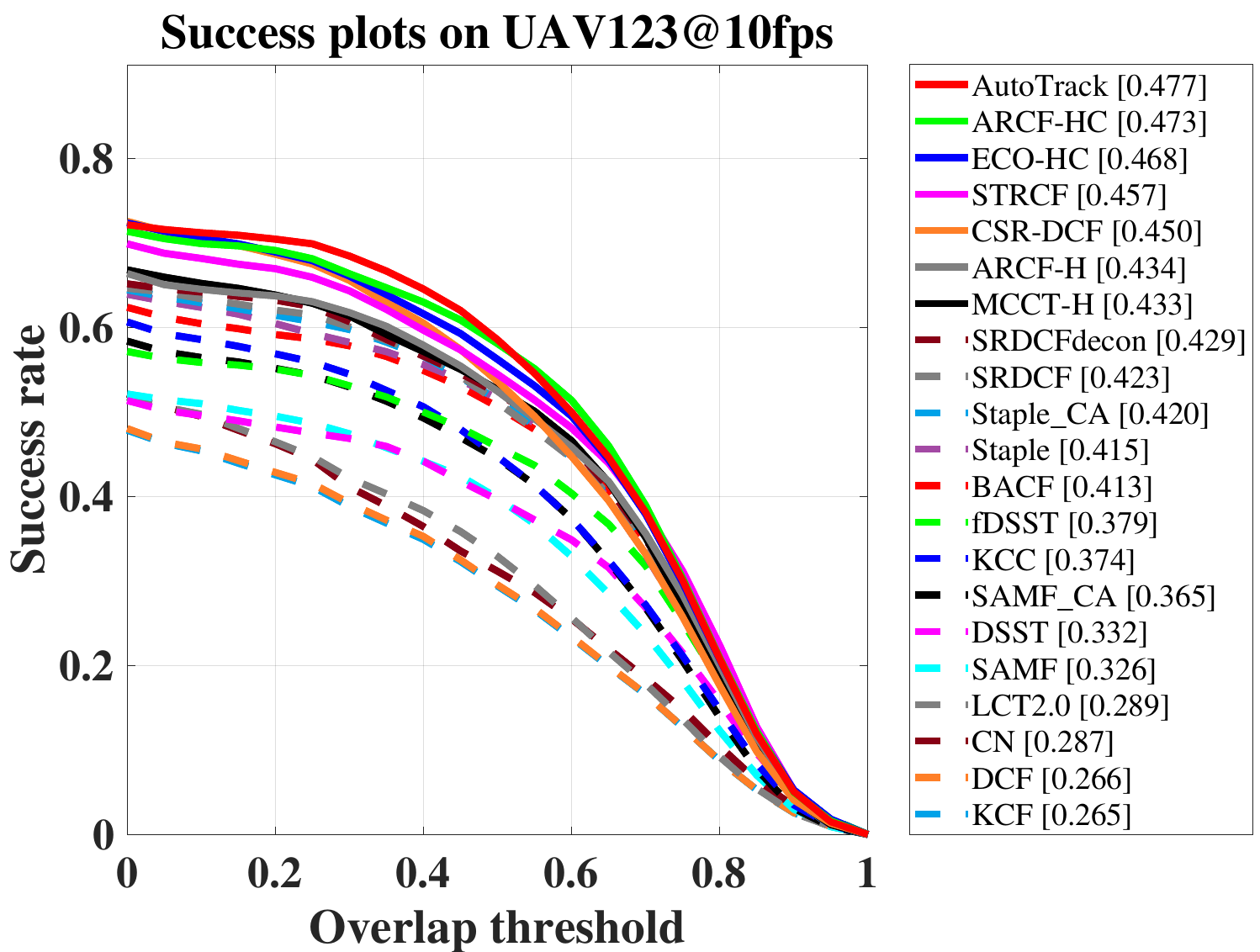}
			\end{minipage}
		}
		\subfigure[] { \label{fig:UAV20L} 
			\begin{minipage}{0.315\textwidth}
				\centering
				\includegraphics[width=1\columnwidth]{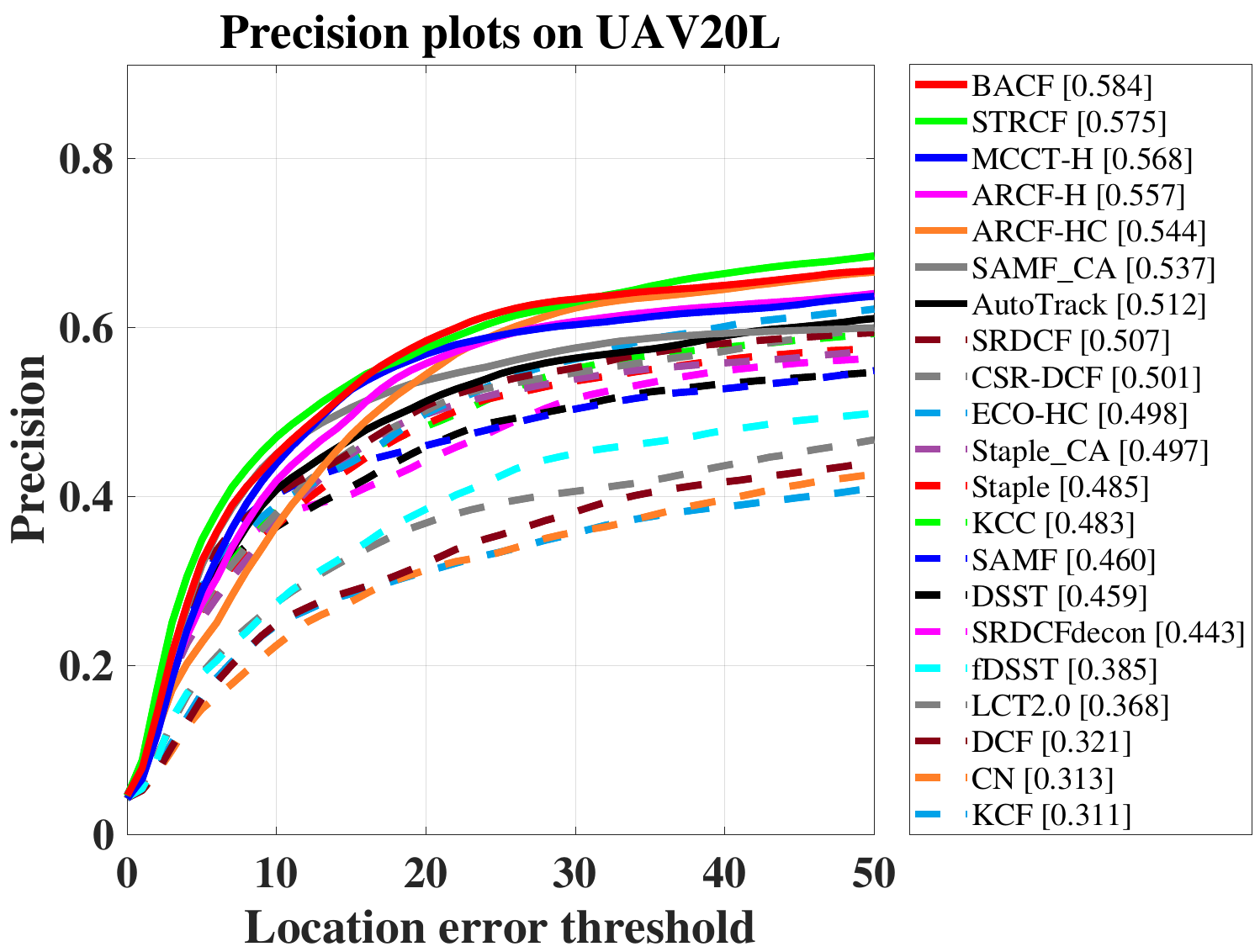}
				\\
				\includegraphics[width=1\columnwidth]{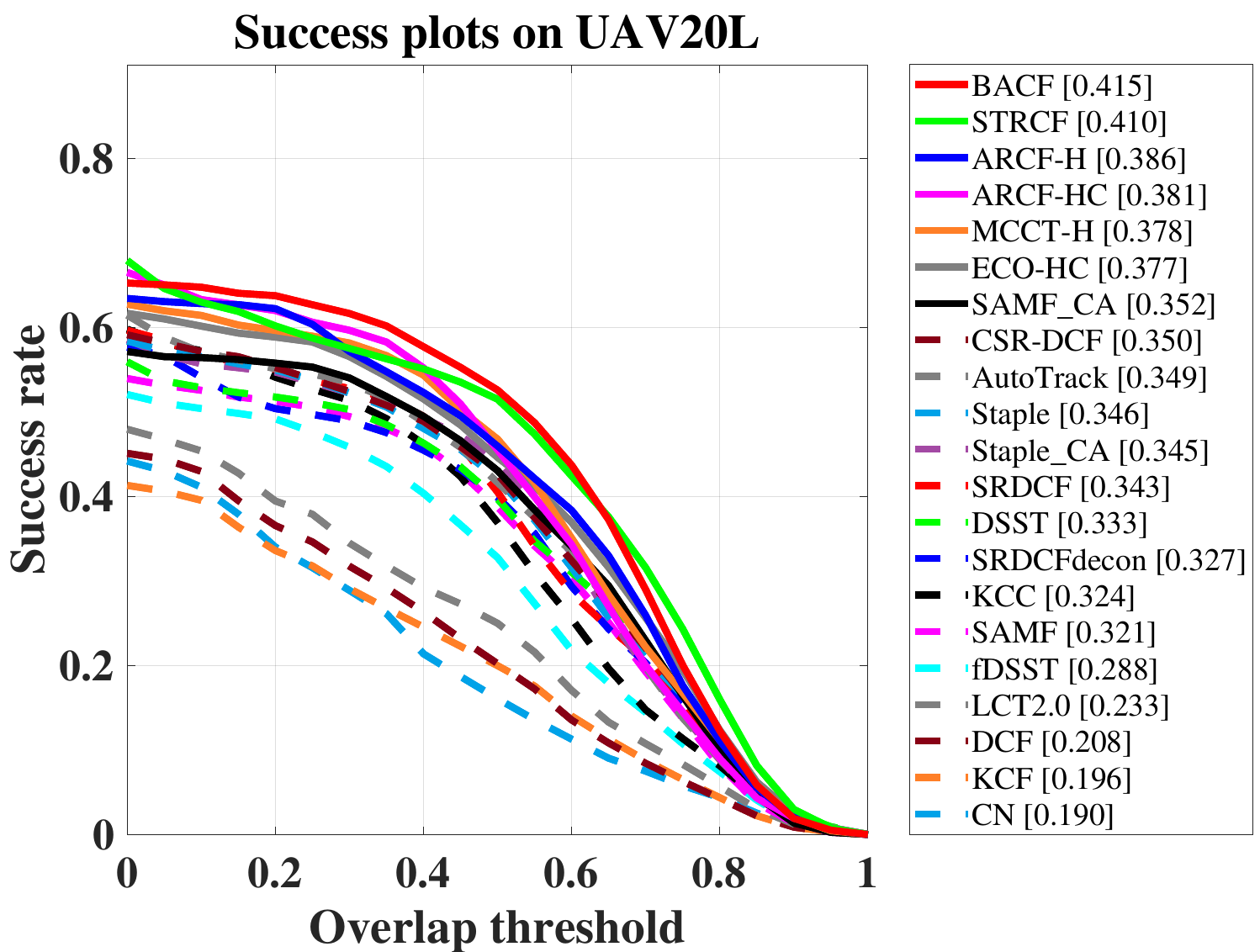}
			\end{minipage}
		}
		\subfigure[] { \label{fig:Visdrone2019} 
			\begin{minipage}{0.315\textwidth}
				\centering
				\includegraphics[width=1\columnwidth]{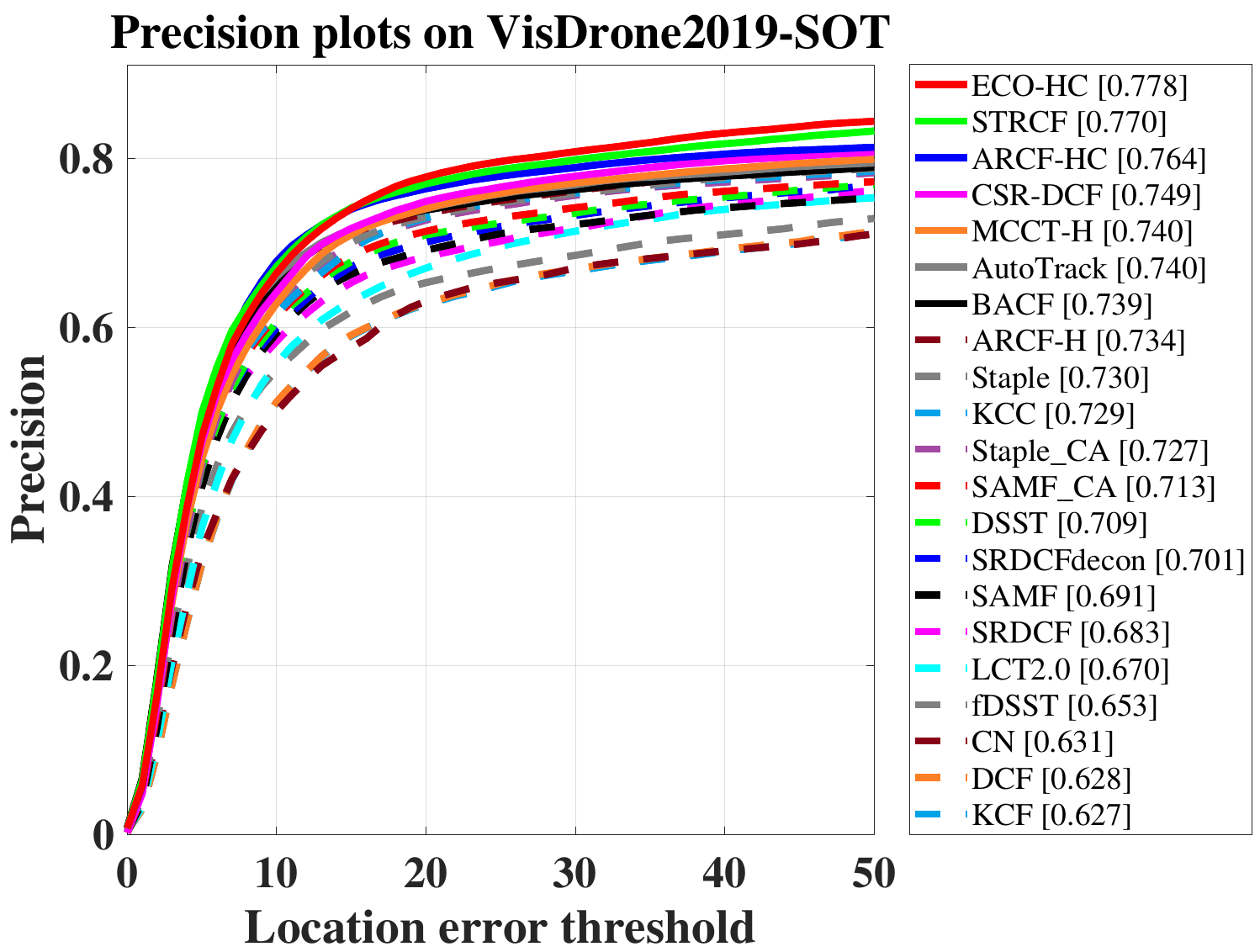}
				\\
				\includegraphics[width=1\columnwidth]{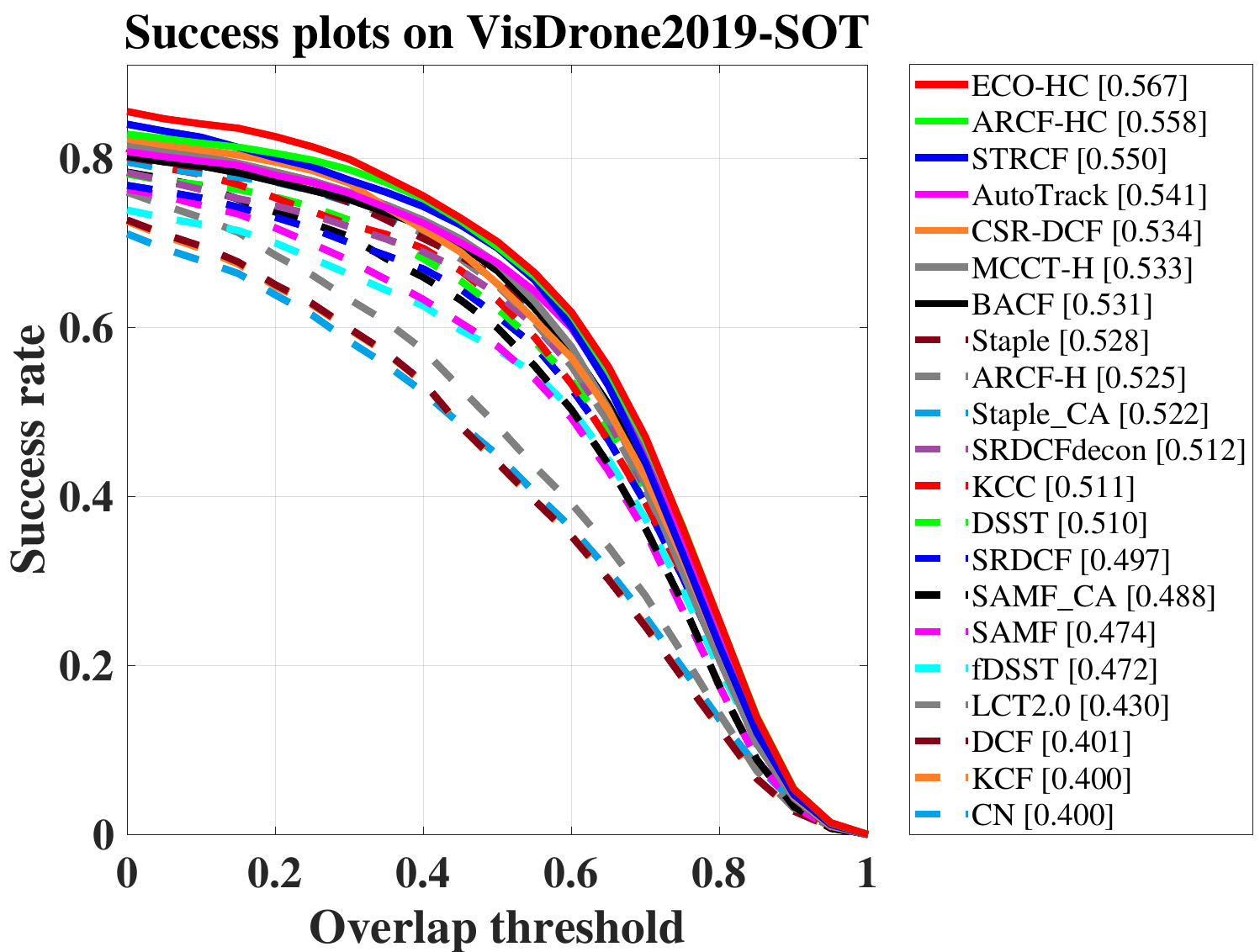}
			\end{minipage}
		}
	\end{center}
	\caption{Overall performance of handcrafted DCF-based trackers on  (a) UAVDT~\cite{du2018ECCV}, (b) UAV123~\cite{mueller2016ECCV}, (c) DTB70~\cite{li2017AAAI}, (d) UAV123@10fps~\cite{mueller2016ECCV}, (e) UAV20L~\cite{mueller2016ECCV}, and (f) Visdrone2019-SOT~\cite{du2019ICCVW}. The ranking standard in precision plot is the precision under CLE = 20 pixels, and the standard in success rate plot is the AUC (area under curve). For better display effect, please refer to the electronic version of this paper.}
	\label{fig:overall}
	%\label{fig:onecol}
\end{figure*}
In order to verify the performance of DCF-based trackers in UAV tracking scenarios, 21 famous DCF-based trackers, using only handcrafted features, are selected, \emph{i.e.}, AutoTrack \cite{Li2020CVPR}, ARCF-H \cite{Huang2019ICCV}, ARCF-HC \cite{Huang2019ICCV}, STRCF \cite{Li2018CVPR}, CSR-DCF \cite{lukezic2017CVPR}, ECO-HC \cite{danelljan2017CVPR}, MCCT-H \cite{wang2018CVPR}, BACF \cite{kiani2017ICCV}, DSST \cite{danelljan2014BMVC}, fDSST \cite{danelljan2017TPAMI}, SAMF \cite{li2014ECCV}, SDRCF \cite{Danelljan2015ICCV}, SDRCFdecon \cite{danelljan2016CVPR}, Staple \cite{bertinetto2016CVPR}, CACF (SAMF\_CA and Staple\_CA) \cite{mueller2017CVPR}, KCF (adopting Gaussian kernel) \& DCF (adopting linear kernel) \cite{Henriques2015TPAMI}, LCT2.0 \cite{ma2018IJCV}, KCC \cite{wang2018AAAI}, and CN \cite{Danelljan2014CVPR}. Their tracking results were obtained on the same platform and the same six authoritative benchmarks \cite{mueller2016ECCV}, \cite{du2018ECCV}, \cite{li2017AAAI}, \cite{du2019ICCVW}. The results show that the DCF-based trackers not only have amazing accuracy and robustness, but many of them also possess the real-time tracking speed on a single CPU, which are ideal algorithms for UAV tracking. 

As shown in Fig.~\ref{fig:overall}, the tracker performance under different benchmarks is usually different. AutoTrack performs best under DTB70, followed by ARCF-{HC}. Under UAV123, the performance of ECO-{HC} is outstanding, followed by AutoTrack. In UAVDT, ARCF-{HC} occupies the first place, followed by AutoTrack. It can be seen that with the development of research in recent years, the performance of DCF-based trackers has been incrementally improved. Table~\ref{tab:3} shows all the state-of-the-art handcrafted DCF-based (most of which achieved real-time on UAV platform) trackers' DP, AUC, and speed comparison.

\begin{table*}[t]
	\centering
	\setlength{\tabcolsep}{1.4mm}
	\fontsize{6.5}{12}\selectfont
	\begin{threeparttable}
		\caption{AUC (area under curve), DP (distance precision at CLE = 20 pixels) and speed (FPS) comparison of the handcrafted DCF-based trackers on DTB70~\cite{li2017AAAI}, UAVDT~\cite{du2018ECCV}, UAV123~\cite{mueller2016ECCV}, UAV123@10fps~\cite{mueller2016ECCV}, UAV20L~\cite{mueller2016ECCV}, and Visdrone2019-SOT~\cite{du2019ICCVW}. \textbf{\textcolor[rgb]{ 1,  0,  0}{Red}}, \textbf{\textcolor[rgb]{ 0,  1,  0}{green}}, and \textbf{\textcolor[rgb]{ 0,  0,  1}{blue}} respectively mean the first, second and third place. Note that the FPS results can vary on different platforms, for which reason the results are the average value. All the trackers maintained their official parameter during evaluation. Typical UAV tracking video can be found at \url{https://youtu.be/LPdmTjt5Nv0}.}
		\vspace{0.1cm}
		\begin{tabular}{cccccccccccccccccccc}
			%\linespread{1.5}
			\toprule[2pt]
			\multirow{2}{*}{\diagbox{Tracker}{Benchmark}}&
			\multicolumn{3}{c}{DTB70}&\multicolumn{3}{c}{UAVDT}&\multicolumn{3}{c}{UAV123}&\multicolumn{3}{c}{UAV123@10fps}&\multicolumn{3}{c}{UAV20L}&\multicolumn{3}{c}{Visdrone2019-SOT}\cr
			\cmidrule(lr){2-4} \cmidrule(lr){5-7} \cmidrule(lr){8-10} \cmidrule(lr){11-13} \cmidrule(lr){14-16} \cmidrule(lr){17-19}
			&AUC&DP&FPS&AUC&DP&FPS&AUC&DP&FPS&AUC&DP&FPS&AUC&DP&FPS&AUC&DP&FPS\cr
			
			\midrule
			
			KCF \cite{Henriques2015TPAMI} &0.280&0.468&\textbf{\textcolor{green}{364.08}}&0.290&0.571&\textbf{\textcolor{green}{826.18}}&0.331&0.523&\textbf{\textcolor{green}{611.65}}&0.265&0.406&\textbf{\textcolor{green}{561.09}}&0.196&0.311&\textbf{\textcolor{green}{371.16}}&0.400&0.627&\textbf{\textcolor{green}{379.95}}\cr
			DCF \cite{Henriques2015TPAMI}&0.280&0.467&\textbf{\textcolor{red}{552.98}}&0.288&0.559&\textbf{\textcolor{red}{1153.62}}&0.332&0.526&\textbf{\textcolor{red}{861.04}}&0.266&0.408&\textbf{\textcolor{red}{811.45}}&0.208&0.321&\textbf{\textcolor{red}{576.70}}&0.401&0.628&\textbf{\textcolor{red}{648.50}}\cr
			CN \cite{Danelljan2014CVPR}&0.315&0.522&\textbf{\textcolor{blue}{225.67}}&0.319&0.603&\textbf{\textcolor{blue}{469.80}}&0.327&0.521&\textbf{\textcolor{blue}{296.60}}&0.287&0.434&\textbf{\textcolor{blue}{308.80}}&0.190&0.313&\textbf{\textcolor{blue}{203.91}}&0.400&0.631&\textbf{\textcolor{blue}{225.67}}\cr
			SAMF \cite{li2014ECCV}&0.339&0.519&7.57&0.313&0.586&12.46&0.399&0.597&10.05&0.326&0.467&9.72&0.321&0.460&9.94&0.474&0.691&6.24\cr
			DSST\cite{danelljan2014BMVC} &0.321&0.463&61.78&0.392&0.681&127.28&0.408&0.586&85.43&0.332&0.448&84.87&0.333&0.459&58.26&0.510&0.709&67.83\cr
			SRDCF \cite{Danelljan2015ICCV}&0.363&0.512&8.41&0.417&0.659&13.11&0.464&0.676&11.08&0.423&0.575&11.18&0.343&0.507&7.53&0.497&0.683&8.22\cr
			SRDCFdecon \cite{danelljan2016CVPR}&0.351&0.504&4.18&0.410&0.644&6.09&0.447&0.630&5.20&0.429&0.584&5.44&0.327&0.443&3.98&0.512&0.701&4.16\cr
			Staple \cite{bertinetto2016CVPR}&0.336&0.489&91.04&0.393&0.678&96.36&0.450&0.666&86.91&0.415&0.573&88.01&0.346&0.485&80.08&0.528&0.730&85.08\cr
			fDSST\cite{danelljan2017TPAMI} &0.357&0.534&138.75&0.383&0.666&223.08&0.405&0.583&162.59&0.379&0.516&160.63&0.288&0.385&88.68&0.472&0.653&135.00\cr
			ECO-HC \cite{danelljan2017CVPR}&\textbf{\textcolor{blue}{0.448}}&0.635&51.86&0.416&0.694&70.04&\textbf{\textcolor{red}{0.496}}&\textbf{\textcolor{red}{0.710}}&63.34&\textbf{\textcolor{blue}{0.468}}&0.640&55.07&0.377&0.499&51.83&\textbf{\textcolor{red}{0.567}}&\textbf{\textcolor{red}{0.778}}&56.16\cr
			CACF (SAMF\_CA) \cite{mueller2017CVPR}&0.345&0.532&7.05&0.303&0.559&11.95&0.415&0.605&9.28&0.365&0.523&8.70&0.352&0.537&8.65&0.488&0.713&5.86\cr
			CACF (Staple\_CA) \cite{mueller2017CVPR}&0.351&0.504&50.66&0.394&0.695&55.50&0.454&0.672&51.62&0.420&0.587&50.12&0.345&0.497&36.92&0.522&0.727&45.86\cr
			BACF \cite{kiani2017ICCV}&0.398&0.581&37.67&\textbf{\textcolor{blue}{0.432}}&0.686&56.07&0.459&0.660&43.45&0.413&0.572&40.99&\textbf{\textcolor{red}{0.415}}&\textbf{\textcolor{red}{0.584}}&32.00&0.531&0.739&37.53\cr
			CSR-DCF \cite{lukezic2017CVPR}&0.438&0.646&11.14&0.389&0.674&12.38&0.449&0.675&10.82&0.450&\textbf{\textcolor{blue}{0.643}}&10.55&0.350&0.501&9.53&0.534&0.749&7.65\cr
			KCC \cite{wang2018AAAI} &0.291&0.440&35.93&0.389&0.649&50.78&0.422&0.620&36.42&0.374&0.531&36.58&0.324&0.483&29.16&0.511&0.729&27.73\cr
			MCCT-H \cite{wang2018CVPR}&0.405&0.604&48.91&0.402&0.668&51.88&0.457&0.659&51.90&0.433&0.596&46.37&0.378&\textbf{\textcolor{blue}{0.568}}&41.16&0.533&0.740&45.15\cr
			STRCF \cite{Li2018CVPR}&0.437&\textbf{\textcolor{blue}{0.649}}&21.88&0.411&0.629&28.64&\textbf{\textcolor{green}{0.481}}&\textbf{\textcolor{blue}{0.681}}&22.58&0.457&0.627&22.40&\textbf{\textcolor{green}{0.410}}&\textbf{\textcolor{green}{0.575}}&17.41&\textbf{\textcolor{blue}{0.550}}&\textbf{\textcolor{green}{0.770}}&20.81\cr
			LCT2.0 \cite{ma2018IJCV}&0.283&0.462&33.60&0.303&0.594&47.78&0.342&0.543&42.68&0.289&0.442&44.08&0.233&0.368&36.36&0.430&0.670&33.16\cr
			ARCF-H \cite{Huang2019ICCV}&0.416&0.607&37.07&0.413&\textbf{\textcolor{blue}{0.705}}&51.63&0.455&0.667&40.36&0.434&0.612&41.95&\textbf{\textcolor{blue}{0.386}}&0.557&31.78&0.525&0.734&37.33\cr
			ARCF-HC \cite{Huang2019ICCV}&\textbf{\textcolor{green}{0.472}}&\textbf{\textcolor{green}{0.694}}&24.28&\textbf{\textcolor{red}{0.458}}&\textbf{\textcolor{red}{0.720}}&29.39&0.468&0.671&24.50&\textbf{\textcolor{green}{0.473}}&\textbf{\textcolor{green}{0.666}}&24.09&0.381&0.544&21.92&\textbf{\textcolor{green}{0.558}}&\textbf{\textcolor{blue}{0.764}}&23.89\cr
			AutoTrack \cite{Li2020CVPR}&\textbf{\textcolor{red}{0.478}}&\textbf{\textcolor{red}{0.716}}&48.63&\textbf{\textcolor{green}{0.450}}&\textbf{\textcolor{green}{0.718}}&56.38&\textbf{\textcolor{blue}{0.472}}&\textbf{\textcolor{green}{0.689}}&48.21&\textbf{\textcolor{red}{0.477}}&\textbf{\textcolor{red}{0.671}}&47.61&0.349&0.512&44.82&0.541&0.740&45.38\cr
			
			\bottomrule[2pt]
			\label{tab:3}
		\end{tabular}
	\end{threeparttable}
\end{table*}

\begin{table*}[!t]
	\centering
	\setlength{\tabcolsep}{6mm}
	\fontsize{6}{8}\selectfont
	\begin{threeparttable}
		\caption{Correspondence between the original attributes (under the line) in the six benchmarks and the new attributes, and the serial number contribution of each benchmark to each new attribute. The attributes' full names are listed in the passage, \emph{e.g.}, CM for camera motion, VC for viewpoint change, \emph{etc.}}
		\vspace{0.08cm}
		\begin{tabular}{ccccccccc}
			\toprule[1.5pt]
			\diagbox{Benchmark}{Attributes} & \multicolumn{2}{c}{VC}& FM & LR & \multicolumn{2}{c}{OCC}& IV \cr
			\midrule
			
			\multirow{2}{*}{UAV123} & \multicolumn{2}{c}{70} & 28 & 30 & \multicolumn{2}{c}{54} & 73\cr
			\cmidrule(lr){2-3} \cmidrule(lr){4-4} \cmidrule(lr){5-5} \cmidrule(lr){6-7} \cmidrule(lr){8-8} 
			&CM&VC&FM&LR&POC&FOC&IV\cr
			\midrule
			
			\multirow{2}{*}{UAV123@10fps} & \multicolumn{2}{c}{70} & 28 & 30 & \multicolumn{2}{c}{54} & 73\cr
			\cmidrule(lr){2-3} \cmidrule(lr){4-4} \cmidrule(lr){5-5} \cmidrule(lr){6-7} \cmidrule(lr){8-8} 
			&CM&VC&FM&LR&POC&FOC&IV\cr
			\midrule
			
			\multirow{2}{*}{UAV20L} & \multicolumn{2}{c}{16} & 7 & 13 & \multicolumn{2}{c}{13} & 18\cr
			\cmidrule(lr){2-3} \cmidrule(lr){4-4} \cmidrule(lr){5-5} \cmidrule(lr){6-7} \cmidrule(lr){8-8} 
			&CM&VC&FM&LR&POC&FOC&IV\cr
			\midrule
			
			\multirow{2}{*}{Visdrone2019-SOT} & \multicolumn{2}{c}{109} & 28 & 20 & \multicolumn{2}{c}{79} &55\cr
			\cmidrule(lr){2-3} \cmidrule(lr){4-4} \cmidrule(lr){5-5} \cmidrule(lr){6-7} \cmidrule(lr){8-8} 
			&CM&VC&FM&LR&POC&FOC&IV\cr
			\midrule
			
			\multirow{2}{*}{DTB70} & \multicolumn{2}{c}{41} & \multirow{2}{*}{\XSolid} & \multirow{2}{*}{\XSolid} & \multicolumn{2}{c}{17} & \multirow{2}{*}{\XSolid} \cr
			\cmidrule(lr){2-3} \cmidrule(lr){6-7} 
			&\multicolumn{2}{c}{FCM}& & &\multicolumn{2}{c}{OCC}& \cr
			\midrule
			
			\multirow{2}{*}{UAVDT} & \multicolumn{2}{c}{30} & \multirow{2}{*}{\XSolid} & 23 & \multicolumn{2}{c}{20} & 28 \cr
			\cmidrule(lr){2-3} \cmidrule(lr){5-5} \cmidrule(lr){6-7} \cmidrule(lr){8-8}
			&\multicolumn{2}{c}{CR}& &SO &\multicolumn{2}{c}{LO}&IV \cr
			\midrule
			
			Total sequences & \multicolumn{2}{c}{336} & 91 & 116 & \multicolumn{2}{c}{237} & 247\cr
			
			\bottomrule[1.5pt]			
			\label{tab:5}
		\end{tabular}
	\end{threeparttable}
\vspace{-0.5cm}
\end{table*}

From the results in Table~\ref{tab:3}, it can be seen that the early DCF-based trackers, \emph{e.g.}, the KCF tracker \cite{Henriques2015TPAMI}, the CN tracker \cite{Danelljan2014CVPR}, due to their simplicity, usually possess high tracking speed, but this results in poor accuracy and robustness. As the development of DCF-based trackers went on, when the various problems and shortcomings of early DCF-based trackers, such as the lack of scale estimation, were paid attention to, settled, and further improved one by one, the tracking results of the DCF-based methods have also been significantly enhanced. To name a few, after the DSST \cite{danelljan2014BMVC} tracker and the SAMF \cite{li2014ECCV} tracker solved the scale estimation problem for the KCF tracker \cite{Henriques2015TPAMI}, the tracking accuracy and robustness have been boosted. After the CACF tracker (Staple\_CA) \cite{mueller2017CVPR} added a context-aware strategy, the performance is greatly better compared with the original tracker Staple \cite{bertinetto2016CVPR}. The AutoTrack tracker \cite{Li2020CVPR}, which uses the STRCF tracker \cite{Li2018CVPR} as the baseline, achieves better results than the STRCF tracker in most benchmarks through the spatio-temporal regularization term that are automatically and adaptively updated.

Among all the DCF-based trackers using handcrafted features, in the case of unadjusted parameters, three trackers, \emph{i.e.}, the AutoTrack tracker \cite{Li2020CVPR}, the ARCF-HC tracker \cite{Huang2019ICCV}, and the ECO-HC tracker \cite{danelljan2017CVPR} have achieved more top-three value (which are marked out by special colors, red, green, and blue in Table~\ref{tab:3}) in terms of the success rate (AUC) and precision (DP) in the six benchmarks than others. This indicates that even with the fixed-parameter, they still possess good universality in various complex scenes in each benchmark, which further demonstrates their practicability in real-world UAV tracking.

\subsection{Performance analysis by UAV special attributes}\label{Sec:4.3}
\begin{figure*}[!t]
	\begin{center}
		\subfigure { \label{fig:VC} 
			\begin{minipage}{0.315\textwidth}
				\centering
				\includegraphics[width=1\columnwidth]{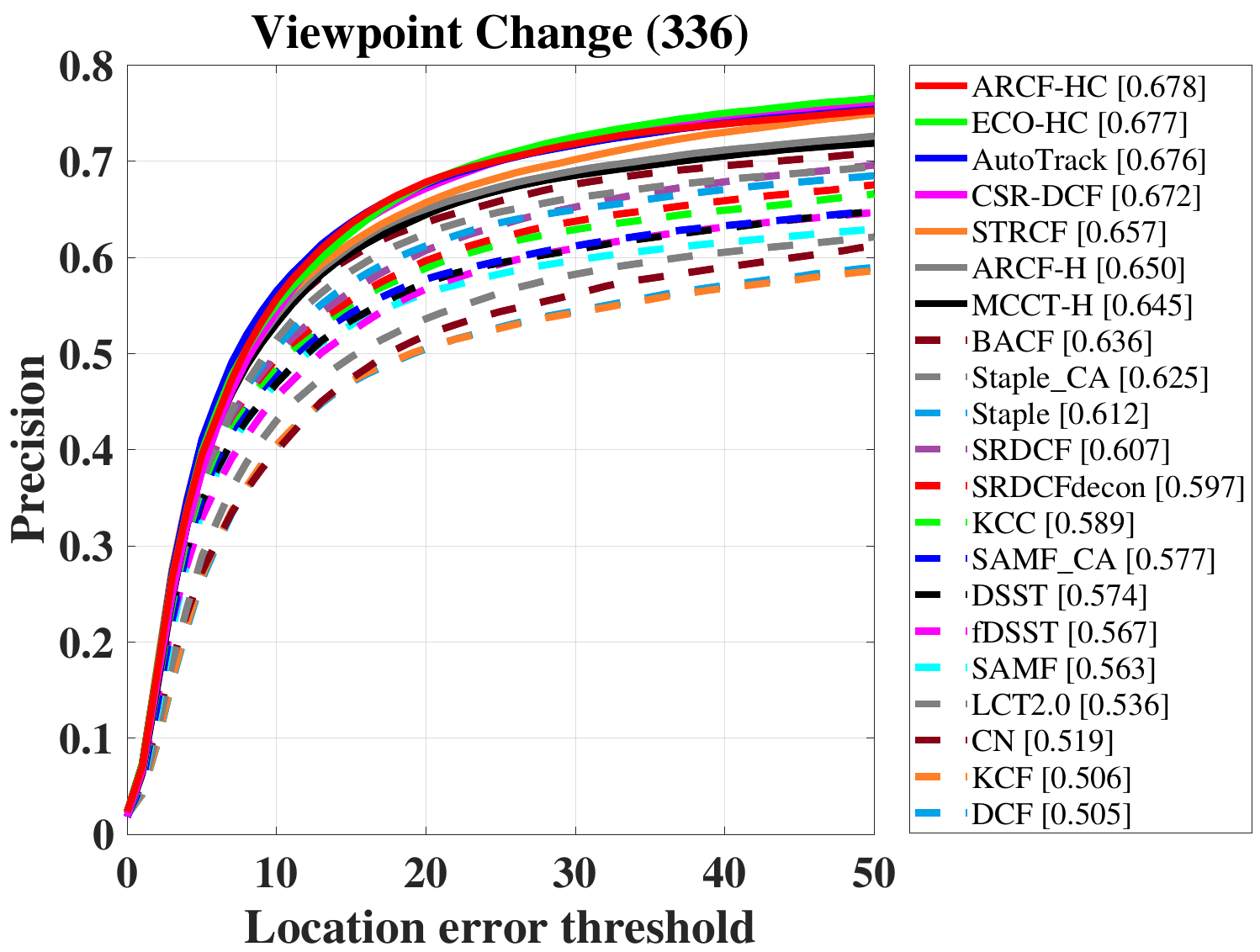}
				\includegraphics[width=1\columnwidth]{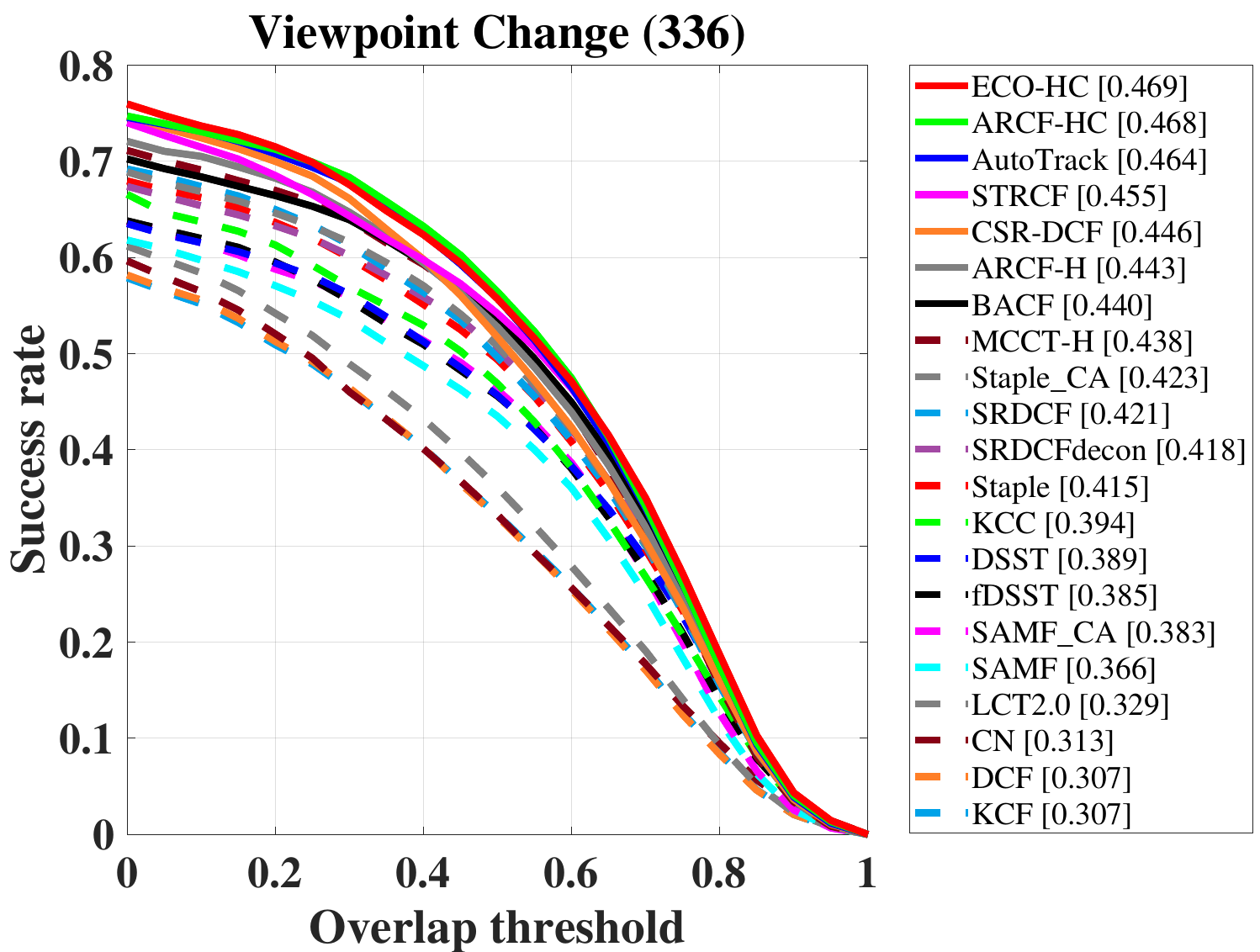}
			\end{minipage}
		}
		\subfigure { \label{fig:FM} 
			\begin{minipage}{0.315\textwidth}
				\centering
				\includegraphics[width=1\columnwidth]{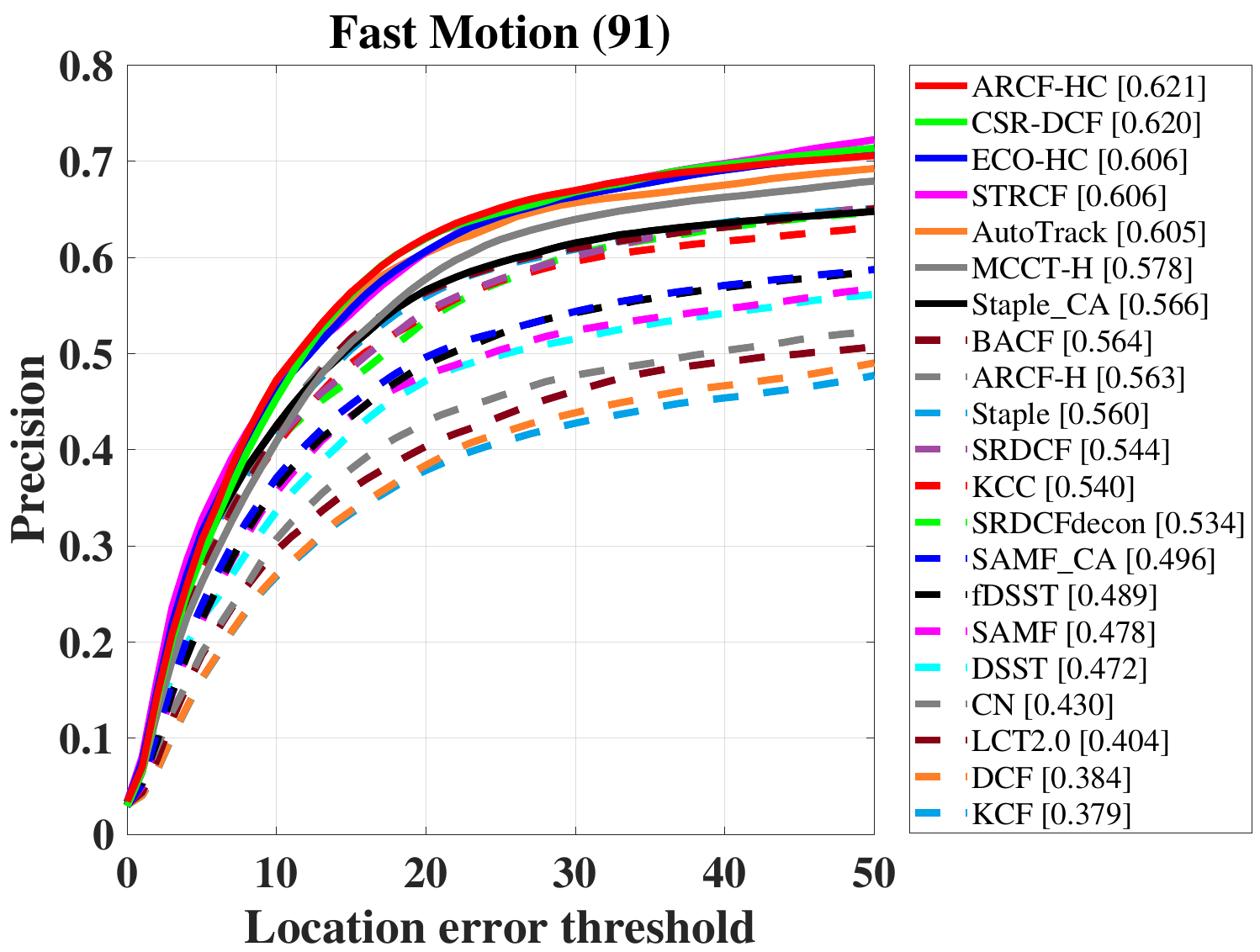}
				\includegraphics[width=1\columnwidth]{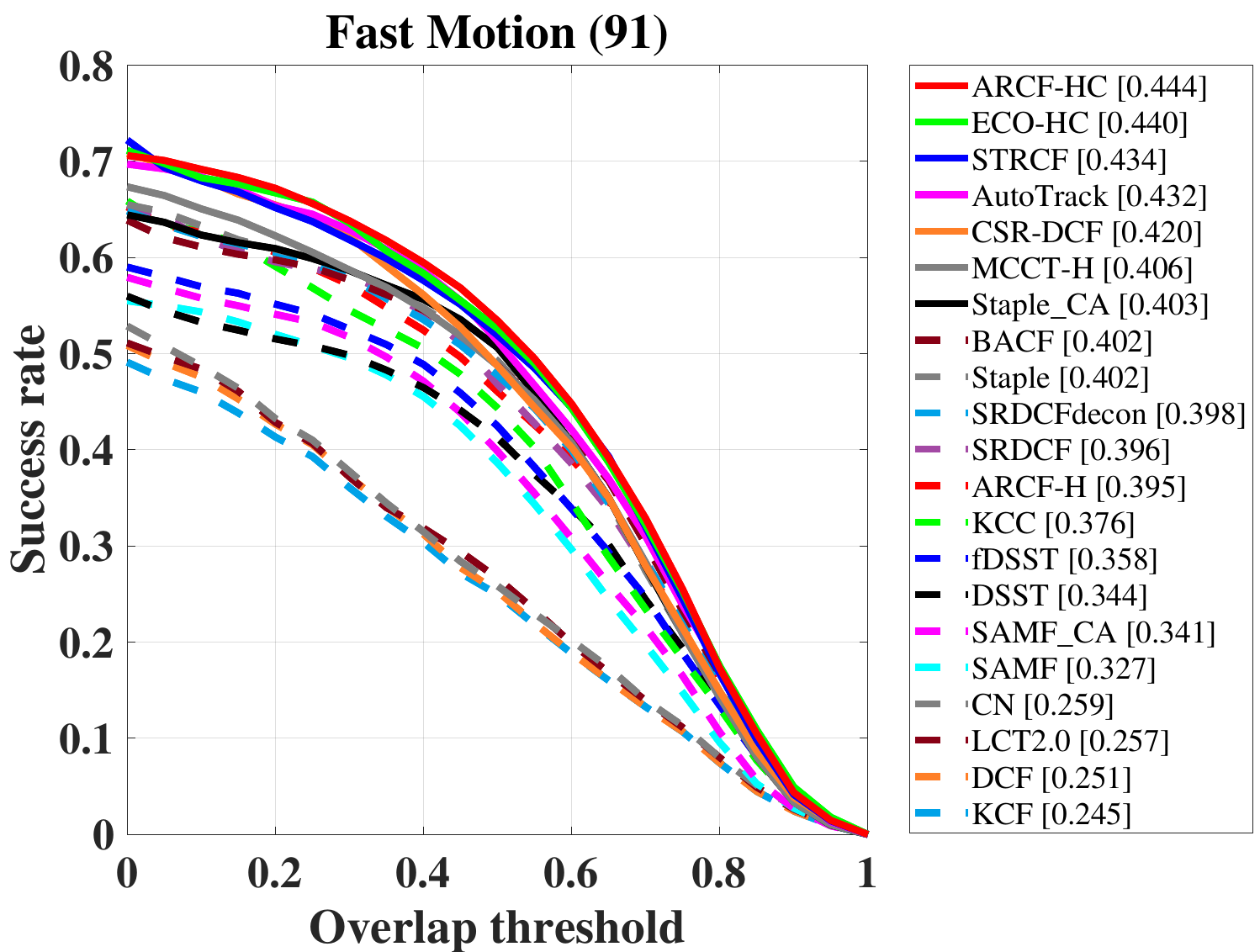}
			\end{minipage}
		}
		\subfigure { \label{fig:LR} 
			\begin{minipage}{0.315\textwidth}
				\centering
				\includegraphics[width=1\columnwidth]{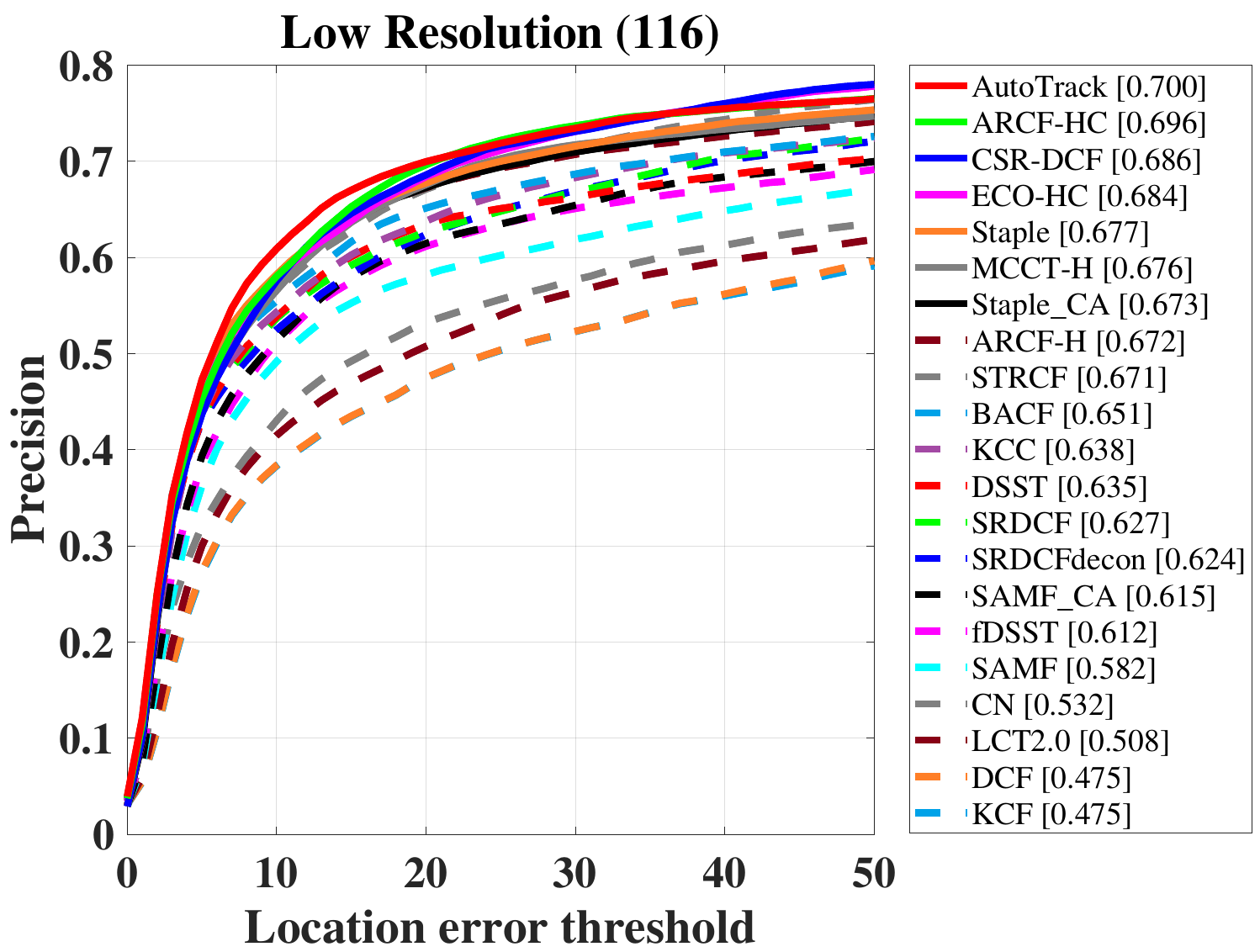}
				\includegraphics[width=1\columnwidth]{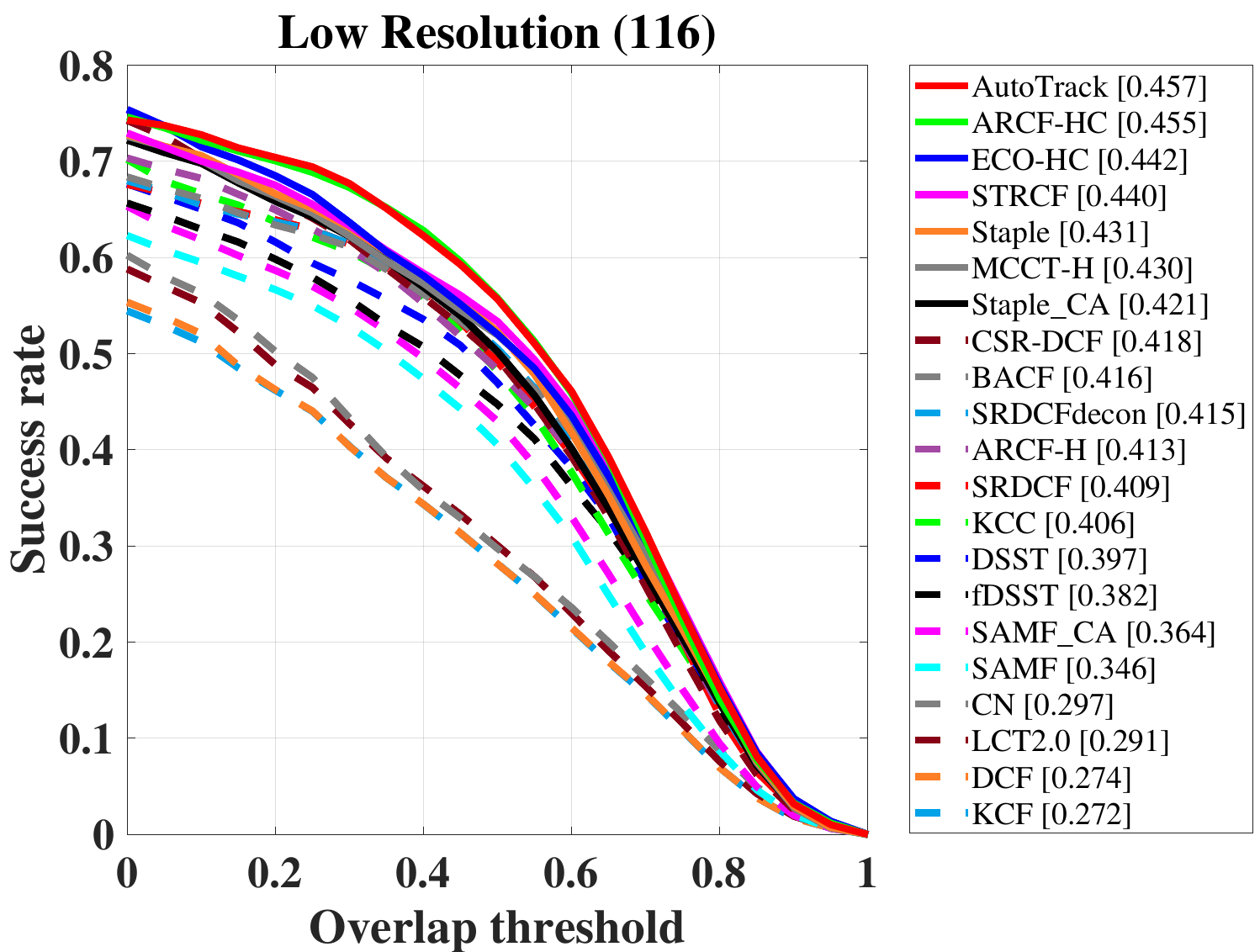}
			\end{minipage}
		}
		\\
		\subfigure { \label{fig:OCC} 
			\begin{minipage}{0.315\textwidth}
				\centering
				\includegraphics[width=1\columnwidth]{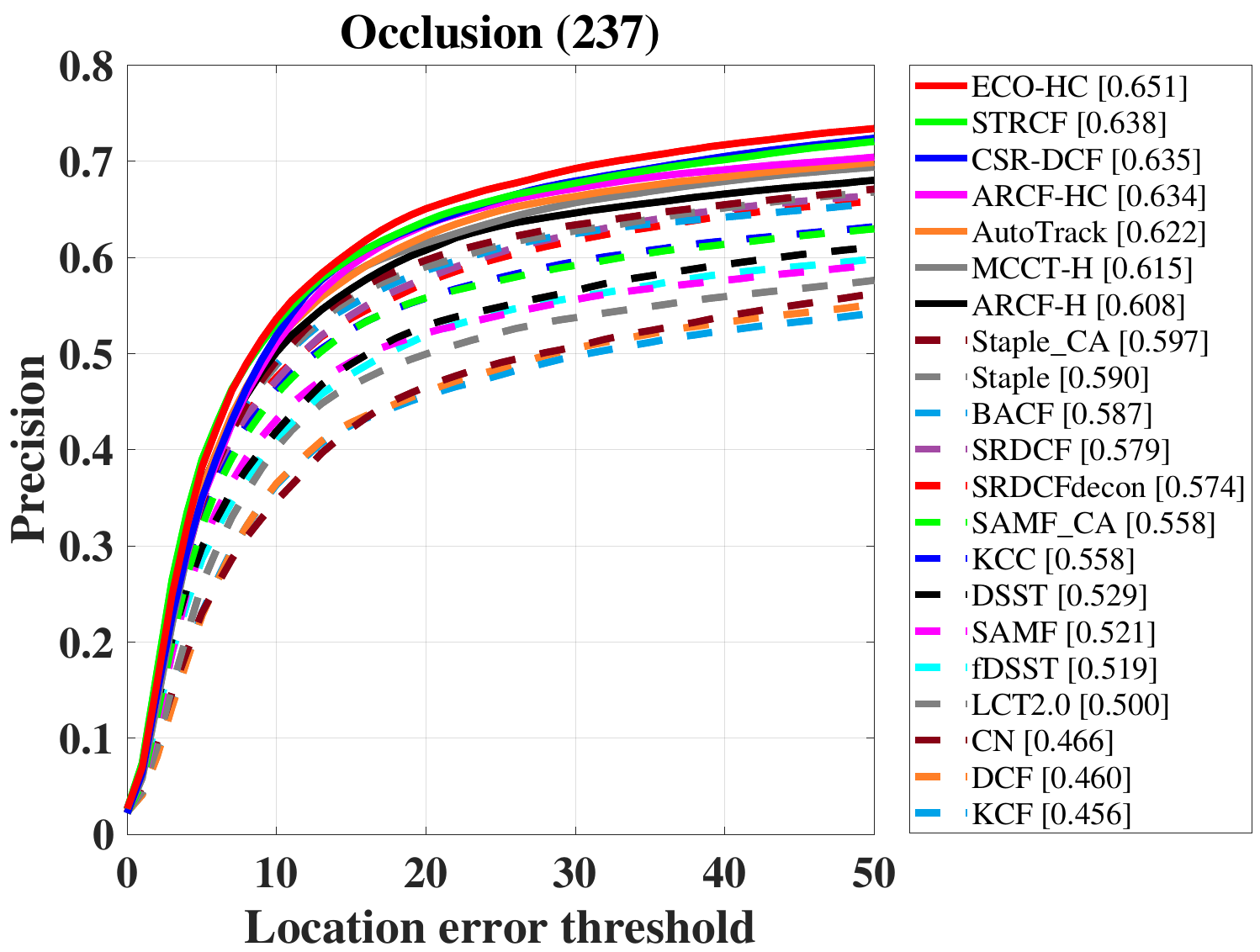}
				\includegraphics[width=1\columnwidth]{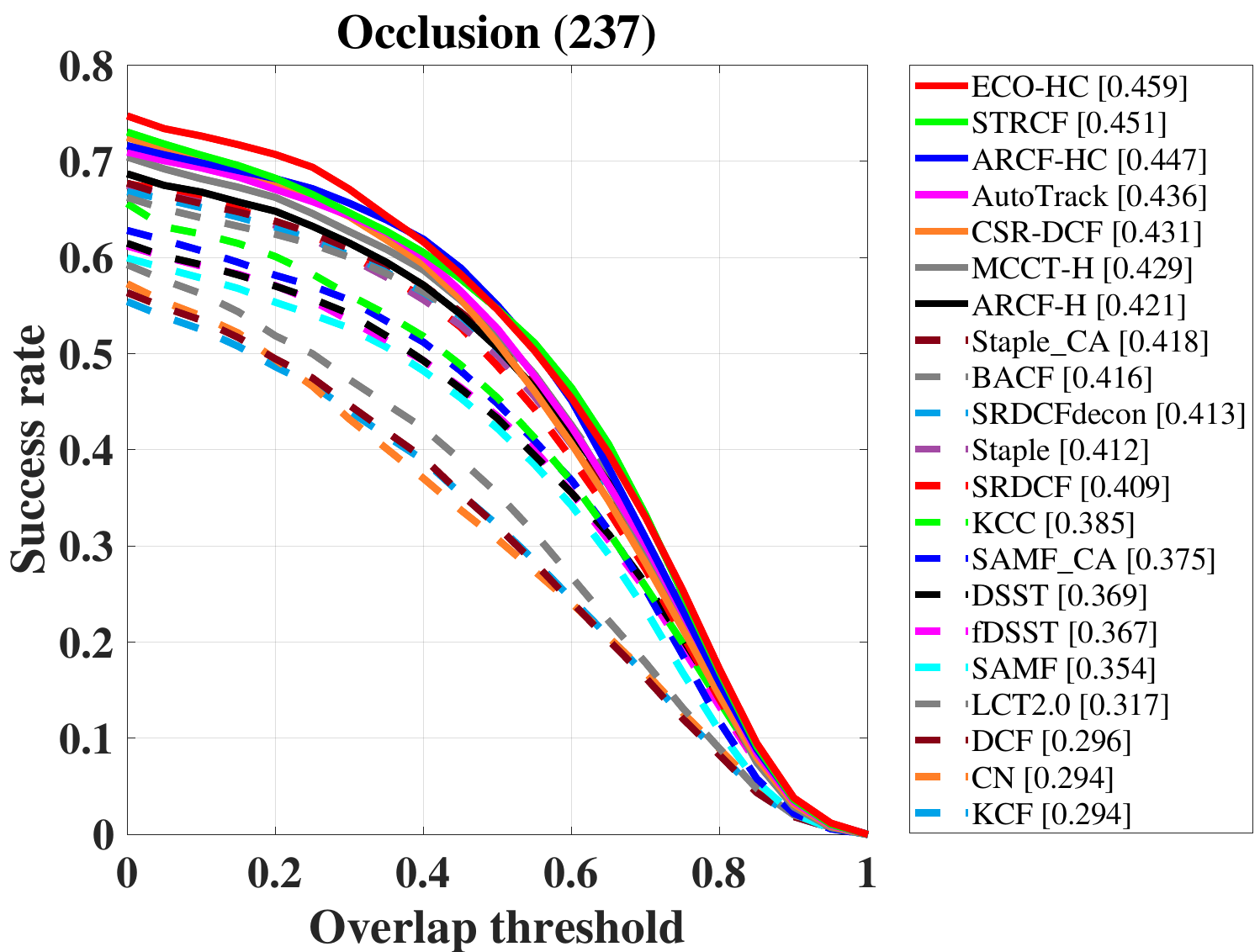}
			\end{minipage}
		}
		\subfigure { \label{fig:IV} 
			\begin{minipage}{0.315\textwidth}
				\centering
				\includegraphics[width=1\columnwidth]{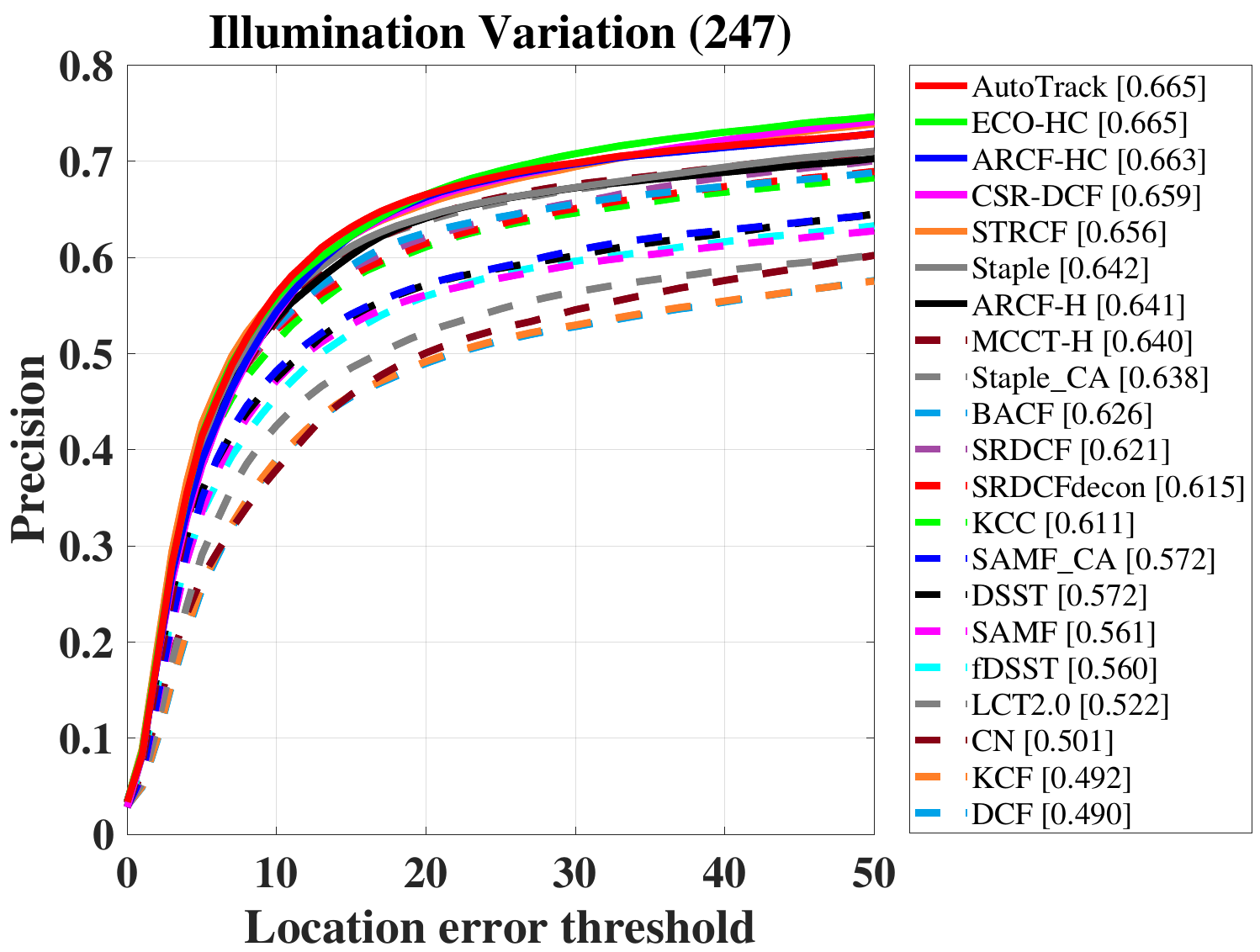}
				\includegraphics[width=1\columnwidth]{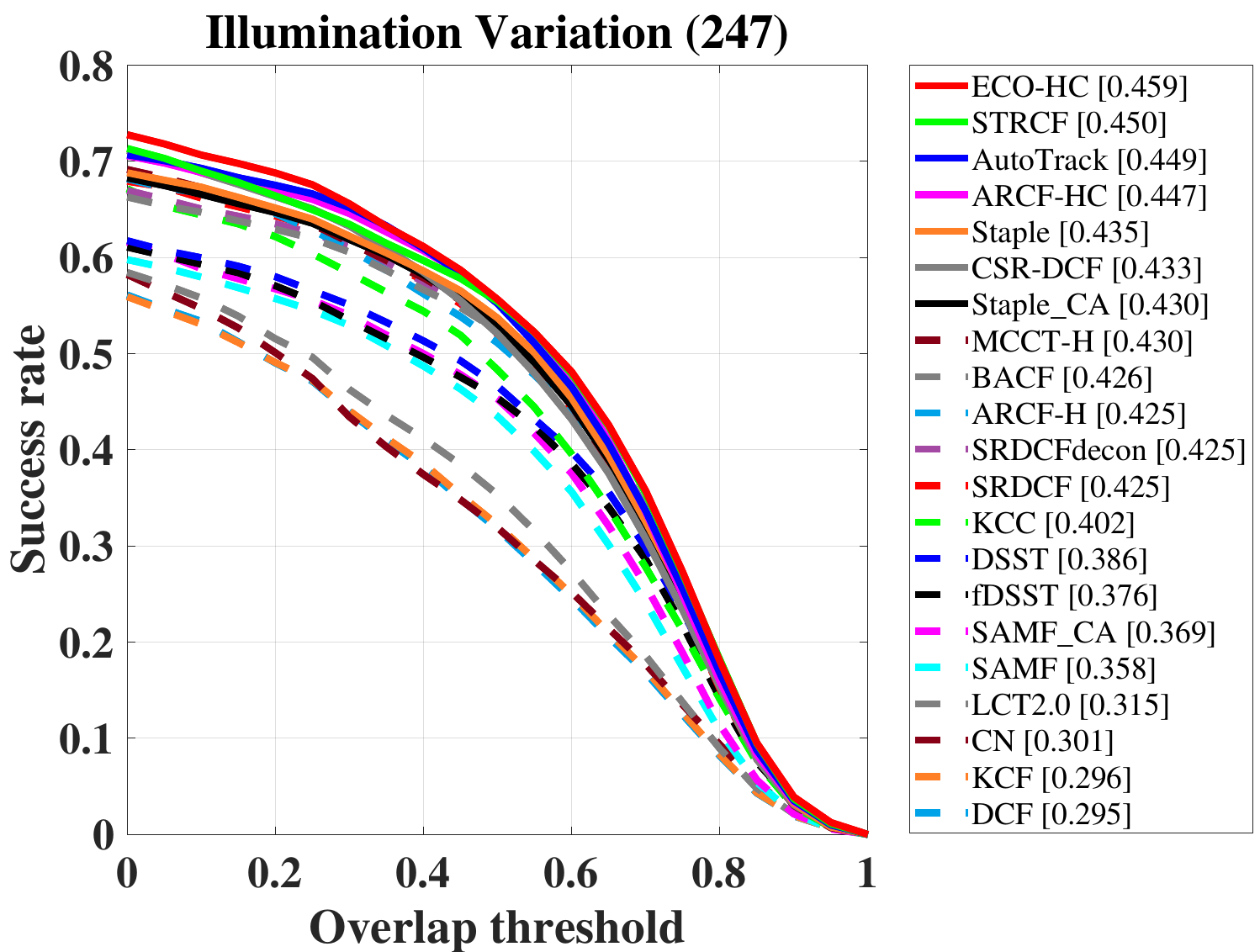}
			\end{minipage}
		}
	\end{center}
\vspace{-0.2cm}
	\caption{Tracking results of each tracker under different attributes, FM, LR, VC, OCC, and IV. The number after the attribute indicates the total number of sequences with the specific attribute, \emph{e.g.}, the total number of sequences with the FM attribute in the six benchmarks is 69. The ranking standard in the precision plot is the precision (DP) under CLE = 20 pixels, and the standard in success rate plot is the area under curve (AUC). For better display effect, please refer to the electronic version of this paper.}
	\label{fig:attributes}
	\vspace{-0.4cm}
	%\label{fig:onecol}
\end{figure*}

In object tracking, to evaluate the performance of the trackers in various challenging scenes, each benchmark puts forward a special series of tracking scenes, which is called the attribute. The defined attributes are indicated in each sequence, whether have or not, for further comparison of the trackers under special attribute. The attributes' full names in Table~\ref{tab:5} are listed as follows: fast motion (FM), camera motion (CM), camera rotation (CR), full occlusion (FOC), occlusion (OCC), partial occlusion (POC), large occlusion (LO), illumination variation (IV), low resolution (LR), small object (SO), viewpoint change (VC).

To better illustrate the performance of the DCF-based trackers in responding to different challenges in UAV tracking scenarios, the six benchmarks' most commonly encountered tracking scenarios are summarized into fast moving (FM), viewpoint change (VC), low resolution (LR), occlusion (OCC), and illumination variation (IV).

The original attributes of each benchmark were firstly mapped to the five attributes, and each sequence was relabeled. For UAV123, UAV123@10fps, UAV20L, and Visdrone2019-SOT, whose original attributes are the same. Their attributes of camera motion and viewpoint change are classified as the VC. The original attributes fast motion, low resolution, and illumination variation are the same as in this work,\emph{i.e.}, FM, LR, and IV. Partial occlusion and full occlusion are classified as OCC. For DTB70, their original attribute fast camera motion is classified as VC, and occlusion is classified as OCC in this work. For UAVDT, their attributes camera rotation, small object, illumination variation, and large occlusion correspond to VC, LR, IV, and OCC respectively. Table~\ref{tab:5} shows the correspondence between the original attributes in the six benchmarks and the new attributes, and the serial number contribution of each benchmark to each new attribute. Above the horizontal line is the sequence number of each benchmark under the new attribute, and below is the original attribute(s) of each benchmark corresponding to the new attribute.

Based on the work above, the tracking results of each tracker under the new attributes were also drawn into ten plots containing success rate plots and precision plots, as is shown in Fig.~\ref{fig:attributes}. Note that the calculation method adopted is to average by sequence, that is, to count the sequences containing specific attributes in all the sequences in the six benchmarks, and the arithmetic average of the results of sequences involved is used as the final result. Figure~\ref{fig:attributes} shows that the performance of trackers under specific attributes is obviously worse than the overall performance in each benchmark, and the challenges that each tracker does well in are also different. Overall speaking, the ARCF-HC tracker \cite{Huang2019ICCV} performs the best in FM and VC, the AutoTrack tracker \cite{Li2020CVPR} is good at VC and IV, and ECO-HC \cite{danelljan2017CVPR} tracker ranked the first in occlusion issues.

\begin{figure}[!t]
	\includegraphics[width=1.02\columnwidth]{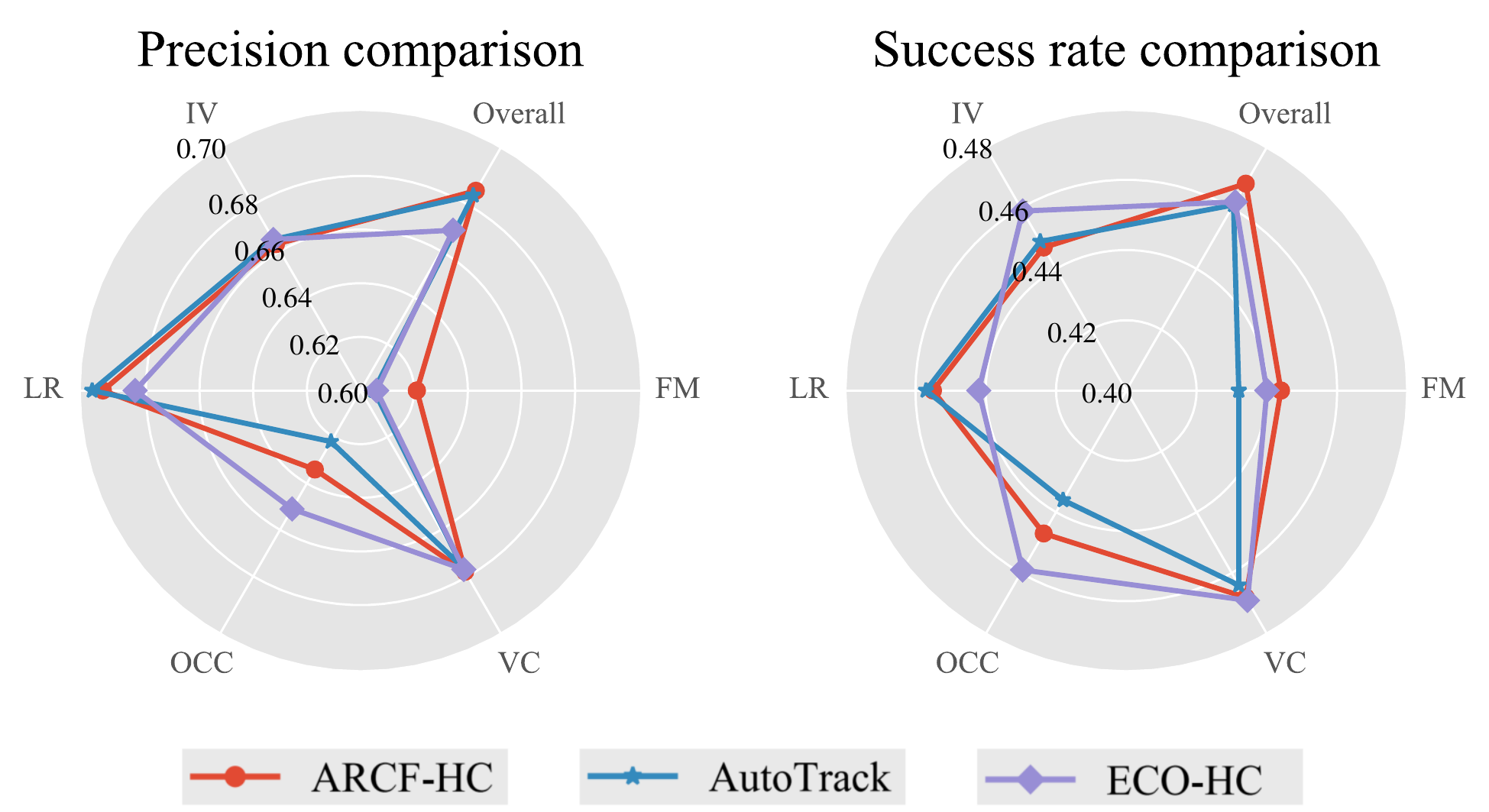}
	\vspace{-0.3cm}
	\caption{The performance comparison of the three best-performing trackers. This figure uses radar charts to show their capabilities under different attributes, whose specific values are in Fig.~\ref{fig:attributes}. Note that the overall data is also the average value of all the sequences in the six benchmarks. For better display effect, please refer to the electronic version of this paper.}
	\label{fig:7}
\end{figure}

Figure~\ref{fig:7} shows the capability comparison of the three best-performing trackers under different challenges, \emph{i.e.}, ARCF-HC tracker \cite{Huang2019ICCV}, AutoTrack tracker \cite{Li2020CVPR}, and ECO-HC tracker \cite{danelljan2017CVPR}. As far as each attribute is concerned, the performance of the three trackers under the two attributes, OCC and FM, is worse than that of IV, LR, and VC, indicating that the two major tracking scenarios of occlusion and fast motion are currently more challenging in UAV tracking. As far as the three trackers are concerned, in the comparison of the success rates, the three have similar performance in overall and VC, while the ECO tracker \cite{danelljan2017CVPR} has a higher success rate in IV, LR, and OCC, and the ARCF-HC tracker \cite{Huang2019ICCV} does better in OCC. The ARCF-HC and AutoTrack tracker both do better in LR compared to the ECO-HC tracker. In the precision comparison, there is little difference among the three in IV, while the overall precision of the ARCF-HC tracker \cite{Huang2019ICCV} and the AutoTrack tracker \cite{Li2020CVPR} is significantly higher than that of the ECO-HC tracker \cite{danelljan2017CVPR}. The ARCF-HC, AutoTrack, and ECO-HC tracker are more accurate under FM, LR, and OCC respectively.

\vspace{-0.3cm}
\subsection{Against deep trackers}\label{Sec:4.4}

\begin{table}[!t]
	\centering
	\setlength{\tabcolsep}{5.7mm}
	\fontsize{8}{12}\selectfont
	\begin{threeparttable}
		\caption{The deep trackers and handcrafted DCF-based tracker's precision at CLE = 20 pixels (DP) and tracking speed (FPS) comparison under UAVDT. The last column indicates whether GPU is used, which is not commonly used in UAV platform. \textbf{\textcolor[rgb]{ 1,  0,  0}{Red}}, \textbf{\textcolor[rgb]{ 0,  1,  0}{green}}, and \textbf{\textcolor[rgb]{ 0,  0,  1}{blue}} respectively mean the first, second and third place.}
		\vspace{0.08cm}
		\begin{tabular}{cccc}
			\toprule[1.5pt]
			Tracker & DP& FPS&GPU\\
			\midrule
			AutoTrack \cite{Li2020CVPR}& \textbf{\textcolor{red}{0.718}} &\textbf{ \textcolor{blue}{56.38}}&\XSolid\\
			SiamFC \cite{Bertinetto2016ECCV}&  \textbf{\textcolor{green}{0.704}} & 39.58&\Checkmark\\
			DSiam \cite{Guo2017ICCV}& \textbf{\textcolor{blue}{0.704}}&20.25&\Checkmark \\
			ASRCF \cite{Dai2019CVPR} & 0.700 & 22.20&\Checkmark\\
			ECO \cite{danelljan2017CVPR}& 0.699 & 16.87&\Checkmark\\
			UDT+ \cite{wang2019CVPR} & 0.696 & \textbf{\textcolor{green}{56.92}}&\Checkmark \\
			CFWCR \cite{he2017CVPR}& 0.691 & 9.46&\Checkmark\\
			CFNet\_conv2 \cite{valmadre2017CVPR}& 0.681 & 44.65&\Checkmark\\
			TADT \cite{li2019CVPR}& 0.677 & 32.32&\Checkmark\\
			MCPF \cite{zhang2017CVPR}& 0.675 & 0.58&\Checkmark\\
			UDT \cite{wang2019CVPR}& 0.674 & \textbf{\textcolor{red}{73.31}}&\Checkmark\\
			MCCT \cite{wang2018CVPR}& 0.671 & 7.86&\Checkmark\\
			DeepSTRCF \cite{Li2018CVPR}& 0.667 & 6.84&\Checkmark\\
			CoKCF \cite{zhang2017PR}& 0.605 & 20.15&\Checkmark\\
			IBCCF \cite{li2017ICCV}& 0.603 & 3.02&\Checkmark\\
			HCFT \cite{Ma2015ICCV}& 0.602 & 19.00&\Checkmark\\
			\bottomrule[1.5pt]			
			\label{tab:6}
		\end{tabular}
	\end{threeparttable}
\vspace{-0.7cm}
\end{table}

\begin{figure*}[!t]
	\centering
	\includegraphics[width=2.1\columnwidth]{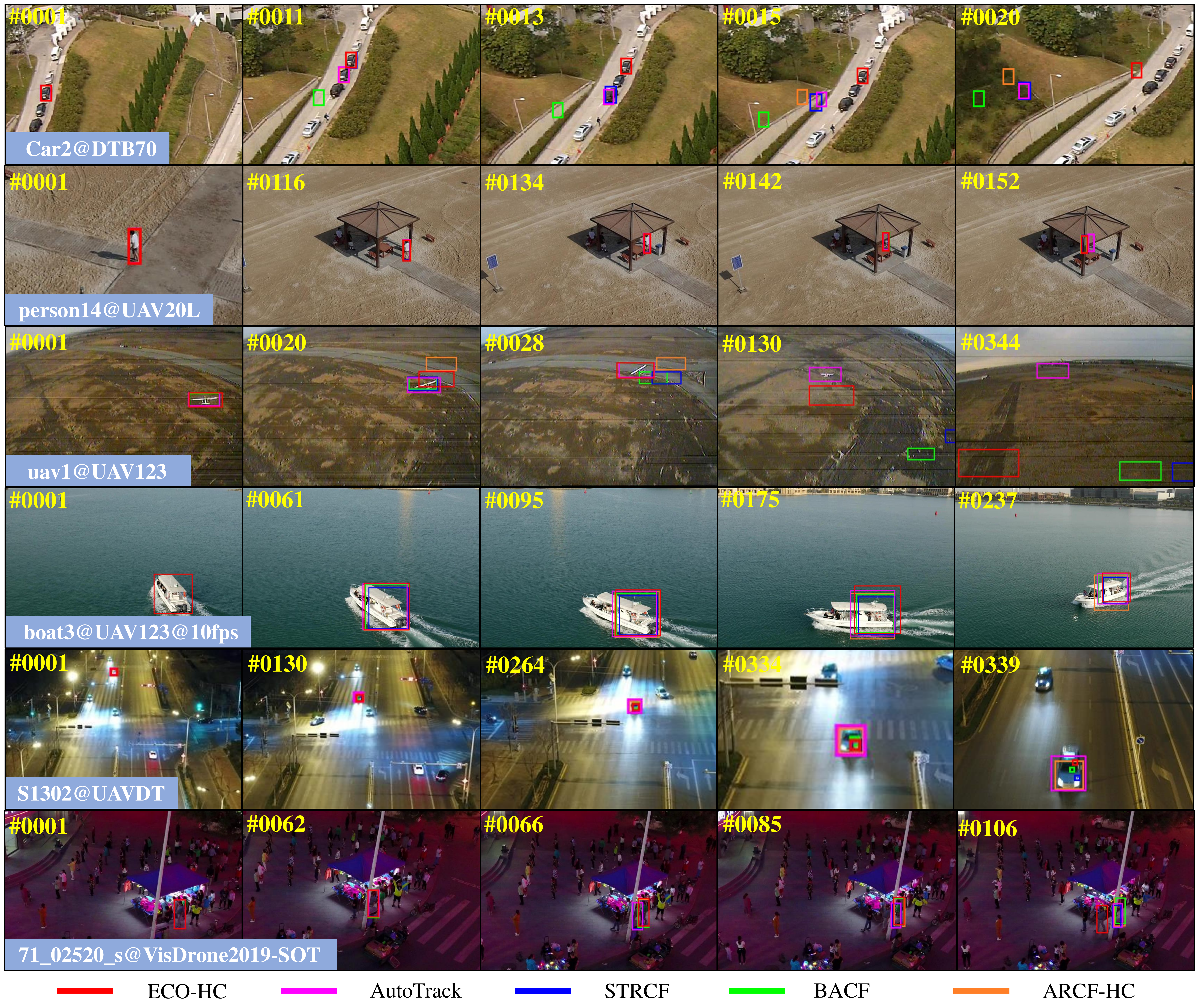}
	\vspace{-0.6cm}
	\caption{Representative tracking failure cases in 6 popular UAV tracking benchmarks. The sequences and the corresponding benchmarks are labeled at the left corner of each row, \emph{i.e.}, Car2@DTB70~\cite{li2017AAAI}, person14@UAV20L~\cite{mueller2016ECCV}, uav1@UAV123~\cite{mueller2016ECCV}, boat3@UAV123@10fps~\cite{mueller2016ECCV}, S1302@UAVDT~\cite{du2018ECCV}, and 71\_02520\_s@Visdrone2019-SOT~\cite{du2019ICCVW} (from the first row to the last). The trackers' predicted boxes are marked out by different colors, \emph{e.g.}, \textbf{\textcolor{red}{red}} for the ECO-HC tracker. For better display effect, please refer to the electronic version of this paper.}
	\label{fig:8}
	\vspace{-0.5cm}
\end{figure*}

\begin{figure*}[!t]
	\centering
	\includegraphics[width=2\columnwidth]{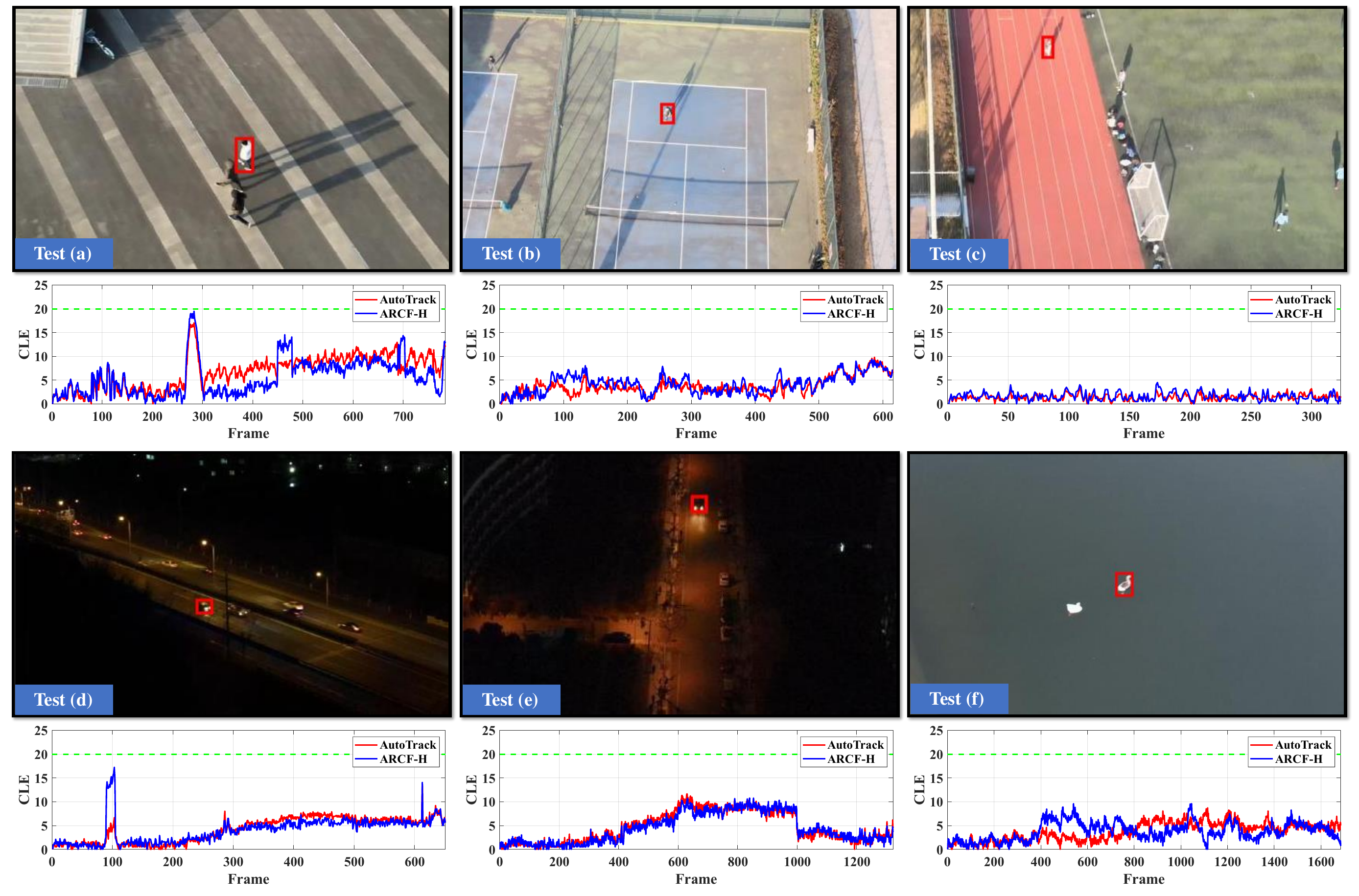}
	\vspace{-0.4cm}
	\caption{Onboard tracking performance in terms of center location error (CLE). The six tests (a), (b), (c), (d), (e), and (f), where the tracking objects are marked out by red boxes, contain the common UAV tracking challenges aforementioned. The satisfying CLE results displayed validated the robustness of AutoTrack \cite{Li2020CVPR} and ARCF-H \cite{Huang2019ICCV} tracker in challenging real-world UAV tracking. For better display effect, please refer to the electronic version of this paper.}
	\vspace{-0.4cm}
	\label{fig:onboardcle}
	
	%\label{fig:onecol}
\end{figure*}

In order to better demonstrate the superiority of the DCF-based trackers using handcrafted features in the UAV tracking scenes, various deep trackers are selected, including those adopting basic CF framework but using CNN features, \emph{i.e.}, the ASRCF tracker \cite{Dai2019CVPR}, the ECO tracker \cite{danelljan2017CVPR}, the CFWCR tracker \cite{he2017CVPR}, the MCCT tracker \cite{wang2018CVPR}, the MPCF tracker \cite{zhang2017CVPR}, the DeepSTRCF tracker \cite{Li2018CVPR}, the CoKCF tracker \cite{zhang2017PR}, the IBCCF tacker \cite{li2017ICCV}, the HCFT tracker \cite{Ma2015ICCV} and also the end-to-end CNN trackers, \emph{i.e.}, the UDT+ tracker \cite{wang2019CVPR}, the SiamFC tracker \cite{Bertinetto2016ECCV}, the CFNet\_conv2 tracker \cite{valmadre2017CVPR}, the TADT  tracker\cite{li2019CVPR}, the UDT tracker \cite{wang2019CVPR}, and the DSiam tracker \cite{Guo2017ICCV}.

\Remark All the deep trackers used GPU acceleration in the experiment, and all the DCF-based trackers using handcrafted features, \emph{e.g.}, the AutoTrack tracker, were evaluated on a single CPU using one core only.

\Remark According to Table~\ref{tab:3} and focusing on AUC and DP, the AutoTrack tracker \cite{Li2020CVPR}, which achieved the most top 3 scores and the most first scores in 6 benchmarks, is considered the most outstanding DCF-based tracker. This subsection chooses AutoTrack to demonstrate DCF-based trackers' superiority against the deep trackers in UAV tracking.

Table~\ref{tab:6} listed their DP and tracking speed under benchmark UAVDT \cite{du2018ECCV}. As is shown in their tracking performance, even with low-cost handcrafted features, the AutoTrack tracker \cite{Li2020CVPR} still prevailed most deep trackers, achieving both excellent precision and considerable tracking speed.

According to the experimental results, even compared with the splendid deep trackers, the handcrafted DCF-based trackers still guard their strong competitiveness in accuracy and possess the real-time performance, which further confirms that the DCF-based trackers using handcrafted features, \emph{e.g.}, the AutoTrack tracker \cite{Li2020CVPR} \emph{etc.}, is the best choice for UAV-based aerial tracking.

\subsection{Failure cases and challenges}\label{Sec:4.5}
Though the brilliant DCF-based methods have exhibited their superiority in UAV tracking, there still exist tracking challenges that have not been well-addressed. This subsection analyzes the classic tracking failure cases of the five best-performing trackers, the AutoTrack tracker \cite{Li2020CVPR}, the ARCF-HC tracker \cite{Huang2019ICCV}, the STRCF tracker \cite{Li2018CVPR}, the ECO-HC tracker \cite{danelljan2017CVPR}, and the BACF tracker \cite{kiani2017ICCV}, in each benchmark in the experiment to illustrate the current limitations and challenges of handcrafted DCF-based trackers. Figure~\ref{fig:8} shows representative tracking failures in the different benchmarks.

(1) Rapid scale changes and other appearance variations are currently difficult for DCF-based trackers to deal with. In the fourth and fifth rows of Fig.~\ref{fig:8}, when the object undergoes rapid appearance variations caused by view point change or scale variations, the trackers can not adapt to the appearance changes in time, making the wrong location and scale estimations. Such scenes usually cause the filter to learn the wrong object information and eventually lead to tracking failure.

(2) The occlusion problem is often hard to tackle by the trackers. In the second row of Fig.~\ref{fig:8}, when the object is completely occluded and appears outside the search region of the trackers again, the trackers cannot predict the object location. In the sixth row of Fig.~\ref{fig:8}, even partial occlusion can seriously affect the object template learned by the tracker, resulting in tracking failure.

(3) Low-resolution objects are much more challenging to track than other objects. As shown in the first and third rows of Fig.~\ref{fig:8}, low-resolution objects lead to insufficient training samples of the filter, which may reduce the filter’s ability to discriminate the object out of the background. When such object undergoes fast motion, it can more easily cause tracking failure due to the filter's poor discriminating ability. In the fifth row of Fig.~\ref{fig:8}, the small object can also cause wrong scale estimation, where when the object scale changes, the tracker can't adapt in time.

(4) Poor lighting conditions and illumination variation can make tracking harder. As shown in the fifth and sixth rows of Fig.~\ref{fig:8}, in the relatively dim environment, the filter cannot learn enough representative object features, thus it is difficult to distinguish the object from the environment. Under such conditions, the presence of scale variation, partial occlusion, or similar object makes robust tracking even harder.

\section{Onboard Evaluation}\label{Section 5}
Apart from the large-scale evaluation experiments above, this work also extended onboard test to further validate the real-time capability and robustness of the outstanding DCF-based trackers \cite{Li2020CVPR,Huang2019ICCV}. This evaluation adopted a typical CPU-based onboard PC for UAV, Intel NUC8i7HVK, which contains a single Intel Core i7-8809G CPU, 32GB RAM, as the test platform.

Onboard tracking performance of six tests have been shown in Fig.~\ref{fig:onboardcle} with the AutoTrack \cite{Li2020CVPR} and ARCF-H \cite{Huang2019ICCV} trackers. The six tests, \textit{i.e.}, (a), (b), (c), (d), (e), and (f), which contain the challenges, \textit{e.g.}, LR, IV, VC, \textit{etc.}, commonly encountered in UAV tracking (including one long-term tracking sequence). Table~\ref{tab:obsped} displayed the challenges in each test and the running speed of the two trackers, where they both surpass 30FPS, realizing real-time processing. In Fig.~\ref{fig:onboardcle}, the CLE of the two trackers in 6 tests are all smaller than 20 pixels, indicating that the brilliant trackers maintained satisfying robustness in real-world challenging UAV tracking scenes.

\Remark The onboard tracking has verified the promising efficiency of DCF-based methods. Such superiority in computation saves the scarce power supply for other energy-consuming functions like self-control in strong wind onboard UAV.

\begin{table}[!b]
	\centering
	\setlength{\tabcolsep}{4mm}
	\caption{Running speed of AutoTrack \cite{Li2020CVPR} and ARCF-H \cite{Huang2019ICCV} tracker and challenges encountered in the onboard test. Clearly, even with limited computation resources, the brilliant trackers can achieve real-time for UAV tracking.}
	\begin{tabular}{cccc}
		\toprule
		\multirow{2}[4]{*}{Test} & \multirow{2}[4]{*}{Challenges} & \multicolumn{2}{c}{FPS} \\
		\cmidrule{3-4}          &       & AutoTrack & ARCF-H \\
		\midrule
		\midrule
		(a)     & LR, OCC   & 33.58 & 50.47 \\
		(b)     & VC, LR  & 37.59 & 68.05 \\
		(c)     & LR   & 36.95 & 81.22 \\
		(d)     & LR, OCC, IV   & 36.42 & 43.47 \\
		(e)     & VC, LR, IV  & 36.92 & 57.48 \\
		(f)     & Long-term  & 40.21 & 57.51 \\
		\bottomrule
	\end{tabular}%
	\label{tab:obsped}%
\end{table}%

\section{Future Works}\label{Section 6}

The future investigation and improvement work on DCF-based trackers can be summarized into four points: 
\begin{itemize}
\item Exploiting more adaptive correlation filters, with different learning rates or adaptive parameters for various tracking scenes and objects.
\item Researching for a more intelligent search strategy, which can better adapt to the situations where objects move faster or reappear at a far location after being occluded.
%\item Add sample preprocessing to suppress noise and distractors in training samples or perform object segmentation in advance. Such preprocessing strategies can make the filter learn more 
\item Embedding image preprocessing strategies into basic tracking structure to boost the trackers' discriminative ability under complex scenes. Such strategies include but are not limited to: low-light image enhancers, object segmentation methods, saliency detection algorithms, \textit{etc.} 
\item Studying multi-modal trackers, such as infrared modalities, to cope with tracking scenes with poor lighting conditions.
\end{itemize}

It can be believed that the prospect of DCF-based methods onboard UAV tracking is promising, which can promote the development and application of UAVs, thus boosting the progress of the remote sensing field.

\section{Conclusion}\label{Section 7}
This work firstly introduces the tracking scenes onboard UAV, the uniqueness and challenge of UAV tracking compared with general tracking scenes, and what makes DCF-based methods suitable for UAV-based aerial tracking in comparison with other types of trackers. Next, the DCF-based tracking algorithms' common basic architecture is proposed for overall understanding. Thirdly, this work introduces the highlights of noted DCF-based tracker, focusing on their contributions, thus integrating DCF-based trackers' developments over the years. Then, in the experiment section, having introduced some implementation information, exhaustive experiments are conducted on six UAV tracking benchmarks to estimate all mentioned DCF-based trackers (both generally and by attribute) and demonstrate the superiority of their tracking. Based on the experiment results, this work further analyzes the current tracking challenges. Moreover, additional onboard tracking tests are extended to validate the real-time capability and robustness of the brilliant DCF-based trackers in challenging real flight UAV tracking tasks. Finally, future research directions and improvement work are summarized, guiding more fruits of DCF for UAV tracking.

\ifCLASSOPTIONcompsoc
\section*{Acknowledgments}
\else
\section*{Acknowledgment}
\fi
The work is supported by the National Natural Science Foundation of China under Grant 61806148 and the Natural Science
Foundation of Shanghai under Grant 20ZR1460100.

% or


\ifCLASSOPTIONcaptionsoff
  \newpage
\fi



\bibliographystyle{IEEEtran}

\bibliography{reference}

\vspace{-1cm}
\begin{IEEEbiography}[{\includegraphics[width=1in,height=1.25in,clip,keepaspectratio]{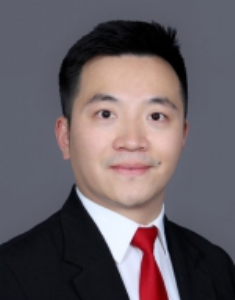}}]{Changhong Fu}
(Member, IEEE) received the Ph.D. degree in robotics and automation from the Computer Vision and Aerial Robotics (CVAR) Laboratory,
Technical University of Madrid, Madrid, Spain, in 2015. During his Ph.D., he held two research positions at Arizona State University, Tempe, AZ, USA, and Nanyang Technological University (NTU), Singapore. After receiving his Ph.D., he worked at NTU as a Post-Doctoral Research Fellow. He is currently an Associate Professor with the School of Mechanical Engineering, Tongji University, Shanghai, China, and leading seven  projects related to the vision for unmanned systems (US). He has worked on more than ten projects related to the vision for UAV. In addition, he has published more than 60 journal and conference papers (including the IEEE TGRS, IEEE TCSVT, IEEE TMM, IEEE TMECH, IEEE TIE, CVPR, ICCV, ICRA, IROS) related to the intelligent vision and control for UAV. His research areas are intelligent vision and control for US in a complex environment.
\end{IEEEbiography}

\vspace{-1cm}

\begin{IEEEbiography}[{\includegraphics[width=1in,height=1.25in,clip,keepaspectratio]{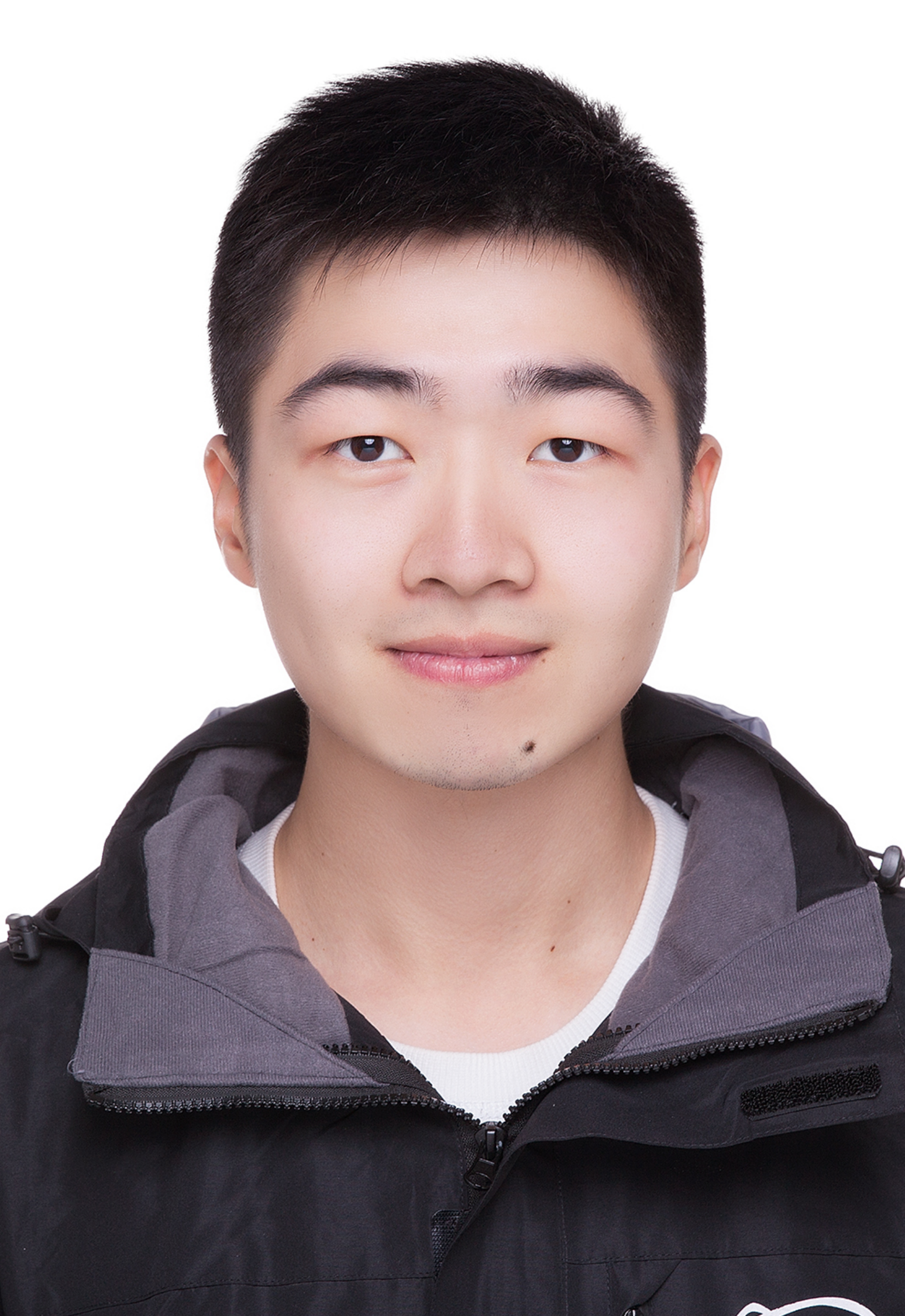}}]{Bowen Li} is pursuing the B.Eng. degree with the School of Mechanical Engineering, Tongji University,
Shanghai, China. His research interests include robotics, artificial intelligence and computer vision.
\end{IEEEbiography}

\vspace{-1.5cm}

\begin{IEEEbiography}[{\includegraphics[width=1in,height=1.25in,clip,keepaspectratio]{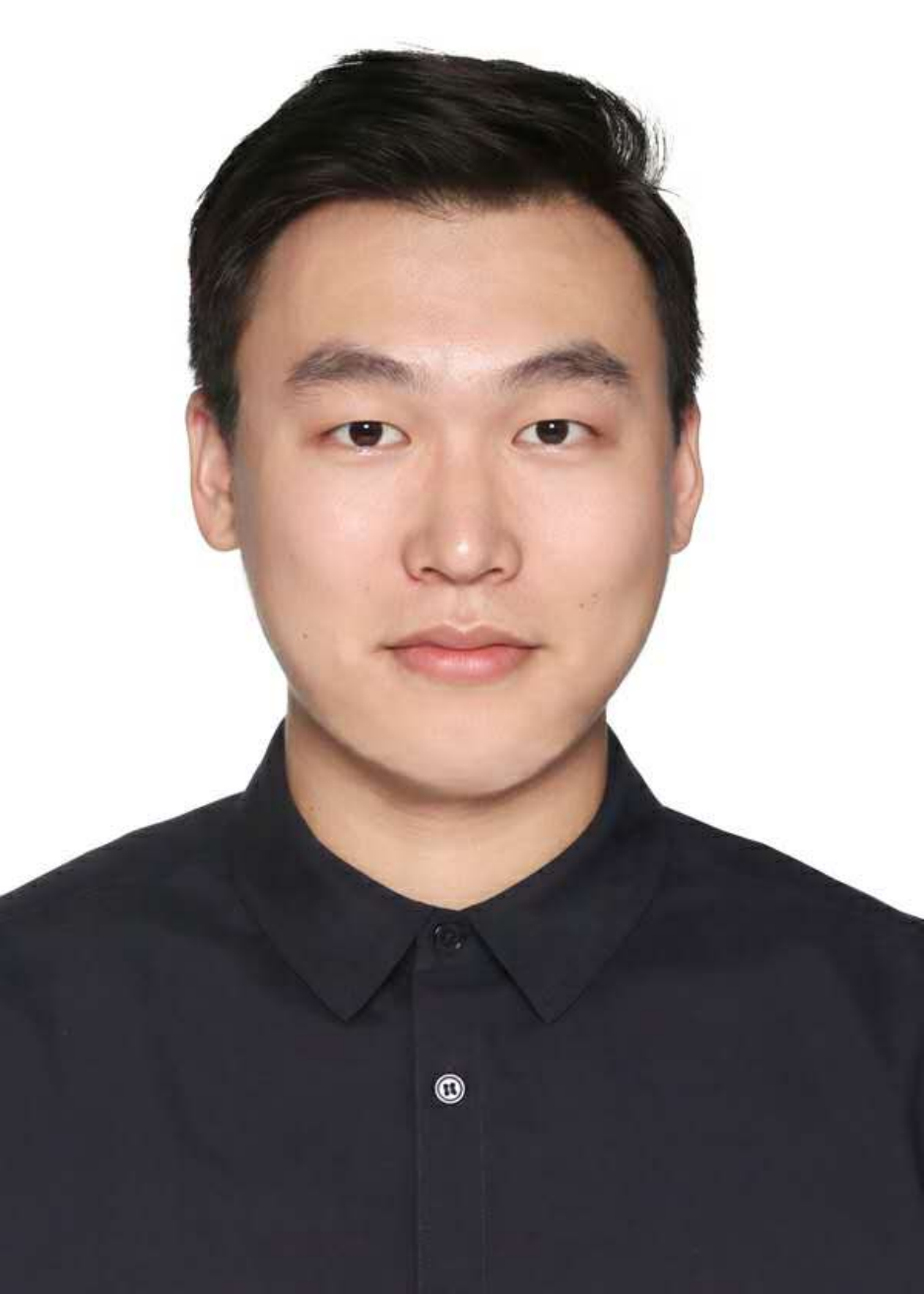}}]{Fangqiang Ding} is pursuing the B.Eng. degree with the School of Mechanical Engineering, Tongji University,
Shanghai, China. His research interests include unmanned aerial vehicle and computer vision.
\end{IEEEbiography}

\vspace{-1.5cm}

\begin{IEEEbiography}[{\includegraphics[width=1in,height=1.25in,clip,keepaspectratio]{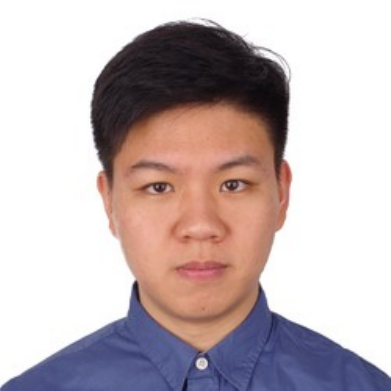}}]{Fuling Lin} received his B.Eng. degree in mechanical engineering from Tongji University, Shanghai, China. He is currently pursuing M.Sc. degree in mechanical engineering in Tongji University, Shanghai, China. His research interests include robotics, visual object tracking and computer vision.
\end{IEEEbiography}

\vspace{-1.5cm}

\begin{IEEEbiography}[{\includegraphics[width=1in,height=1.25in,clip,keepaspectratio]{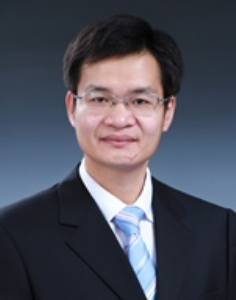}}]{Geng Lu} received the B.E., M.E., and the Ph.D. degrees from the Department of Automation, Tsinghua
University, Beijing, China, in 1999, 2002, and 2004, respectively. From 2004 to 2006, he was a Postdoctorate Scholar with the Department of Electrical Engineering, Tsinghua University. Since 2006, he has been with the Department of Automation, Tsinghua University, where he is currently an Associate Professor. His research interests include robust control, nonlinear control, signal processing, and aerial robot.
\end{IEEEbiography}

\end{document}